\documentclass{article}

\usepackage{microtype}
\usepackage{graphicx}
\usepackage{subcaption}
\usepackage{comment}
\usepackage{booktabs} 

\usepackage{hyperref}

\usepackage[accepted]{icml2026}

\usepackage{amsmath}
\usepackage{amssymb}
\usepackage{mathtools}
\usepackage{amsthm}
\usepackage{enumitem}
\usepackage{multirow}

\usepackage{xcolor}

\newcommand{\model}{EpiNode}

\usepackage{pifont}

\newcommand{\cmark}{\ding{51}}  
\newcommand{\xmark}{\ding{55}}  

\usepackage[capitalize,noabbrev]{cleveref}

\theoremstyle{plain}

\theoremstyle{definition}

\theoremstyle{remark}

\usepackage[textsize=tiny]{todonotes}

\icmltitlerunning{How (Not) to Hybridize Neural and Mechanistic Models for Epidemiological Forecasting}

\begin{document}

\twocolumn[
  \icmltitle{
  How (Not) to Hybridize Neural and Mechanistic Models \\for Epidemiological Forecasting
  }



  \icmlsetsymbol{equal}{*}

  \begin{icmlauthorlist}
    \icmlauthor{Yiqi Su}{yyy}
    \icmlauthor{Ray Lee}{yyy}
    \icmlauthor{Jiaming Cui}{yyy}
    \icmlauthor{Naren Ramakrishnan}{yyy}
  \end{icmlauthorlist}

  \icmlaffiliation{yyy}{Department of Computer Science, Virginia Tech, Alexandria, USA}

  \icmlcorrespondingauthor{Naren Ramakrishnan}{naren@vt.edu}

  \icmlkeywords{Machine Learning, ICML}

  \vskip 0.3in
]



\printAffiliationsAndNotice{}  


\begin{abstract}
    Epidemiological forecasting from surveillance data is a hard problem and hybridizing mechanistic compartmental models with neural models is a natural direction. The mechanistic structure helps keep trajectories epidemiologically plausible, while neural components can capture non-stationary, data-adaptive effects. In practice, however, many seemingly straightforward couplings fail under partial observability and continually shifting transmission dynamics driven by behavior, waning immunity, seasonality, and interventions. We catalog these failure modes and show that robust performance requires making non-stationarity explicit: we extract multi-scale structure from the observed infection series and use it as an interpretable control signal for a controlled neural ODE coupled to an epidemiological model. Concretely, we decompose infections into trend, seasonal, and residual components and use these signals to drive continuous-time latent dynamics while jointly forecasting and inferring time-varying transmission, recovery, and immunity-loss rates. 
    Across early outbreak and multi-wave regimes, our approach attains the lowest RMSE on five datasets (21-63\% reduction over the strongest default-configured baseline), 
    achieves the best peak detection accuracy,
    and infers time-varying epidemiological rates within ground-truth ranges, without relying on auxiliary covariates.

\end{abstract}

\section{Introduction}
Given historical epidemic curves, e.g., daily infections, hospitalizations, or deaths, the goal of epidemiological forecasting is to predict future trajectories~\cite{ali2021influenza, held2019forecasting}.
Epidemiologists have developed a broad range of models that have been applied across outbreaks such as H1N1 \cite{reich2019collaborative}, Ebola \cite{viboud2018rapidd}, and COVID-19 \cite{cramer2022evaluation}.

Most forecasting models fall into three main families: compartmental models, statistical or machine learning models, and hybrid models~\cite{rodriguez2024machine}. Compartmental models built on ordinary differential equations (ODEs), such as the SIR (Susceptible-Infectious-Recovered) model, have long formed the foundation of epidemiological modeling \cite{anderson1991infectious,hethcote2000mathematics}. However, they often rely on strong assumptions about parameter stationarity. Meanwhile, statistical approaches such as ARIMA, state-space models, and Gaussian processes have also been widely used, particularly for short forecasting horizons \cite{box2015time,harvey1989structural,rasmussen2006gaussian}. More recently, deep learning models, including recurrent neural networks, temporal convolutional networks, and transformer-based architectures have also achieved strong empirical performance \cite{hochreiter1997long,bai2018empirical,li2019enhancing,zhou2021informer}. Yet these models typically operate as black boxes, lack epidemiological interpretability, and may produce physically inconsistent forecasts for long horizons or under regime shifts \cite{Rudin_2019}. To address this issue, hybrid frameworks that embed neural components within compartmental models have been proposed, including neural ODEs \cite{neural-odes}, latent ODEs \cite{latent-odes}, and physics-informed neural networks \cite{raissi2019physics}. 
In principle, hybrid models can combine mechanistic interpretability with data-driven flexibility; in practice, however, they remain fundamentally challenged by partial observability and non-stationarity.

{\noindent\textbf{Partial observability.} 
Epidemiological data is inherently incomplete: typically only cases/deaths/hospitalizations (or proxies such as symptomatic rates) are observed, while other states (e.g., susceptible or exposed) are latent \cite{fairchild2018epidemiological,ryu2022epidemiology}. Such undetermined systems make both inference and forecasting ill-posed, since multiple latent trajectories can explain the same observations, and full-state supervision is rarely available.

{\noindent \textbf{Non-stationarity.} 
Key epidemiological parameters (e.g., transmission or recovery rates) evolve over time, influenced by factors such as climate, mobility, behavior change, policy interventions, and accumulated immunity \cite{transmission, dureau2013capturing}. For example, influenza transmission rates show substantial seasonal variation \cite{shaman2010absolute}, so assuming constant parameters can induce systematic bias and degrade peak predictions.
Modeling time-varying parameters is therefore essential, but increases estimation complexity
from $O(n)$ to $O(nT)$.

Crucially, these challenges interact and amplify each other: latent states complicate parameter estimation, while parameter misestimation feeds back into incorrect latent-state reconstruction. This vicious cycle worsens with longer forecast horizons.
While short-term forecasts benefit from temporal autocorrelation, accuracy deteriorates rapidly as the horizon grows \cite{cramer2022evaluation}, with errors accumulating due to uncertainty in latent compartments and evolving dynamics. The effect is especially consequential near epidemic peaks, since small parameter uncertainties can lead to widely divergent epidemic trajectories, making peak timing and peak magnitude increasingly difficult to predict \cite{castro2020turning}. This motivates methods that improve long-horizon stability and peak prediction, and a careful accounting of when `hybridization' helps versus hurts.

Our key contributions are:
\vspace{-3mm}

\begin{itemize}
\item 
We characterize and 
demonstrate key failure modes in neural–mechanistic hybrid epidemic models, highlighting how the wrong architecture can destabilize long-horizon rollouts and bias peak predictions.
\vspace{-3mm}
\item 
We propose \model{}, a decomposition-controlled hybrid framework that couples a neural ODE with mechanistic SIRS dynamics to address partial observability and non-stationarity by explicitly modeling the multi-scale structure of the observed infection signal. The mechanistic SIRS layer enforces epidemiological consistency and supports stable long-horizon rollouts (Figure \ref{fig:history_accuracy}). The neural component infers time-varying transmission, recovery, and immunity-loss rates within bounded, interpretable ranges.
\vspace{-3mm}
\item We demonstrate the broad applicability of the proposed model to seasonal epidemics across early-outbreak and multi-wave forecasting regimes, through extensive synthetic and real-world benchmarks. 
\end{itemize}
}

\vspace{-3mm}

    


\begin{figure}[t]
    \centering
    \includegraphics[width=\columnwidth]{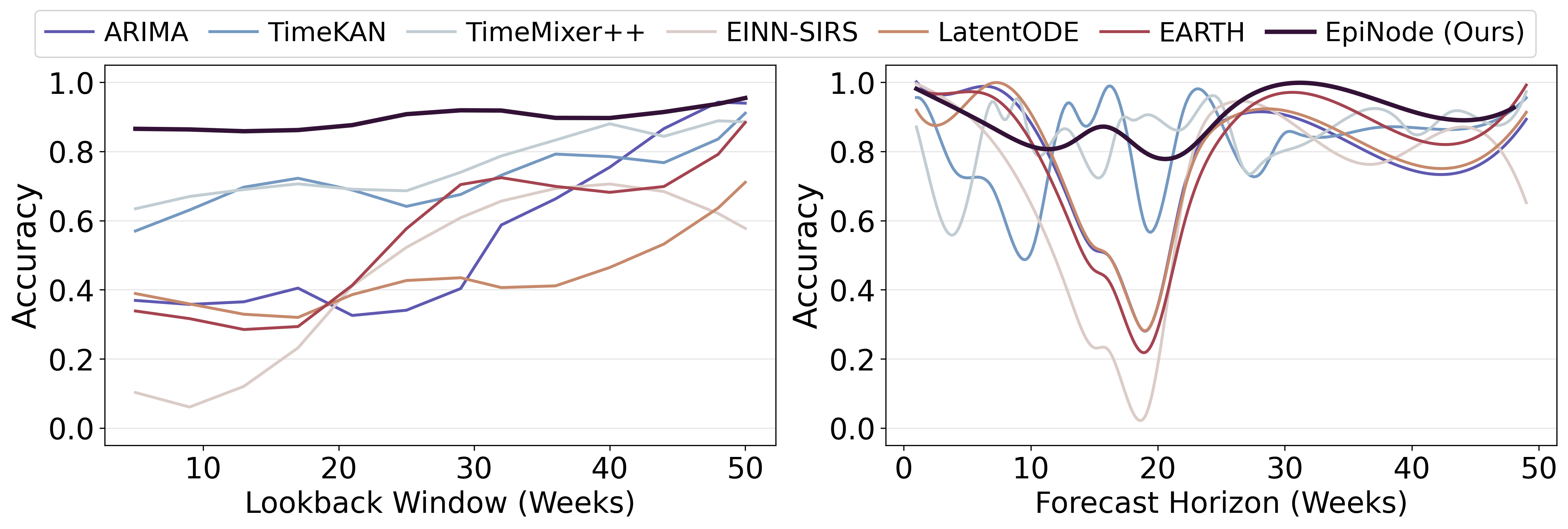}
    \caption{\model{} accuracy as a function of lookback window and forecast horizon.
    } \label{fig:history_accuracy}
\end{figure}

\section{Failure Modes}
\label{sec:failure-modes}

We begin by outlining failure modes specific to epidemiological modeling when using neural ODEs and related machine learning approaches. These modes can be viewed as `antipatterns', i.e., what not to do.

\subsection{Neural ODE architectures fail at long-horizon forecasting even under full observability.}
\label{sec:fm-datadriven}

\begin{figure}[htbp]
    \centering
    \begin{subfigure}[t]{0.49\columnwidth}
        \centering
        \includegraphics[width=\linewidth, height=3cm]{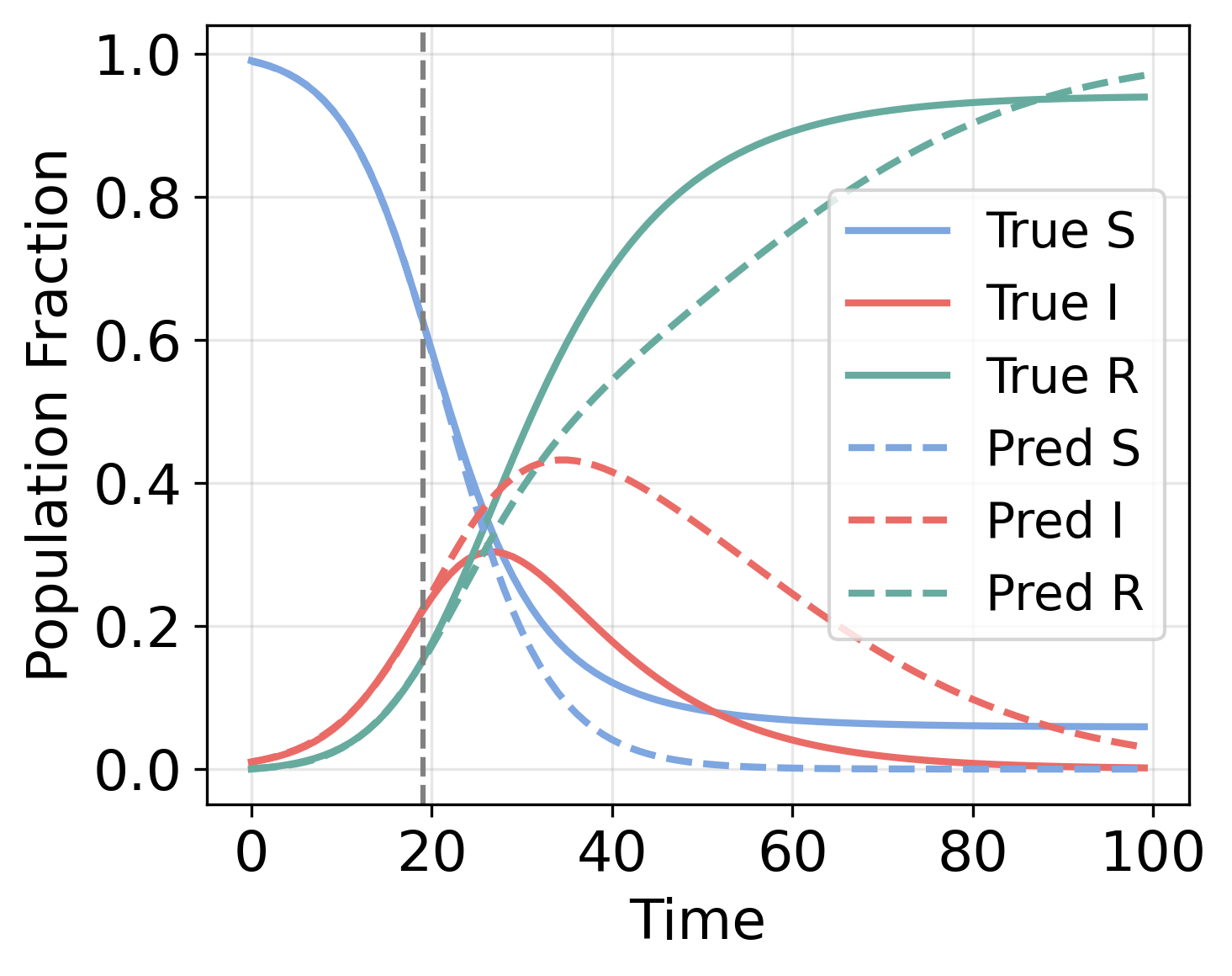}
        \caption{\small Vanilla Neural ODE with all states under SIR} \label{fig:vanilla_node}
    \end{subfigure}\hfill
    \begin{subfigure}[t]{0.49\columnwidth}
        \centering
        \includegraphics[width=\linewidth, height=3cm]{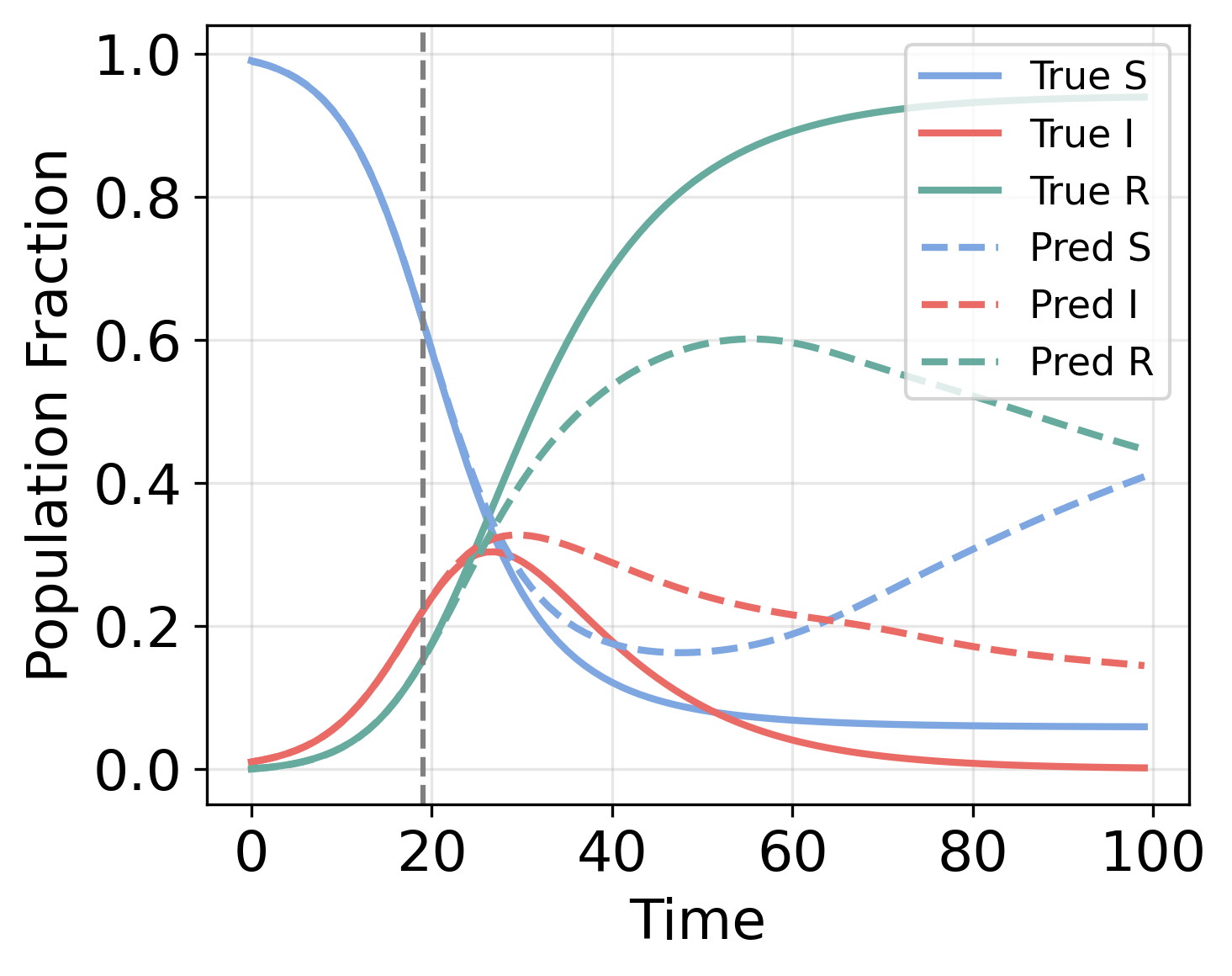}
        \caption{\small AE-NODE with all states under SIR}\label{fig:ae-node}
    \end{subfigure}
    \caption{Failure of Neural ODE architectures under full observability at a train/forecast split of 0.2/0.8.}
    \label{fig:fm_datadriven}
    \vspace{-1mm}
\end{figure}   

We first evaluated vanilla neural ODEs \cite{neural-odes}
and a deterministic autoencoder-based neural ODE (AE-NODE) architecture on synthetic data generated from an SIR compartmental model. In the former, the predicted SIR states evolve according to a vector field parameterized by a neural network. In the latter, an encoder maps the observed time series to an initial latent state,
which is evolved continuously in time via a learned vector field. A decoder then maps the latent trajectory back to observation space. We use a train/forecast split of 0.2/0.8, and impose basic structural constraints on model outputs, namely mass conservation ($S + I + R = 1$) and non-negativity of each compartmental state. 
As shown in Figure~\ref{fig:fm_datadriven}, while these models succeed in generating near exact fits in the training window, they fail to reliably forecast accurate SIR dynamics beyond it.

\subsection{Bidirectional objectives fail to resolve identifiability and learn implausible latent dynamics.}
\begin{figure}[htbp]
    \centering
    \begin{subfigure}[t]{0.49\columnwidth}
        \centering
        \includegraphics[width=\linewidth, height=3cm]{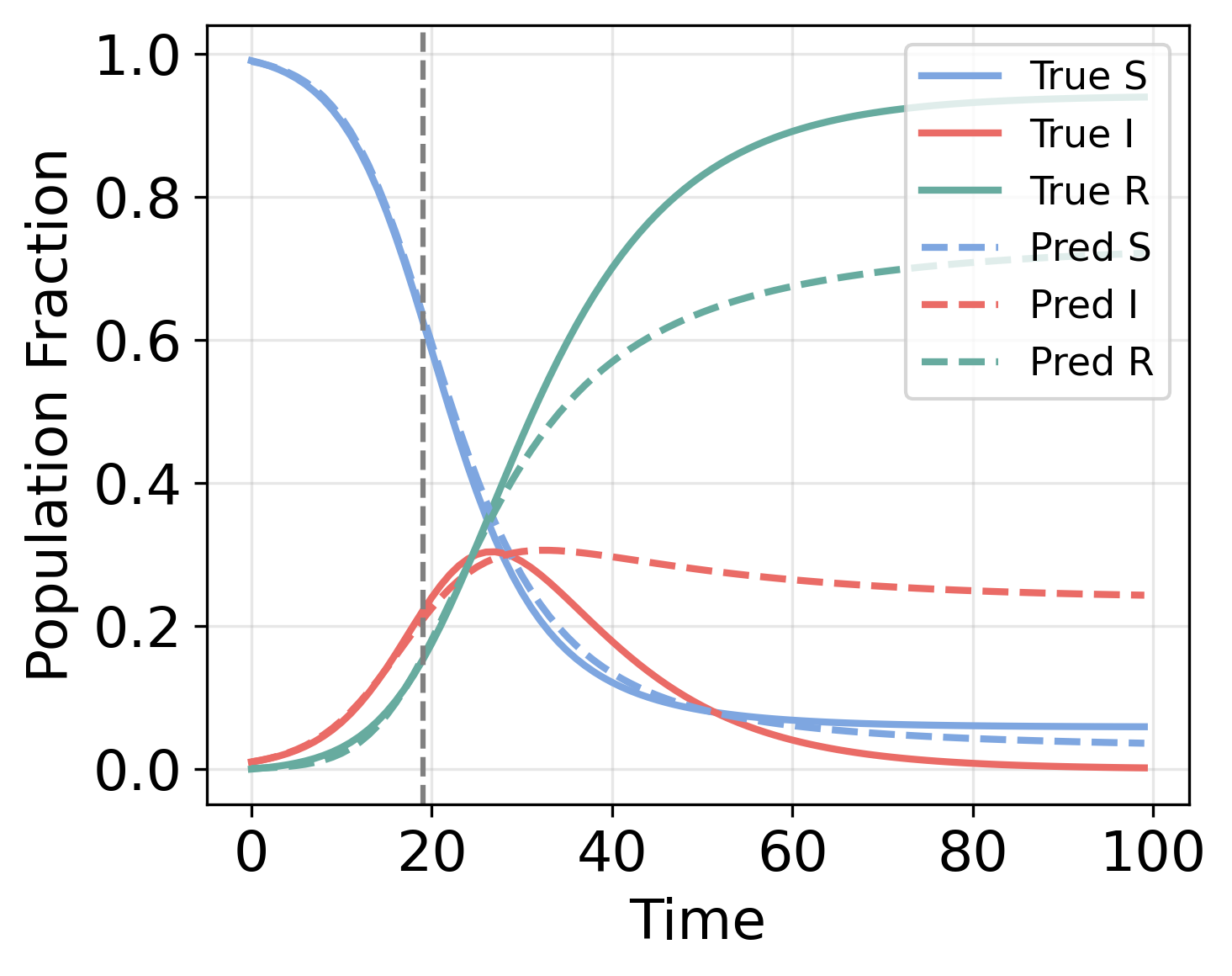}
         \caption{\small Bidirectional AE-NODE with all states under SIR} \label{fig:bidir_full}
    \end{subfigure}\hfill
    \begin{subfigure}[t]{0.49\columnwidth}
        \centering
        \includegraphics[width=\linewidth, height=3cm]{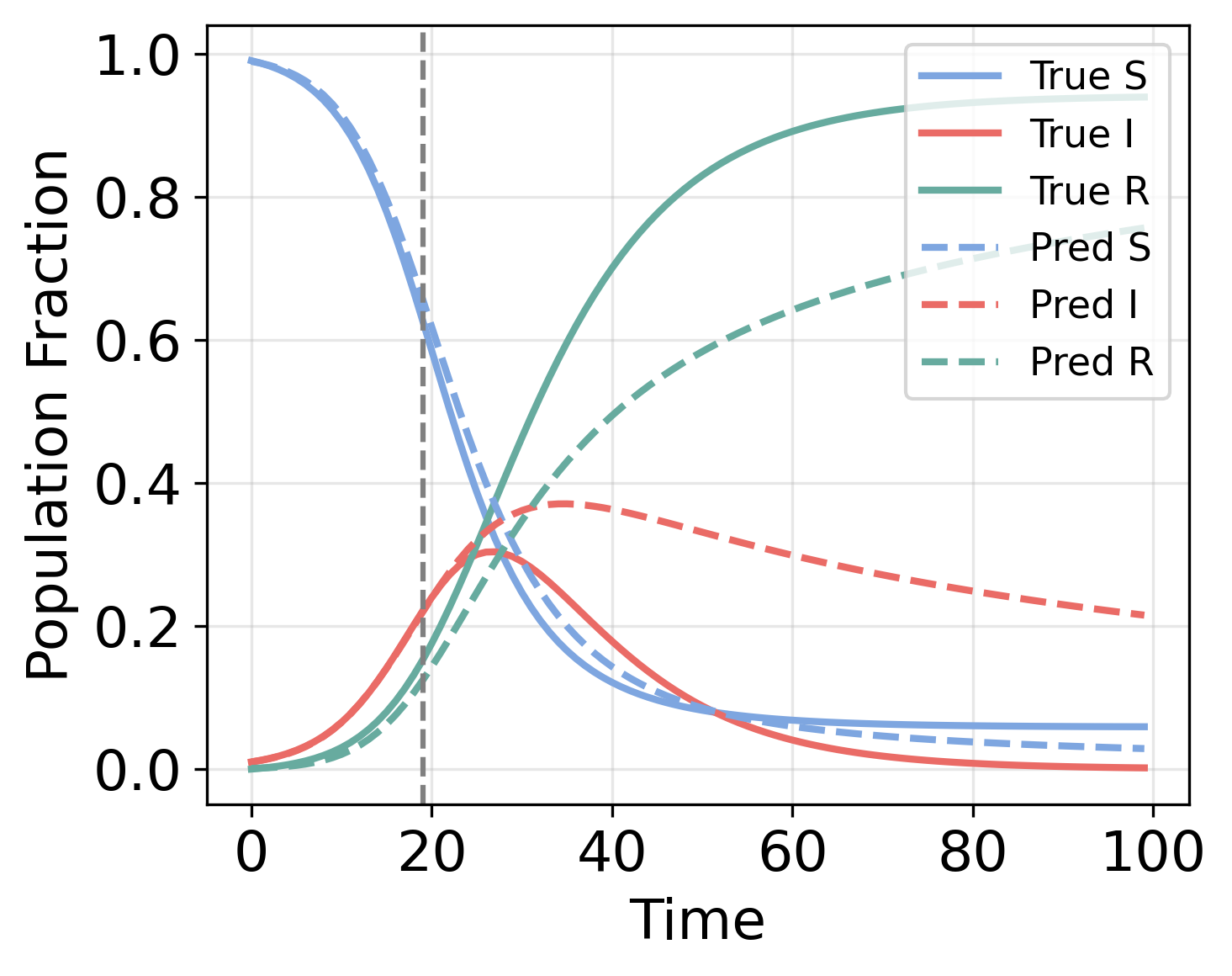}
        \caption{\small Bidirectional AE-NODE with I-only under SIR}\label{fig:bidir-ionly}
    \end{subfigure}
    \caption{
    Failure of SIR forecasts from AE-NODE under short training windows, with bidirectional training.
    }
    \label{fig:fail-koopman}
\end{figure}
\vspace{-1mm}

Inspired by consistent Koopman Autoencoders \cite{koopman}, we explored bidirectional training objectives for the AE-NODE implemented in \ref{sec:fm-datadriven}. In addition to a reconstruction loss over the full training window, we compute local forward and backward losses anchored by sliding origins. For a given time origin $s$, the encoder infers a latent state based on observations up to $t = s$. The latent state is then evolved both forward and backward over five time steps within the training window, and compared with the corresponding observations after decoding. 

While bidirectional training encourages learned latent states to be consistent with nearby observations, it does not yield accurate forecasts 
(Figure~\ref{fig:fail-koopman}). When only the infected trajectory I(t) is observed (Fig.~\ref{fig:bidir-ionly}), the backward reconstruction introduces no new information about unobserved compartments or time-varying drivers. When all compartments are observed (Fig. \ref{fig:bidir_full}), the additional loss terms are still insufficient to constrain the learned dynamics toward the correct epidemic trajectory. Moreover, epidemic processes are intrinsically forward-evolving, governed by causal transmission and recovery mechanisms. This highlights that bidirectional consistency 
must be complemented with structured, forward-driving signals to reliably learn interpretable and stable epidemic dynamics.

\subsection{Physics-informed losses struggle with partially observed dynamics under complex dynamics.}
\label{sec:fm-PINN}

\begin{figure}[htbp]
    \centering
    \begin{subfigure}[t]{0.49\columnwidth}
        \centering
        \includegraphics[width=\linewidth, height=3cm]{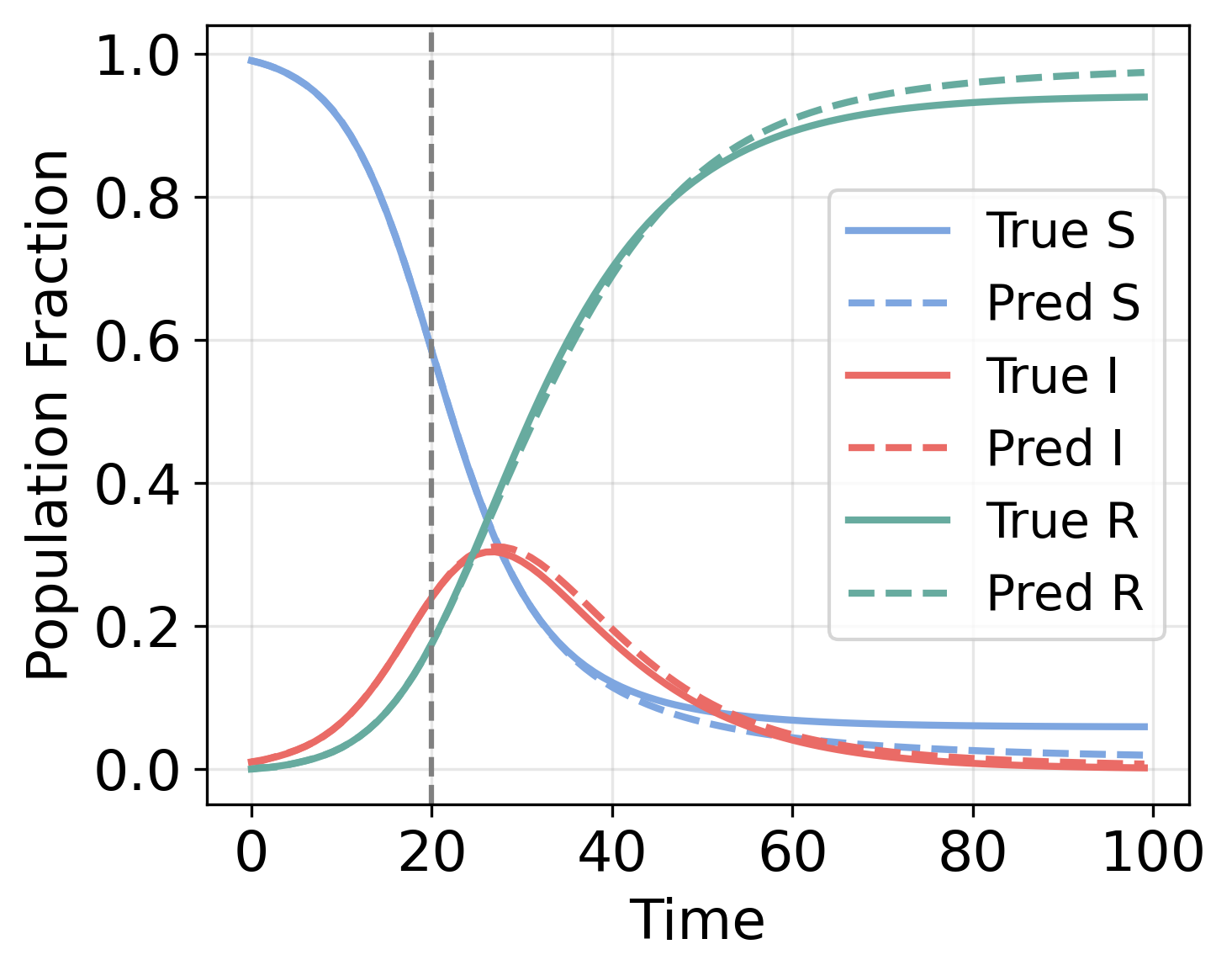}
        \caption{\small Physics-informed loss with all states under SIR} \label{fig:pinn_full_sir}
    \end{subfigure}\hfill
    \begin{subfigure}[t]{0.49\columnwidth}
        \centering
        \includegraphics[width=\linewidth, height=3cm]{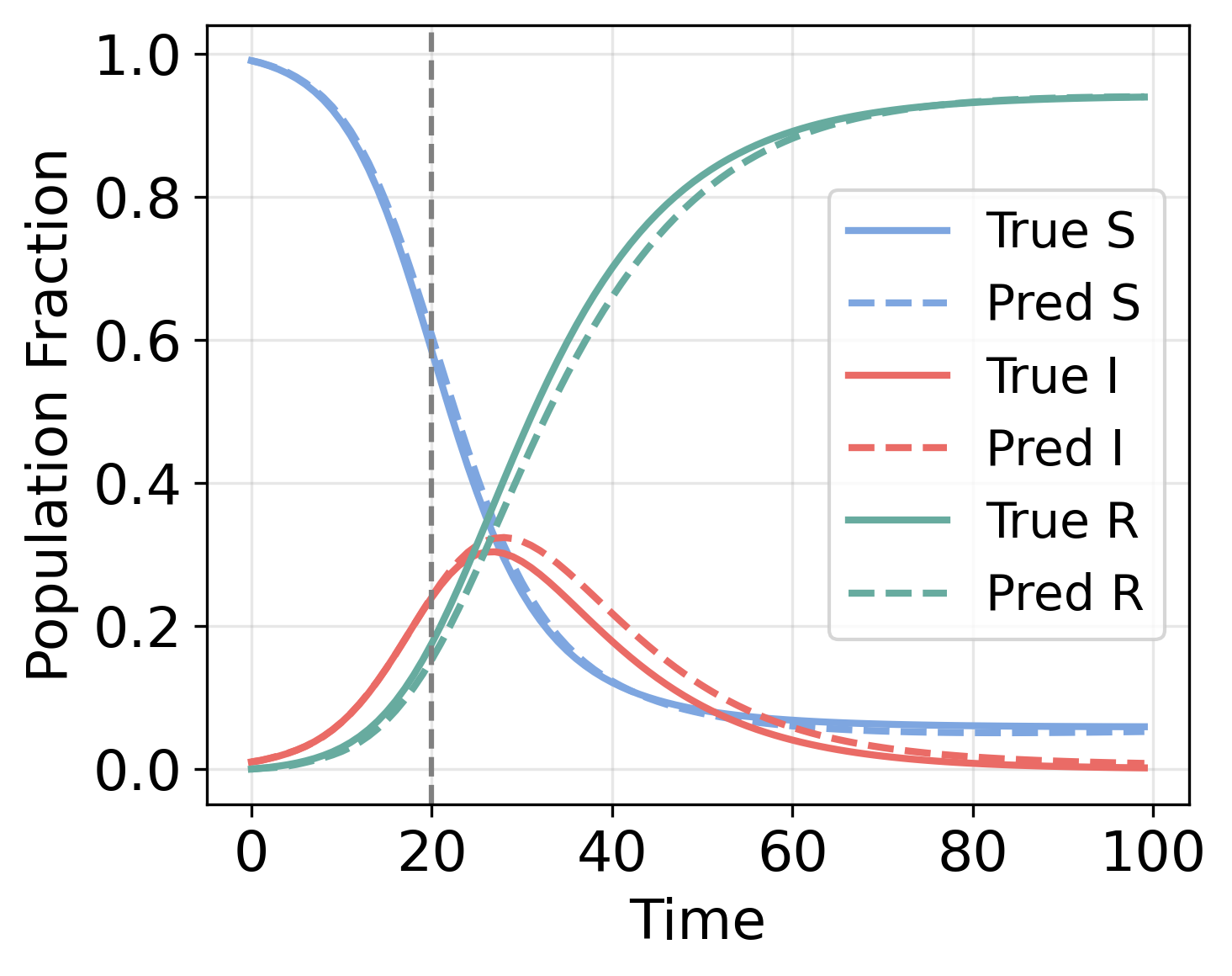}
        \caption{\small Physics-informed loss with I-only under SIR}\label{fig:pinn_i_sir}
    \end{subfigure}
    \caption{Incorporating physics-informed loss with an SIR model assumption. Forecasts are accurate both when all states are observed (a) and when only $I(t)$ is observed (b).}
    \vspace{-1mm}
\end{figure}   

\begin{figure}[htbp]
    \centering
    \begin{subfigure}[t]{0.49\columnwidth}
        \centering
        \includegraphics[width=\linewidth, height=3cm]{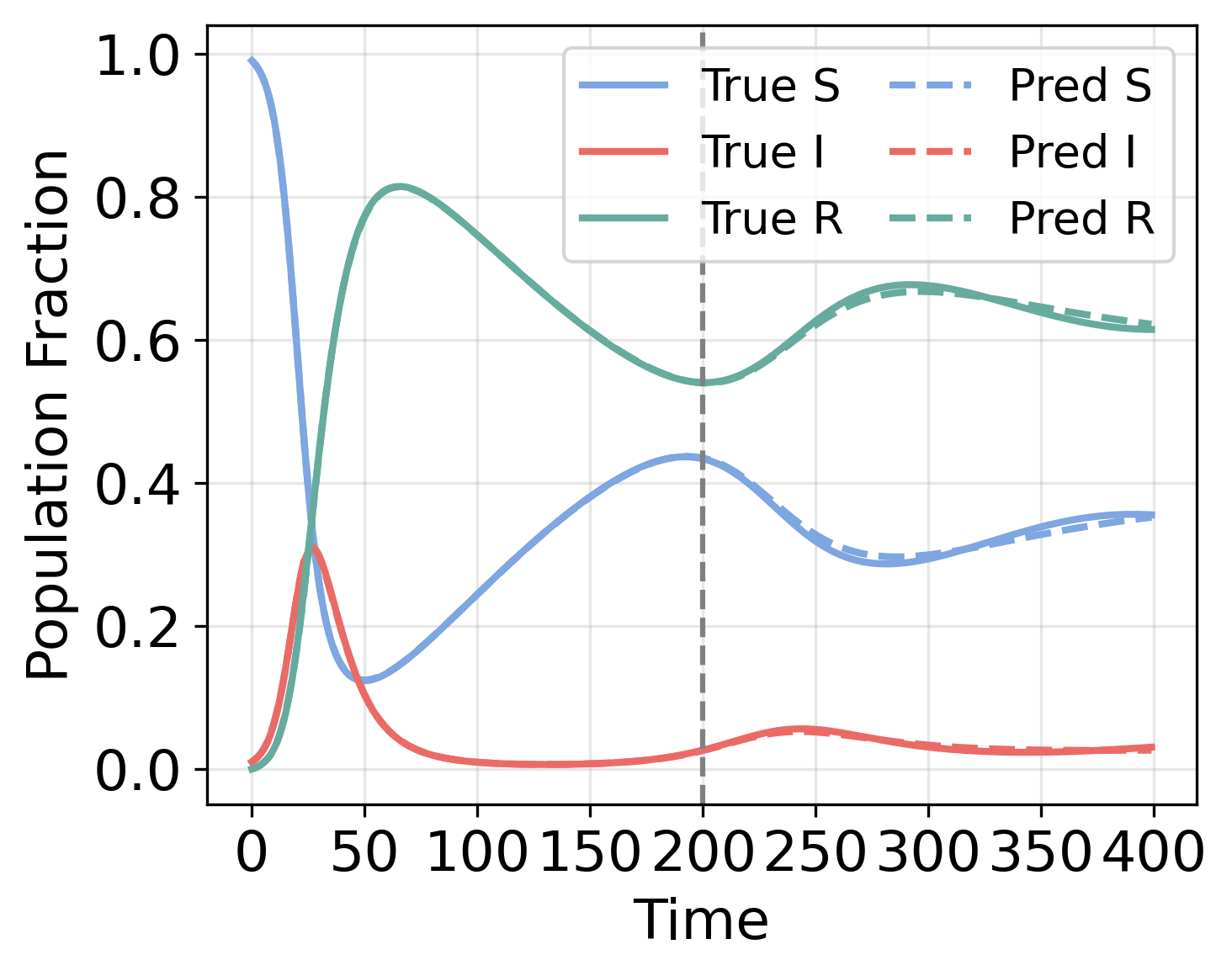}
        \caption{\small Physics-informed loss with all states under SIRS} \label{fig:pinn_full_sirs}
    \end{subfigure}\hfill
    \begin{subfigure}[t]{0.49\columnwidth}
        \centering
        \includegraphics[width=\linewidth, height=3cm]{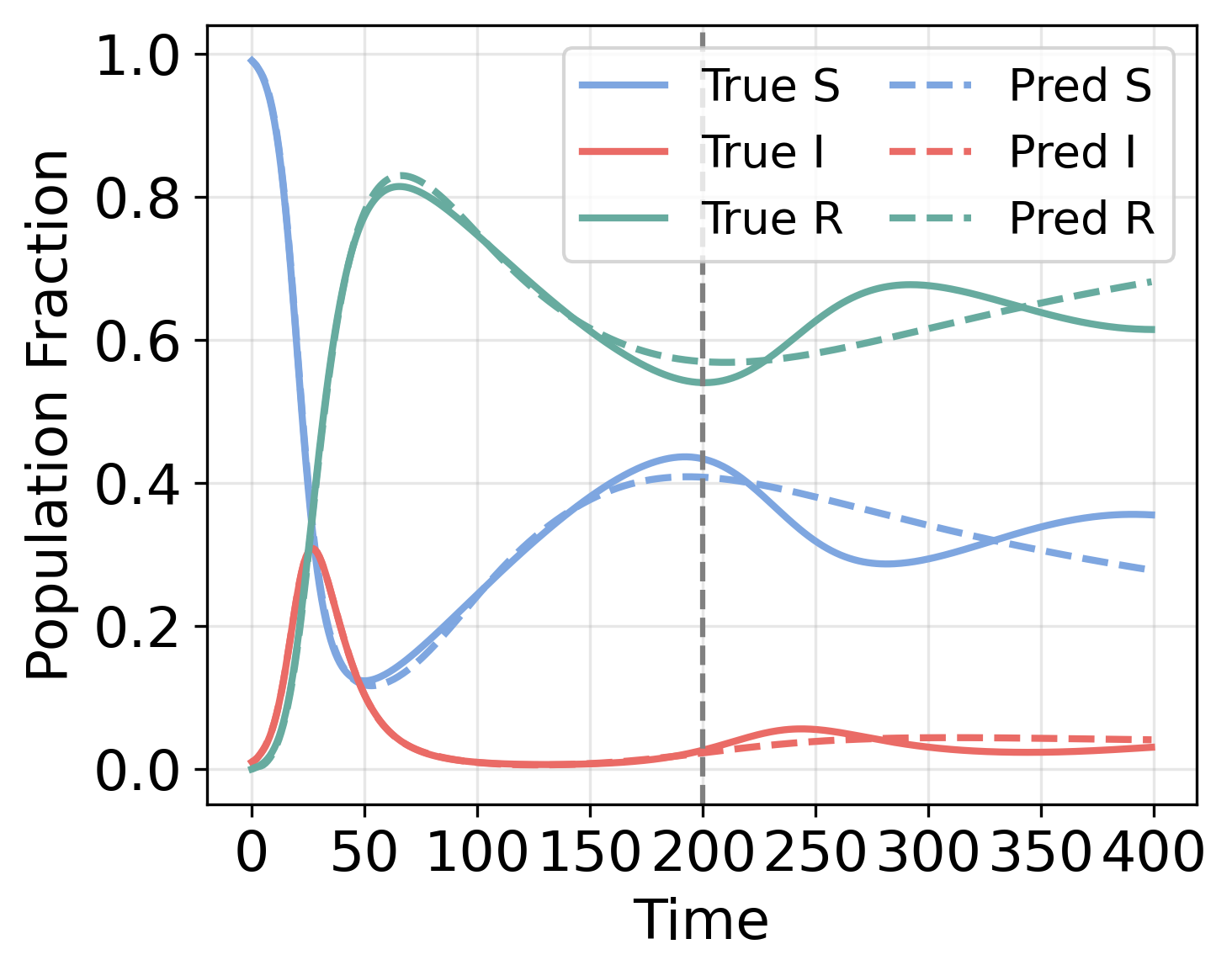}
        \caption{\small Physics-informed loss with I-only under SIRS}\label{fig:pinn_i_sirs}
    \end{subfigure}
    \caption{Incorporating physics-informed loss with an SIRS model assumption. Forecasts trained on synthetic SIRS data are accurate when all compartments are observed (a), but degrade with only $I(t)$ observed (b).}
    \vspace{-2mm}
\end{figure} 



Next, we incorporate physics-informed losses \cite{raissi2019physics} that penalize violations of SIR/SIRS differential equations during training. These formulations aim to
improve physical feasibility and extrapolation when
strong prior knowledge is available. The compartmental states are parameterized via a neural network, optimized jointly with rate constants corresponding to the appropriate governing compartmental system. The loss penalizes both the mean squared error of the learned trajectories within the training window, and the SIR/SIRS residual induced by the governing equations at collocation points along the full time window. 
The models perform well on both simulated SIR and SIRS trajectories when all compartments are observed (Figure \ref{fig:pinn_full_sir}, \ref{fig:pinn_full_sirs}). 
When only $I(t)$ is available, forecasts remain similarly accurate under SIR dynamics (Figure \ref{fig:pinn_i_sir}). However, in the context of an SIRS system, the model produces qualitative errors in unobserved compartments and long-horizon dynamics (Figure \ref{fig:pinn_i_sirs}). This suggests that waning immunity introduces additional uncertainty in the unobserved states and model parameters, leaving the physics-informed loss too weak to recover the underlying system.


\subsection{Mechanistic Neural ODEs fail to capture multi-wave dynamics.}

 \begin{figure}[htbp]
    \centering
    \begin{subfigure}[t]{0.49\columnwidth}
        \centering
        \includegraphics[width=\linewidth, height=3cm]{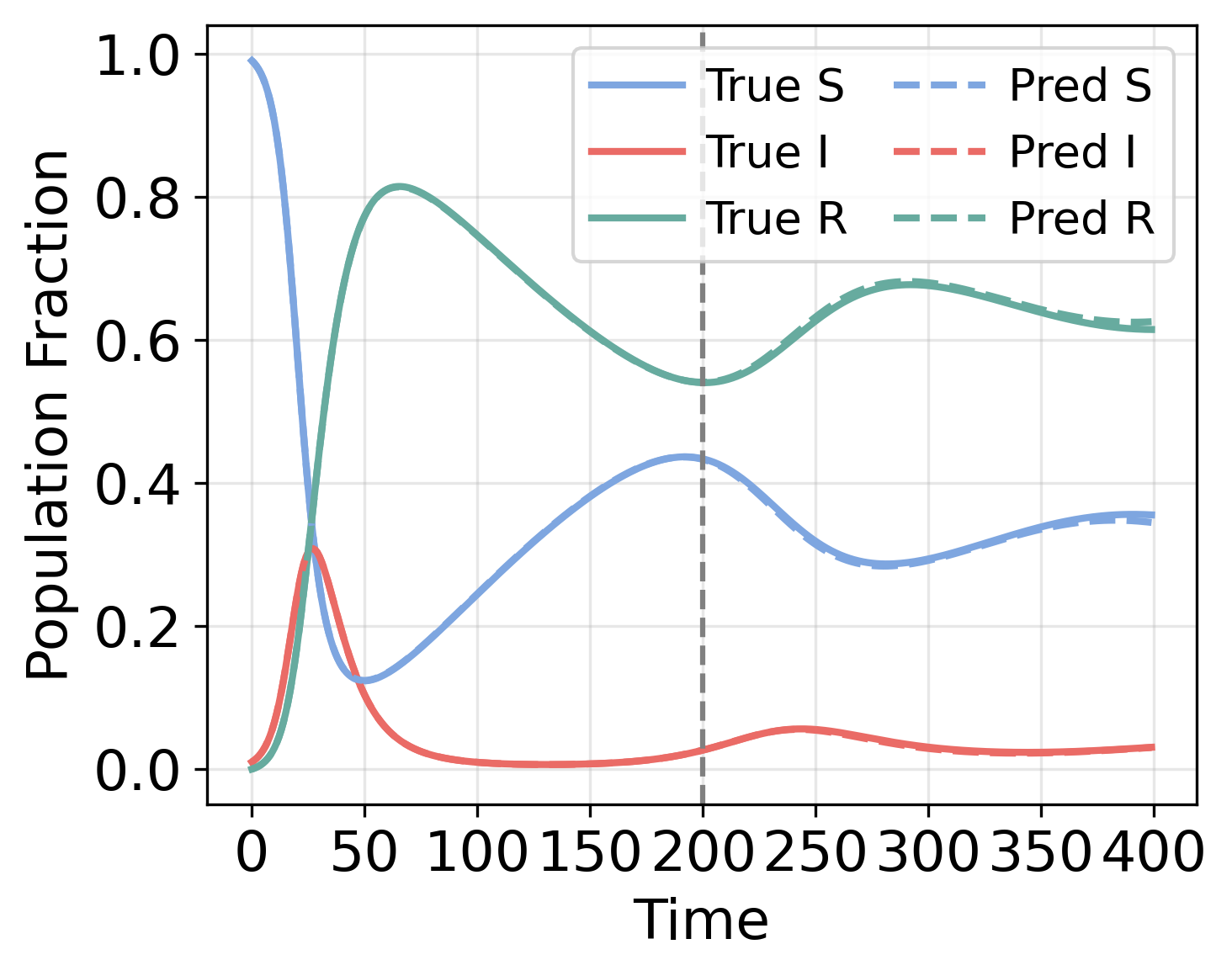}
        \caption{\small Mechanistic neural ODE with I-only under SIRS} \label{fig:mech_node_sirs_i}
    \end{subfigure}\hfill
    \begin{subfigure}[t]{0.49\columnwidth}
        \centering
        \includegraphics[width=\linewidth, height=3cm]{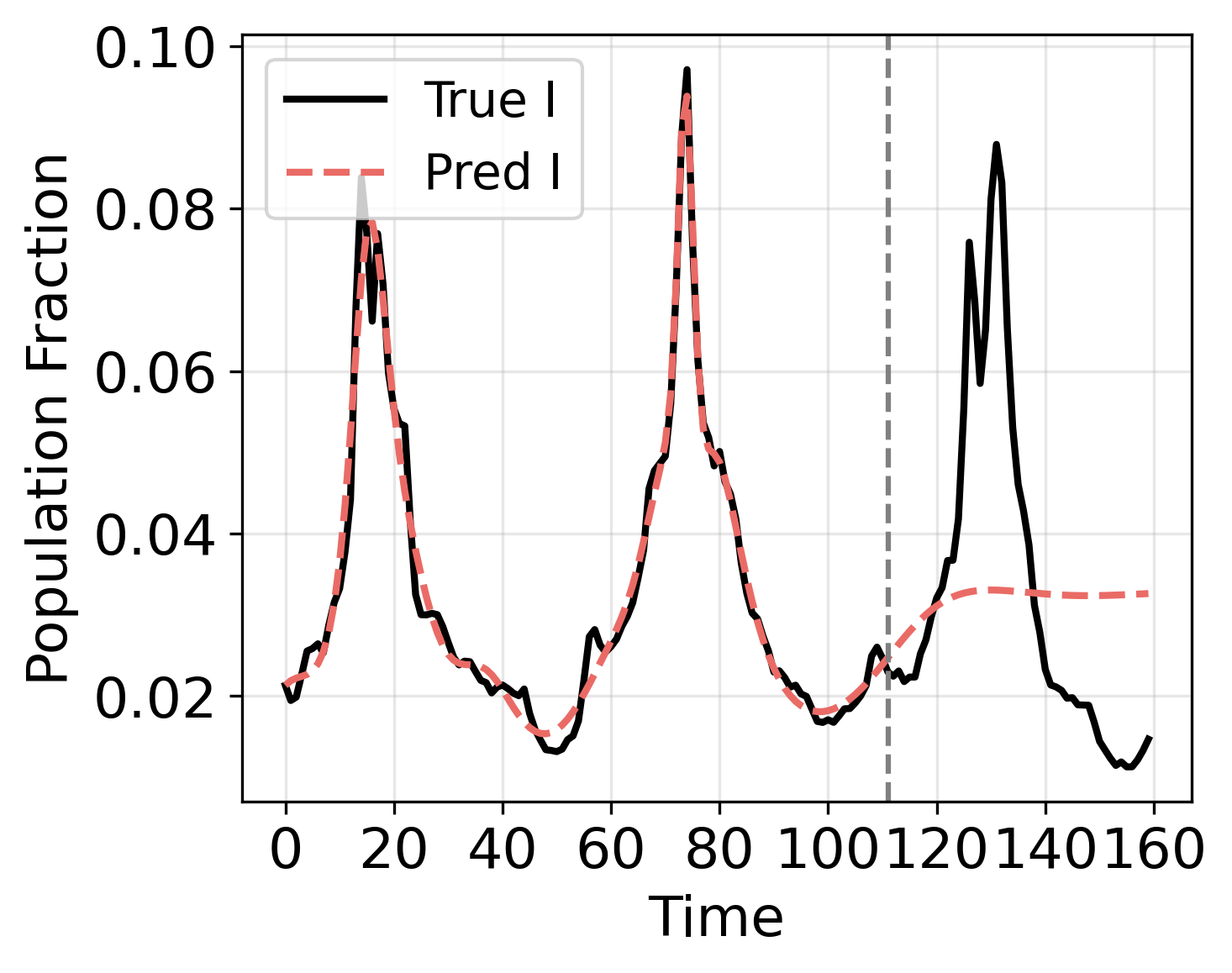}
        \caption{\small Mechanistic neural ODE with I-only under ILI}\label{fig:mech_node_ili_i}
    \end{subfigure}
    \caption{Failure of mechanistic neural ODE forecasts on real data under partial observability. Forecasts perform well for the SIRS model (a), but fail to capture trends for ILI data (b).}
    \vspace{-1mm}
\end{figure}

We then evaluate a hybrid mechanistic neural ODE \cite{rackauckas2021universal}, where the SIRS compartmental model structure is assumed, while the rate parameters (transmission and recovery rates, and immunity waning) are learned via neural networks to generate the prediction. Under synthetic SIRS dynamics, the model reproduces full trajectories with high accuracy at the given train/test split when only $I(t)$ is observed (Figure \ref{fig:mech_node_sirs_i}). However, it struggles with real-world, multi-wave influenza-like illness (ILI) data (Figure \ref{fig:mech_node_ili_i}). This limitation arises because multi-wave dynamics are driven by a variety of latent forcing processes, such as seasonality and immunity waning, that cannot be uniquely inferred from $I(t)$ alone. Consequently, the learned epidemic dynamics generalize poorly beyond the observed training period, resulting in inaccurate long-horizon forecasts across successive epidemic waves.

\section{\model{} Framework}
\label{sec:method}

\begin{figure*}[t]
    \centering
    \includegraphics[width=\linewidth]{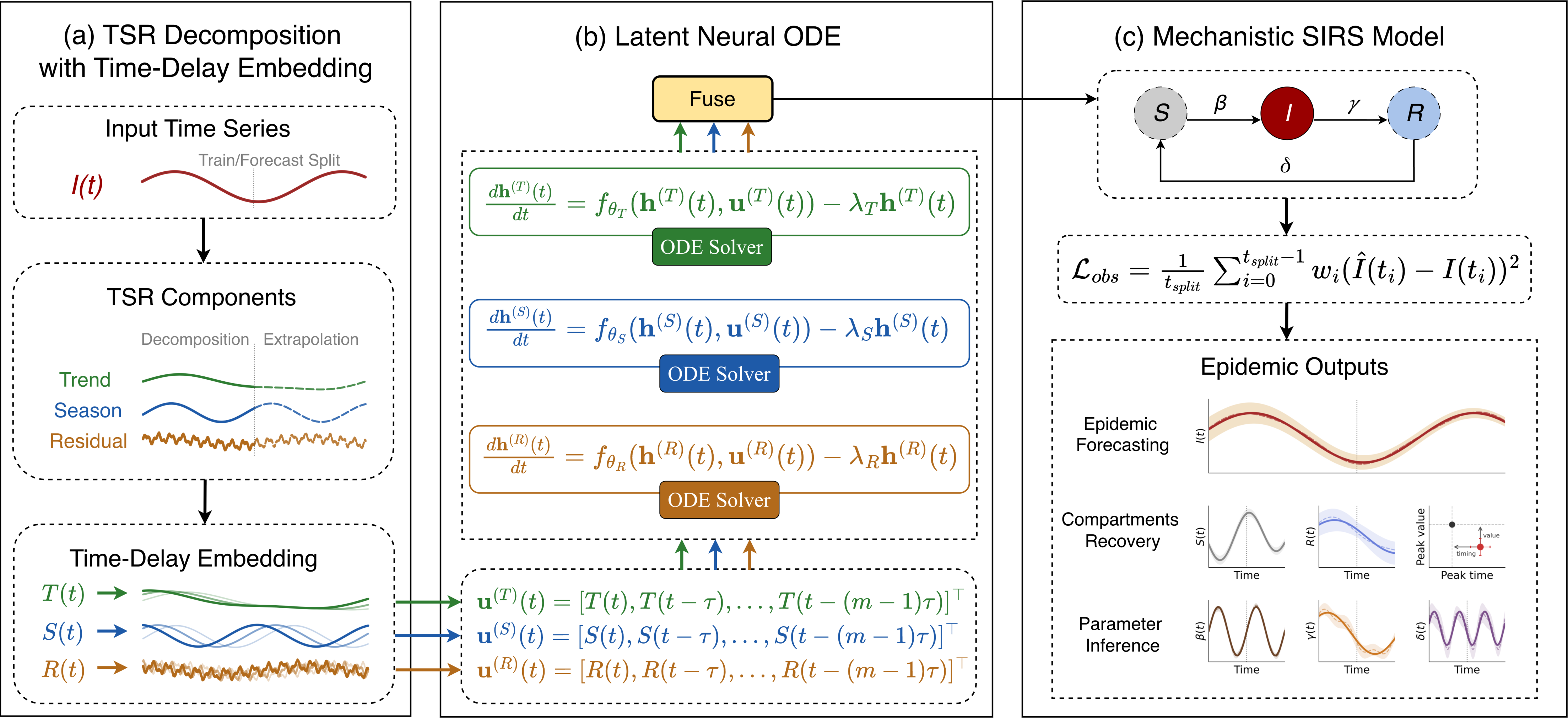}
    \caption{
    Overview of \textbf{\model{}} framework. The observed epidemic time series is decomposed into \emph{trend}, \emph{seasonal}, and \emph{residual} components, which serve as multi-scale control signals for latent Neural ODEs. The fused latent representation is decoded into time-varying epidemiological parameters and coupled with mechanistic SIRS dynamics to produce physically consistent forecasts and interpretable parameter trajectories.
    }
    \label{fig:model}
\end{figure*}


Toward addressing the above issues,
we present \textbf{\model}{} (Figure~\ref{fig:model}), a hybrid neural-physical framework for epidemic forecasting from partial observations. \model{} integrates multi-scale signal decomposition, controlled latent continuous-time dynamics, and mechanistic SIRS evolution to jointly forecast epidemic trajectories and infer time-varying epidemiological parameters. 


\paragraph{Problem setup.}
Let $\{(t_i, I_i)\}_{i=0}^{T-1}$ denote an epidemic time series, where only the infected compartment $I(t)$ is observed at discrete times $t_i$. The latent epidemic state is
\begin{equation}
\mathbf{y}(t) = [S(t), I(t), R(t)]^\top \in \mathbb{R}_{\ge 0}^3,
\label{eq:state}
\end{equation}
with unknown, time-varying parameters $\beta(t)$ (transmission), $\gamma(t)$ (recovery), and $\delta(t)$ (immunity waning).

We assume $I(0)$ is known and initialize
\begin{equation}
S(0) = 1 - I(0), \qquad R(0) = 0.
\label{eq:init}
\end{equation}

Our objective is to forecast $I(t)$ beyond the observation window, accurately predict epidemic peaks, and infer interpretable parameter trajectories.

\paragraph{Variational mode decomposition (VMD).}
\label{sec:vmd}

Given a real-valued epidemic signal $x(t)$, we use 
VMD~\cite{vmd} to decompose it into $K$ intrinsic mode functions $\{u_k(t)\}_{k=1}^K$, each associated with a center frequency $\omega_k$, by minimizing the total bandwidth of the modes subject to exact signal reconstruction. The constrained variational problem is defined as:
\begin{equation}
\begin{aligned}
\min_{\{u_k\},\,\{\omega_k\}} \quad &
\sum_{k=1}^{K}
\left\|
\partial_t \left[
\left(\delta_D(t) + \frac{j}{\pi t}\right) * u_k(t)
\right]
e^{-j\omega_k t}
\right\|_2^2 \\
\text{s.t.}\quad &
\sum_{k=1}^{K} u_k(t) = x(t),
\end{aligned}
\label{eq:vmd-objective}
\end{equation}
where $(\delta_D(t) + \frac{j}{\pi t}) * u_k(t)$ denotes the analytic signal of $u_k(t)$ obtained via the Hilbert transform, $\delta_D$ is the Dirac delta, $\partial_t$ is the temporal derivative, and $*$ denotes convolution. This objective encourages each mode to be compact around its center frequency while collectively reconstructing the original signal.

\paragraph{Augmented Lagrangian formulation. }
The constrained problem in \eqref{eq:vmd-objective} is solved using the alternating direction method of multipliers (ADMM)~\cite{boyd2011distributed} by forming the augmented Lagrangian:

\begin{equation}
\begin{aligned}
\mathcal{L}(\{u_k\},\{\omega_k\},\lambda) \\
= \alpha \sum_{k=1}^{K}
\left\|
\partial_t
\Bigl(
(\delta_D(t)+\tfrac{j}{\pi t}) * u_k(t)
\Bigr)
e^{-j\omega_k t}
\right\|_2^2 \\
+ \left\|
x(t)-\sum_{k=1}^{K}u_k(t)
\right\|_2^2 \\
+ \left\langle
\lambda(t),
x(t)-\sum_{k=1}^{K}u_k(t)
\right\rangle .
\end{aligned}
\label{eq:vmd-lagrangian}
\end{equation}
where $\lambda(t)$ is the Lagrange multiplier and $\alpha>0$ controls the bandwidth penalty.

\paragraph{ADMM updates. }

Let $\hat{x}(\omega)$, $\hat{u}_k(\omega)$, and $\hat{\lambda}(\omega)$ denote the Fourier transforms of $x(t)$, $u_k(t)$, and $\lambda(t)$, respectively. The ADMM updates admit closed-form solutions in the frequency domain. At iteration $n+1$, the mode update is
\begin{equation}
\hat{u}_k^{\,n+1}(\omega)
=
\frac{
\hat{x}(\omega)
-
\sum_{i\neq k} \hat{u}_i^{\,n}(\omega)
+
\frac{1}{2}\hat{\lambda}^{\,n}(\omega)
}{
1 + 2\alpha\left(\omega - \omega_k^{\,n}\right)^2
},
\label{eq:vmd-uk}
\end{equation}
and the center frequency is updated as the energy-weighted mean frequency:
\begin{equation}
\omega_k^{\,n+1}
=
\frac{
\int_{0}^{\infty} \omega \left|\hat{u}_k^{\,n+1}(\omega)\right|^2\, d\omega
}{
\int_{0}^{\infty} \left|\hat{u}_k^{\,n+1}(\omega)\right|^2\, d\omega
}.
\label{eq:vmd-omega}
\end{equation}
The Lagrange multiplier is updated by
\begin{equation}
\hat{\lambda}^{\,n+1}(\omega)
=
\hat{\lambda}^{\,n}(\omega)
+
\tau\left(
\hat{x}(\omega) - \sum_{k=1}^{K} \hat{u}_k^{\,n+1}(\omega)
\right),
\label{eq:vmd-lambda}
\end{equation}
where $\tau>0$ is the dual ascent step size. Iterations continue until convergence.
VMD produces an ensemble of band-limited modes with distinct frequency characteristics. In our framework, these modes are represented by three frequency levels: (i) trend $T(t)$ aims to capture slow shifts such as those from accumulated immunity, behavioral drift, policy interventions, and pathogen evolution, dominating long-horizon accuracy as trend errors compound over time;
(ii) seasonal $S(t)$ captures periodic forcing from climate, school terms, and indoor contact patterns at annual or semi-annual scales, critical for peak-timing prediction; and (iii) 
residual $R(t)$ absorbs high-frequency reporting artifacts, super-spreader fluctuations, and stochastic dynamics, keeping the trend and seasonal estimates clean and stable.
The decomposed modes are used as structured forcing signals for downstream epidemic modeling:
\begin{equation}
I(t) = T(t) + S(t) + R(t),
\label{eq:tsr}
\end{equation}
where $T(t)$ captures low-frequency trends, $S(t)$ captures periodic or seasonal structure, and $R(t)$ represents residual fluctuations. This decomposition isolates distinct temporal scales that would otherwise be confounded in a single observation stream.

\paragraph{Channel-aware extrapolation past the observation window.}
Equation~\eqref{eq:tsr} holds only where $I(t)$ is observed, i.e.\ on $[0, t_{\mathrm{split}})$. To make the channels available to downstream components on the full window, we extend each channel into $[t_{\mathrm{split}}, T)$ by a rule matched to its spectral character \cite{harvey1989structural,cleveland1990stl}. The trend is extrapolated as a linear drift fit by least squares on the last $L_{\mathrm{fit}}$ training samples \cite{zeng2023dlinear}:
\begin{equation}
T(t) \;=\; a\, t + b,
\label{eq:extrap-trend}
\end{equation}
where $t \ge t_{\mathrm{split}}$ and $(a, b)$ minimize $\sum_{j=t_{\mathrm{split}}-L_{\mathrm{fit}}}^{t_{\mathrm{split}}-1}\bigl(T(t_j) - a\,t_j - b\bigr)^2$.
The seasonal channel is tiled at its dominant period $p_S$, obtained from a periodogram of $\{S(t_j)\}_{j=0}^{t_{\mathrm{split}}-1}$ \cite{bloomfield2000fourier}, yielding a seasonal-naive forecaster at the data-driven period \cite{hyndman2021forecasting}:
\begin{equation}
S(t) \;=\; S\!\bigl(t - p_S\, k(t)\bigr), \quad k(t) = \bigl\lceil (t - t_{\mathrm{split}} + 1)/p_S \bigr\rceil,
\label{eq:extrap-season}
\end{equation}
where the RHS argument lies in $[t_{\mathrm{split}} - p_S,\, t_{\mathrm{split}})$ and is therefore defined by the causal VMD output. 
Because the residual is a high-frequency signal with no extrapolable structure, we set it to its training-window mean:
\begin{equation}
R(t) \;=\; \bar{R}_{\mathrm{train}}.
\label{eq:extrap-resid}
\end{equation}
By VMD’s bandwidth regularization, modes with $\omega_k>0$ have approximately zero mean.


\paragraph{Time-delay embedding of control signals.}
To provide temporal context, we apply a time-delay embedding to each component. For a component $c(t)\in\{T(t), S(t), R(t)\}$, corresponding to trend, seasonality, and residual respectively, we construct the lag-augmented control:
\begin{equation}
\mathbf{u}^{(c)}(t) =
\big[c(t), c(t-\tau), \dots, c(t-(m-1)\tau)\big]^\top \in \mathbb{R}^m,
\label{eq:delay}
\end{equation}

where $\tau$ is the delay and $m$ is the embedding dimension.
Each delay-embedded control $\mathbf{u}^{(c)}(t)$ is then provided to its own latent Neural ODE, allowing the model to capture component-specific temporal dependencies before fusion. 

\paragraph{Collaborative latent neural ODEs.}
We maintain three latent states corresponding to the TSR components:
\begin{equation}
\mathbf{h}^{(T)}(t)\in\mathbb{R}^{d_T}, \quad
\mathbf{h}^{(S)}(t)\in\mathbb{R}^{d_S}, \quad
\mathbf{h}^{(R)}(t)\in\mathbb{R}^{d_R}.
\label{eq:latents}
\end{equation}

Each latent state evolves according to a Neural ODE:
\begin{equation}
\frac{d\mathbf{h}^{(c)}(t)}{dt}
=
f_{\theta_c}\big(\mathbf{h}^{(c)}(t), \mathbf{u}^{(c)}(t)\big)
-
\lambda_c\, \mathbf{h}^{(c)}(t), 
\label{eq:controlled-ode}
\end{equation}
where $f_{\theta_c}$ is a neural vector field, $c\in\{T,S,R\}$, $\mathbf{u}^{(c)}(t)$ is the TSR-based control input from \eqref{eq:delay}, and $\lambda_c\ge 0$ is a learnable damping coefficient implemented via $\mathrm{softplus}(\cdot)$. The ODEs are numerically integrated between observation times using an explicit solver (RK4 in our experiments).

\paragraph{Latent fusion and parameter decoding.}

The latent states are fused into a shared representation:
\begin{equation}
\mathbf{h}(t) =
\mathrm{Fuse}\!\left(
[\mathbf{h}^{(T)}(t);\mathbf{h}^{(S)}(t);\mathbf{h}^{(R)}(t)]
\right),
\label{eq:fuse}
\end{equation}
where $\mathrm{Fuse}(\cdot)$ is a multilayer perceptron.
From $\mathbf{h}(t)$, we decode epidemiological parameters using a bounded parameter network:
\begin{equation}
(\tilde{\beta}(t), \tilde{\gamma}(t), \tilde{\delta}(t)) = g_\phi(\mathbf{h}(t)),
\qquad \tilde{\cdot}\in(0,1),
\label{eq:param-raw}
\end{equation}
followed by affine scaling into disease-specific ranges:
\begin{equation}
\begin{aligned}
\beta(t) &= \beta_{\min} + (\beta_{\max}-\beta_{\min})\tilde{\beta}(t), \\
\gamma(t) &= \gamma_{\min} + (\gamma_{\max}-\gamma_{\min})\tilde{\gamma}(t),\\
\delta(t) &= \delta_{\min} + (\delta_{\max}-\delta_{\min})\tilde{\delta}(t). 
\end{aligned}
\label{eq:param}
\end{equation}

\paragraph{Evolving epidemic trajectories.}
Given the decoded parameters, the epidemic state evolves according to the SIRS equations:
\begin{equation}
\begin{aligned}
\dot{S}(t) &= -\beta(t)\, S(t)\, I(t) + \delta(t)\, R(t), \\
\dot{I}(t) &= \beta(t)\, S(t)\, I(t) - \gamma(t)\, I(t), \\
\dot{R}(t) &= \gamma(t)\, I(t) - \delta(t)\, R(t).
\end{aligned}
\label{eq:sirs}
\end{equation}

We integrate \eqref{eq:sirs} using an RK4 step with $\Delta t_i = t_{i+1}-t_i$. After each step, we enforce non-negativity and approximate mass conservation:
\begin{equation}
\mathbf{y}(t_{i+1}) \leftarrow
\alpha \frac{\mathbf{y}(t_{i+1})}{\|\mathbf{y}(t_{i+1})\|_1}
+ (1-\alpha)\mathbf{y}(t_i),
\label{eq:mass}
\end{equation}
with $\alpha=0.9$ in all experiments. The predicted infection trajectory is $\hat{I}(t) = (\mathbf{y}(t))_2$.

\paragraph{Training objective.}

Let $t_{\mathrm{split}}$ denote the end of the observation window. We train \model{} end-to-end by minimizing a weighted mean squared error on the infected compartment:
\begin{equation}
\mathcal{L}_{\mathrm{obs}}
=
\frac{1}{t_{\mathrm{split}}}
\sum_{i=0}^{t_{\mathrm{split}}-1}
w_i \big(\hat{I}(t_i) - I(t_i)\big)^2,
\label{eq:loss}
\end{equation}
where $w_i$ increases linearly near the end of the training window to emphasize alignment at the forecast boundary. Gradients are backpropagated through all components, including the ODE solvers.

\section{Experimental Results}
Our experiments are intended to answer the
below questions:

\begin{enumerate}[label=(\arabic*)]

    \item  \textbf{Performance evaluation (Section~\ref{sec:performance}, Section~\ref{sec:evaluation_appendix})} 
        \begin{enumerate}
            \item \textbf{Forecasting accuracy.} How accurately does \model{} predict the infection trajectory over time? (Section~\ref{sec:forecast_accuracy}, Appendix~\ref{sec:app_forecast_accuracy})
            
            \item \textbf{Peak errors.} How accurately does \model{} predict the magnitude and timing of the infection peak? (Section~\ref{sec:peak}, 
            Appendix~\ref{sec:peak_results}) 

        \end{enumerate}

    \item \textbf{Applications (Section~\ref{sec:application}, Appendix~\ref{sec:application_appendix})}
    
    \vspace{-1mm}
        \begin{enumerate}

        \item \textbf{Parameter inference.} Is \model{} able to recover meaningful and smooth trajectories of time-dependent epidemiological parameters? (Section~\ref{sec:parameter-inference} Appendix~\ref{sec:inference_appendix}) 

        \item \textbf{Regional dynamics.} Is \model{} able to capture regional dynamics of infectious disease transmission? 
        (Appendix~\ref{sec:hhs}) 
        \end{enumerate}
    \item \textbf{Ablation analysis (Appendix~\ref{sec:ablation})}
        \begin{enumerate}
            \item  \textbf{Decomposition $vs$ non-decomposition. } Does decomposing the observed infection time series improve forecasting performance compared to using the raw signal directly? (Appendix~\ref{sec:ablation_arc}) 
    
            \item  \textbf{Effect of decomposition order. } How does the number of decomposed components affect performance? (Appendix~\ref{sec:ablation_noComp}) 
            
            \item  \textbf{Contribution of time delay. } How does time delay embedding influence prediction accuracy and learned dynamics? (Appendix~\ref{sec:ablation_delay_results})
            
            \item  \textbf{Decomposition methods. } How do different decomposition choices affect model accuracy? (Appendix~\ref{sec:ablation_decomp_meth})
            
        \end{enumerate}
\end{enumerate}
\vspace{-2mm}


\subsection{Ablation Procedures}
    \vspace{-1mm}
\paragraph{Single latent ODE.} We replace the three collaborative Neural ODEs in \eqref{eq:controlled-ode} with a single latent state $\mathbf{z}(t)$ as
\begin{equation}
\frac{d\mathbf{z}(t)}{dt}
=
f_\theta\big(\mathbf{z}(t),
[\mathbf{u}^{(T)}(t);\mathbf{u}^{(S)}(t);\mathbf{u}^{(R)}(t)]\big)
-
\lambda \mathbf{z}(t),
\label{eq:tsr1ode}
\end{equation}
followed by the same fusion, parameter decoding, and SIRS rollout. This variant isolates the benefit of disentangling multi-scale dynamics into separate latent flows.

\vspace{-1mm}
\paragraph{Number of decomposed signals.} To study the impact of the number of decomposition components, we compare the accuracy over 1-component (1C) vs 2C vs 3C. In the case of 1C, the input time-series data will not be decomposed and thus this reverts to a vanilla neural ODE. 

\vspace{-1mm}
\paragraph{Choice of decomposition methods.} We further examine which decomposition methods are most effective at producing band-limited components with well-separated frequency content. The decomposition techniques evaluated include MA (moving average) \cite{zhang2022detrendattendrethinkingattention}, STL (Seasonal–Trend decomposition using Loess) \cite{cleveland1990stl}, VMD, Wavelet \cite{mallat1989theory}, SSA–VMD \cite{GAO2023335}, and Neural Koopman–based approaches \cite{NEURAL-KOOPMAN}, covering a broad range of design philosophies (see details in Appendix~\ref{sec:decomposition-details}). 

\vspace{-1mm}
\paragraph{Time delay.} 
When delay embedding is disabled, the control reduces to the instantaneous signal, i.e., $\mathbf{u}^{(c)}(t)=c(t)$.

\subsection{Benchmarking Models}
We compare against a diverse set of baselines, including 
(1)   classical statistical and nonlinear sequence baselines (ARIMA, RNN-based models (LSTM)), 
(2) state-of-the-art univariate forecasters that leverage decomposition or multiscale mixing to capture long-range temporal patterns (TimeKAN \cite{timekan} and TimeMixer++ \cite{timemixer++}), 
(3) a physics-informed neural model (EINN \cite{einn}), 
(4) ODE-based continuous-time models (Neural ODE \cite{neural-odes}, Latent ODE \cite{latent-odes}, and KAN-ODE \cite{kan-odes}), and 
(5) a graph-based neural ODE model (EARTH \cite{pmlr-v267-wan25b}) (see details in Appendix~\ref{benchmark-details}).
\textcolor{black}{Specifically, as EINN employs SEIRm physics, we adapt the framework in two ways: a) replace SEIRm with SIRS, b) predict I rather than m. In both options, I is the only observed compartment. 
For EARTH, 
since we are focused on single region forecasting, we treat the single region as a graph with $N = 1$ node and set adjacency $A = [1]$.}
All models are evaluated under a \emph{single-variate input} setting.

    \vspace{-1mm}

\subsection{Datasets}

\subsubsection{Synthetic datasets}
\paragraph{SIRS with time-fixed and time-varying parameters. }
We generate synthetic epidemics using the SIRS model under multiple parameter regimes: (1) in the fixed setting (SIRS (Fixed)), transmission ($\beta$), recovery ($\gamma$), and immunity-loss ($\delta$) rates remain constant over time, serving as a baseline for identifiability under stationarity; in the time-varying setting (SIRS (Varying)), parameters evolve periodically to emulate seasonal forcing, capturing recurring epidemic patterns.

\vspace{-3mm}
\paragraph{Mismatched epidemic physics. }
To assess robustness to physics mismatch, we simulate data from alternative compartmental models: (1) the SIR setting removes immunity waning, testing the model’s ability to adapt when the assumed SIRS structure is over-parameterized; (2) the SEIRS setting introduces an exposed compartment, increasing latent-state complexity and evaluating performance when the true dynamics deviate from the assumed model class.

\subsubsection{Real-world datasets}
We use weekly Influenza-like illness (ILI) surveillance data collected by the U.S. Centers for Disease Control and Prevention (CDC) (available at \url{https://gis.cdc.gov/grasp/fluview/fluportaldashboard.html}) from all 10 U.S. Department of Health and Human Services (HHS) regions 
(Week 30, 2022–Week 30, 2025). 

\subsection{Evaluation Protocol}
We evaluate all models under a unified protocol: 
\textbf{(1) Forecast accuracy:} evaluate forecasting performance using standard pointwise error metrics on the infection trajectory. 
\textbf{(2) Peak detection accuracy:} 
evaluate peak detection performance by comparing the {predicted and true peak values} (magnitude) and the {predicted and true peak times} (timing). 
For multi-wave sequences, we focus on the dominant peak within the forecasting window.
\textbf{(3) Parameter estimation:} for models that infer epidemiological parameters, evaluate the quality of the {estimated parameter trajectories}. 
\vspace{-5mm}

\paragraph{Evaluation metrics.}
For forecast accuracy we report root-mean-square error (RMSE) on $I(t)$ over the entire forecast window. For peak detection we report signed and absolute errors for both peak timing and peak magnitude. For parameter inference across synthetic datasets, where ground-truth parameters are available, we assess recovery quality using metrics tailored to the underlying parameter dynamics. For SIRS (Varying), where rates evolve over time, we evaluate the inferred parameter dynamics using the sign agreement rate, defined as the fraction of timesteps where the inferred direction of parameter change matches the ground truth. For fixed-parameter datasets, we compare the inferred parameter trajectories against the true values to assess convergence and stability. All metrics are reported as mean $\pm$ std over random seeds to ensure statistical robustness.

\subsection{Performance Evaluation}
\label{sec:performance}

\subsubsection{Forecasting accuracy}
\label{sec:forecast_accuracy}

\begin{figure*}[t]
\centering
\begin{subfigure}{0.32\textwidth}\centering
    \includegraphics[width=\textwidth]{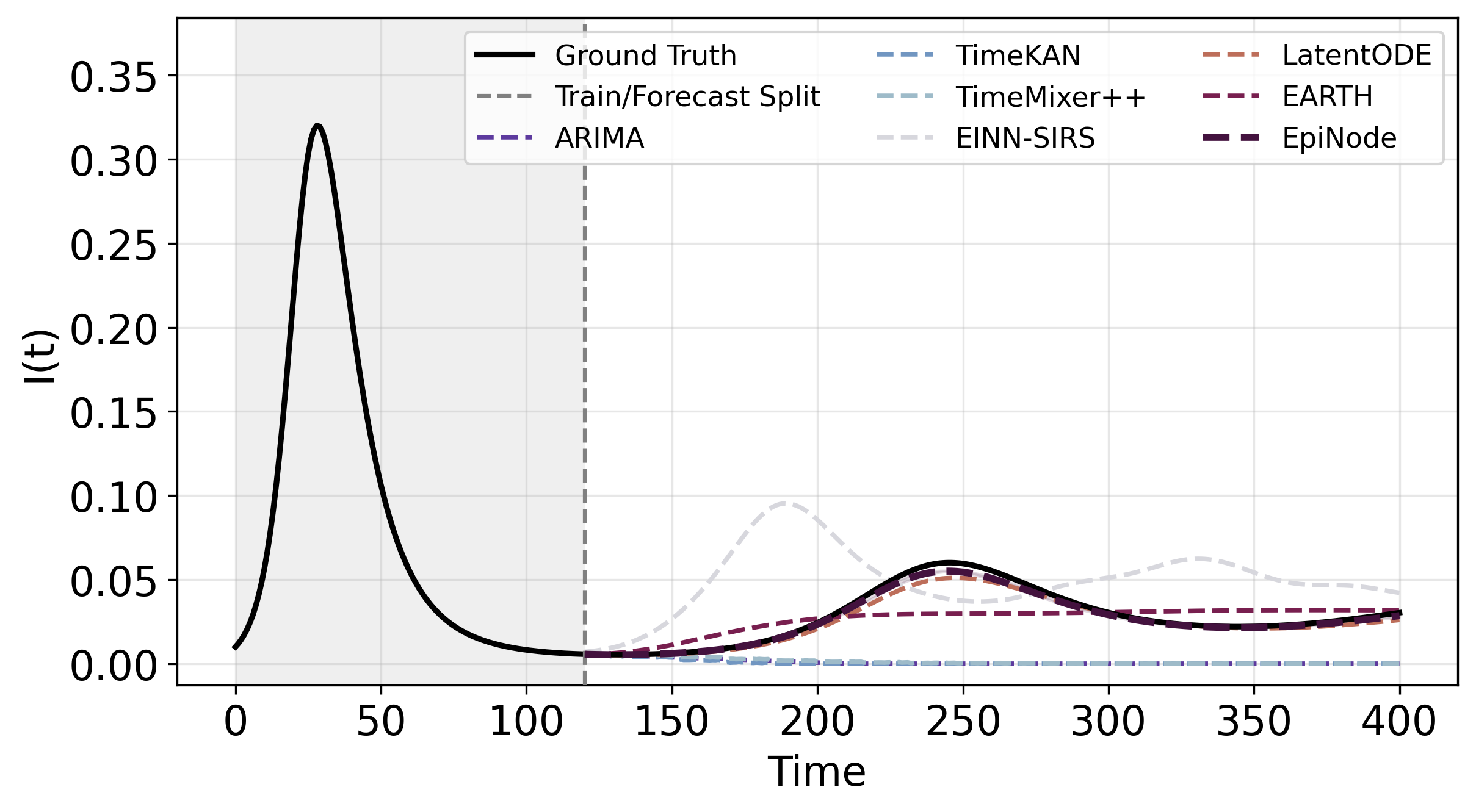}
    \caption{SIRS (Fixed) at split=0.3}
    \label{fig:forecasts_sirs-fixed}
\end{subfigure}\hfill%
\begin{subfigure}{0.32\textwidth}\centering
    \includegraphics[width=\textwidth]{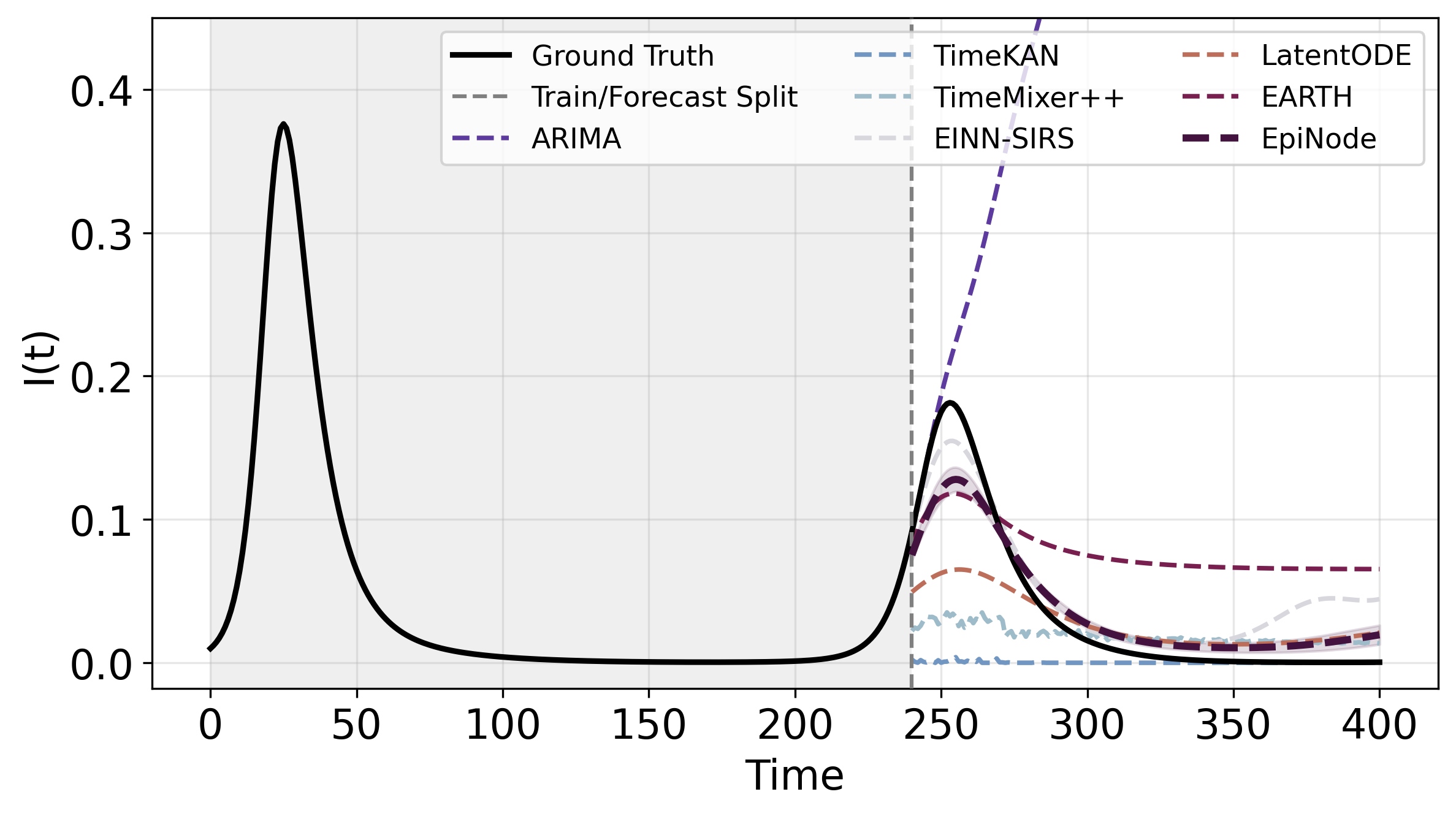}
    \caption{SIRS (Varying) at split=0.6}
    \label{fig:forecasts_sirs-varying}
\end{subfigure} \hfill%
\begin{subfigure}{0.32\textwidth}\centering
    \includegraphics[width=\textwidth]{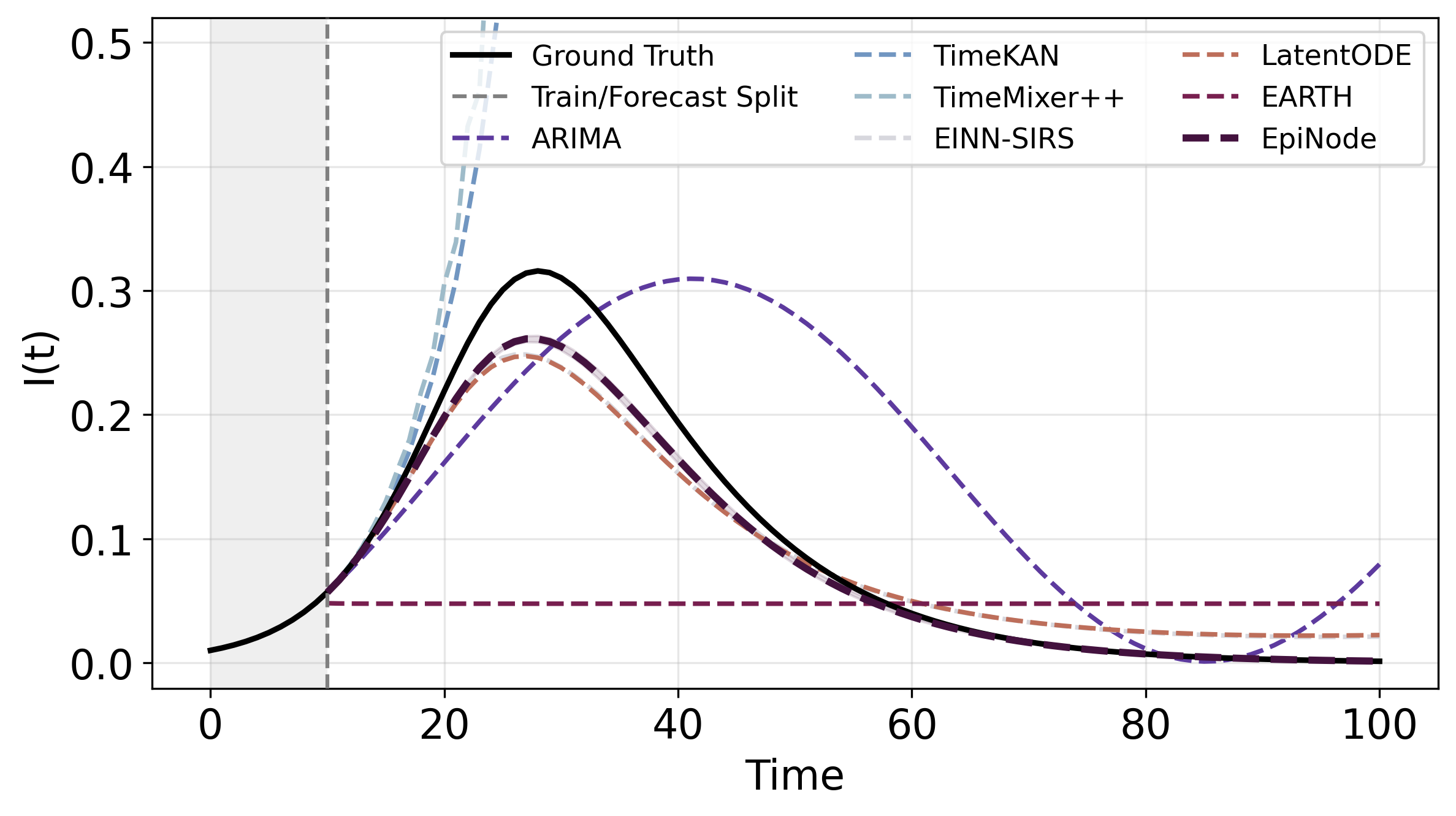}
    \caption{SIR at split=0.1}
    \label{fig:forecasts_sir}
\end{subfigure}

\vspace{2mm}
\begin{subfigure}{0.32\textwidth}\centering
    \includegraphics[width=\textwidth]{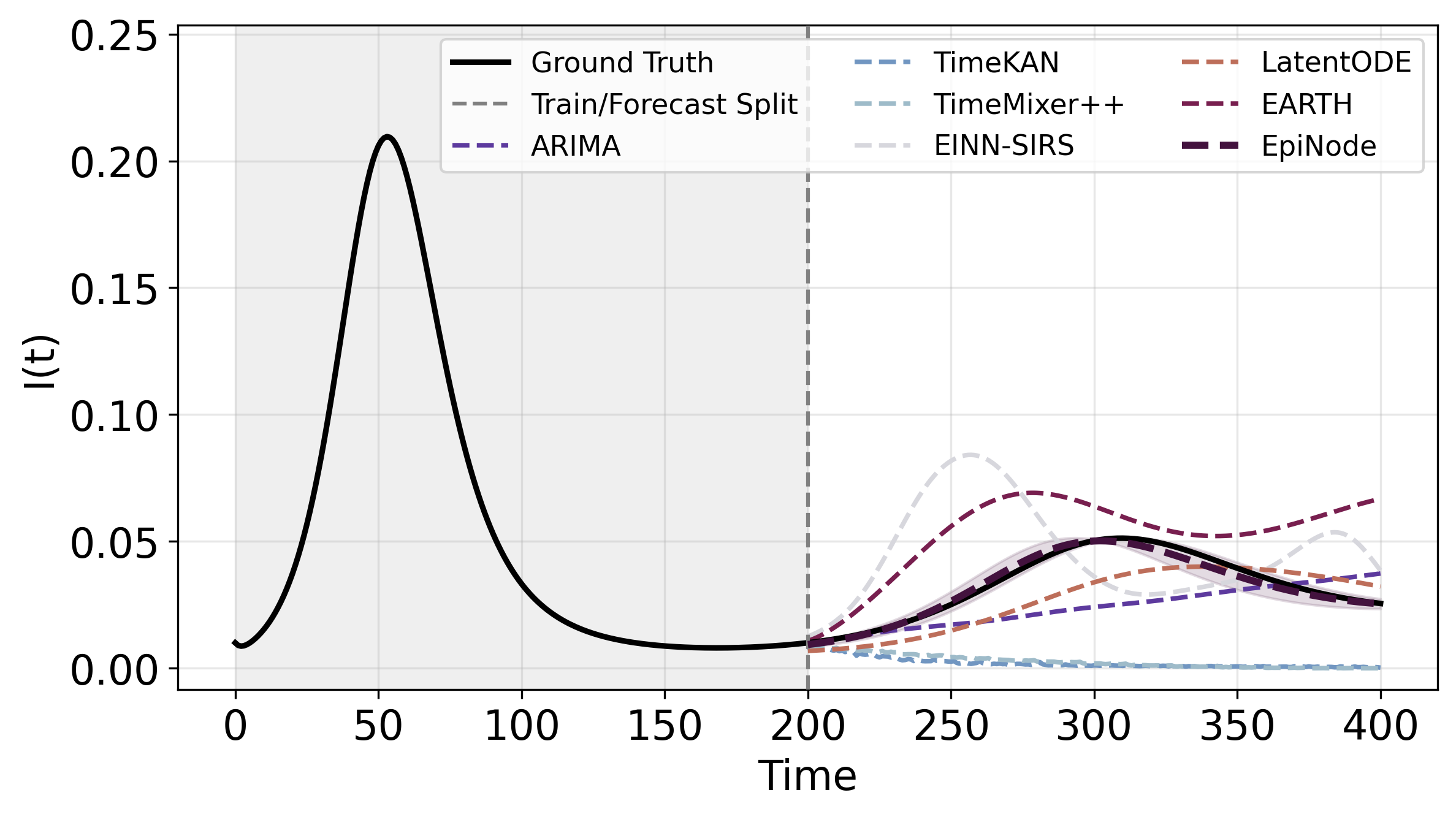}
    \caption{SEIRS at split=0.5}
    \label{fig:forecasts_seirs}
\end{subfigure}\hfill%
\begin{subfigure}{0.32\textwidth}\centering
    \includegraphics[width=\textwidth]{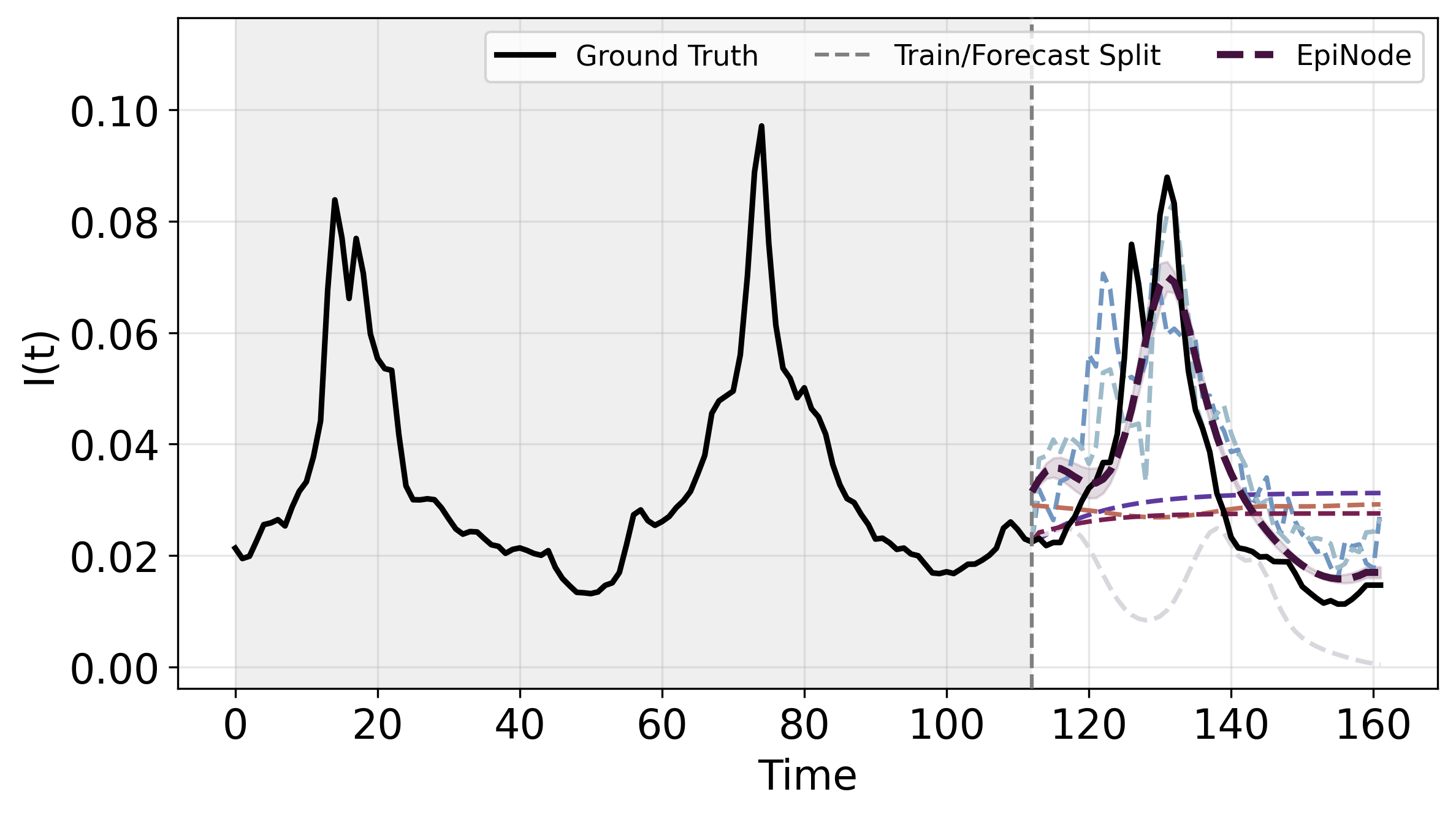}
    \caption{ILI at split=0.7}
    \label{fig:forecasts_ili}
\end{subfigure}\hfill%
\begin{subfigure}{0.32\textwidth}\centering
    \includegraphics[width=\textwidth]{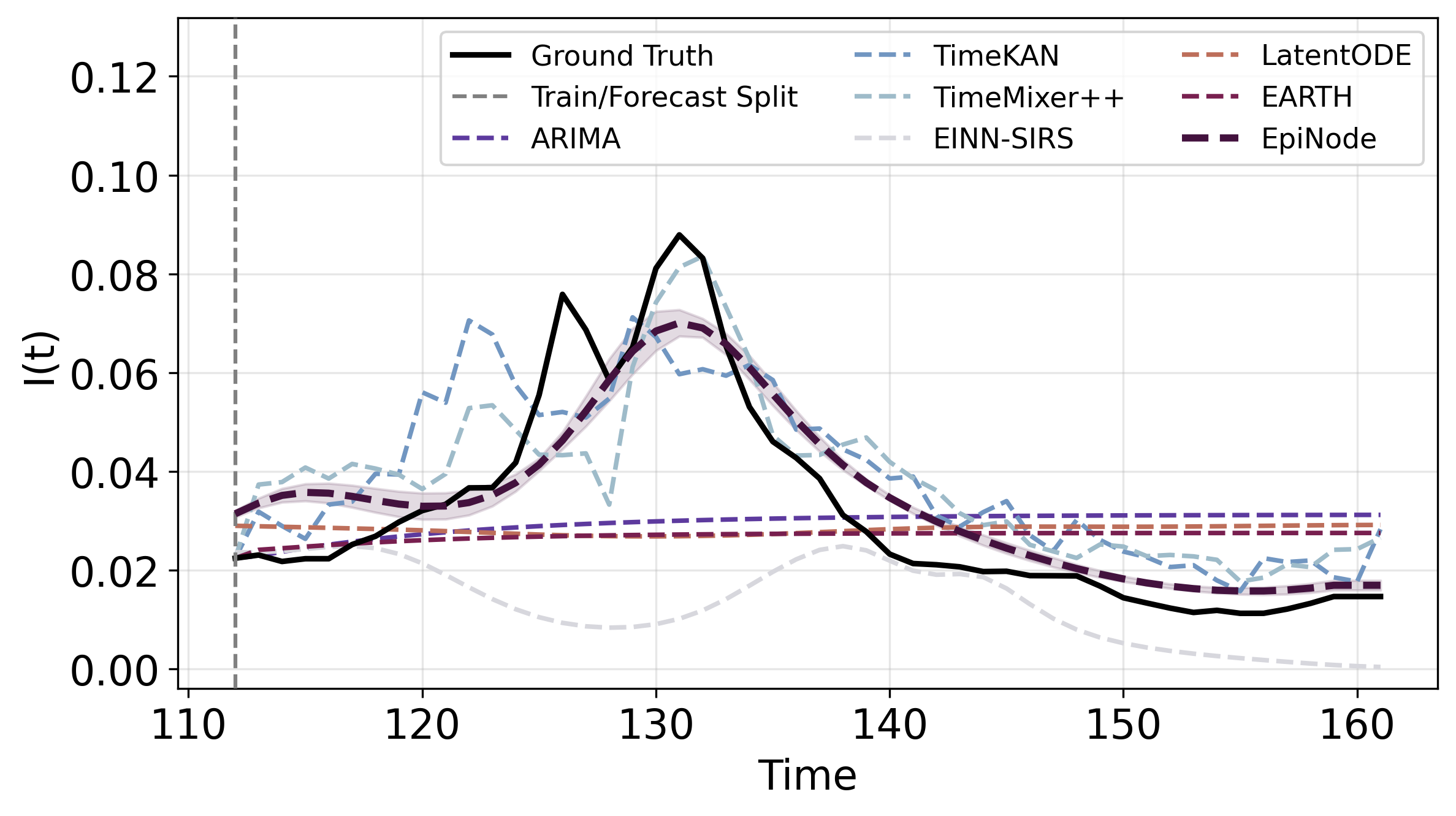}
    \caption{ILI (Year 3 zoom)}
    \label{fig:forecasts_ili_zoom}
\end{subfigure}

\caption{Forecast comparison across synthetic and real datasets at the headline train/forecast split per dataset. }
\label{fig:benchmark_forecasts}
\end{figure*}

\begin{table*}[!t]
\centering
\fontsize{9}{9.5}\selectfont
\caption{Benchmark comparison across synthetic and real datasets (RMSE, mean $\pm$ std).}
\label{tab:forecast_accuracy_benchmark}

\begin{tabular}{lccccc}
\toprule
\textbf{Method} &
\textbf{SIRS (Fixed)} &
\textbf{SIRS (Varying)} &
\textbf{SIR (Fixed)} &
\textbf{SEIRS (Fixed)} &
\textbf{ILI} \\
\midrule

ARIMA        & 0.0322 & 0.8179 & 0.0961 & 0.0146 & 0.0216 \\
LSTM         & 0.0324 (0.0000) & 0.1310 (0.0130) & 0.1132 (0.0008) & 0.0815 (0.0091) & 0.0224 (0.0045) \\
EINN-SIRS    & 0.0345 (0.0031) & \underline{0.0247} (0.0049) & \underline{0.0305} (0.0050) & 0.0295 (0.0047) & 0.0302 (0.0033) \\
EINN-SEIRm   & 0.0752 (0.0308) & 0.1038 (0.0290) & 0.1400 (0.0109) & 0.0563 (0.0236) & 0.0425 (0.0194) \\
NeuralODE    & 0.0269 (0.0102) & 0.0497 (0.0103) & 0.0977 (0.0295) & 0.0307 (0.0098) & 0.0299 (0.0069) \\
LatentODE    & \underline{0.0046} (0.0039) & 0.0405 (0.0036) & 0.0326 (0.0113) & \underline{0.0111} (0.0093) & 0.0228 (0.0035) \\
KAN-ODEs     & 0.0637 (0.0002) & 0.0564 (0.0005) & 0.1234 (0.0027) & 0.0259 (0.0000) & \underline{0.0209} (0.0000) \\
EARTH        & 0.0243 (0.0093) & 0.0588 (0.0115) & 0.1186 (0.0002) & 0.0259 (0.0185) & 0.0219 (0.0003) \\
\midrule
\textbf{EpiNode (Ours)}
& \textbf{0.0022} (0.0003) & \textbf{0.0195} (0.0044) & \textbf{0.0223} (0.0017) & \textbf{0.0041} (0.0009) & \textbf{0.0093} (0.0006) \\

\bottomrule
\end{tabular}

\vspace{2pt}
{\footnotesize
\raggedright
\textit{\ \ \ \ \ \ \ \ \ \ \ \ \ \ \ \ \ Note: ARIMA is deterministic and therefore no standard deviation is reported.} 
\par}
\end{table*}

\begin{figure*}[!t]
\centering
\begin{subfigure}[t]{0.32\linewidth}
    \centering
    \includegraphics[width=\linewidth]{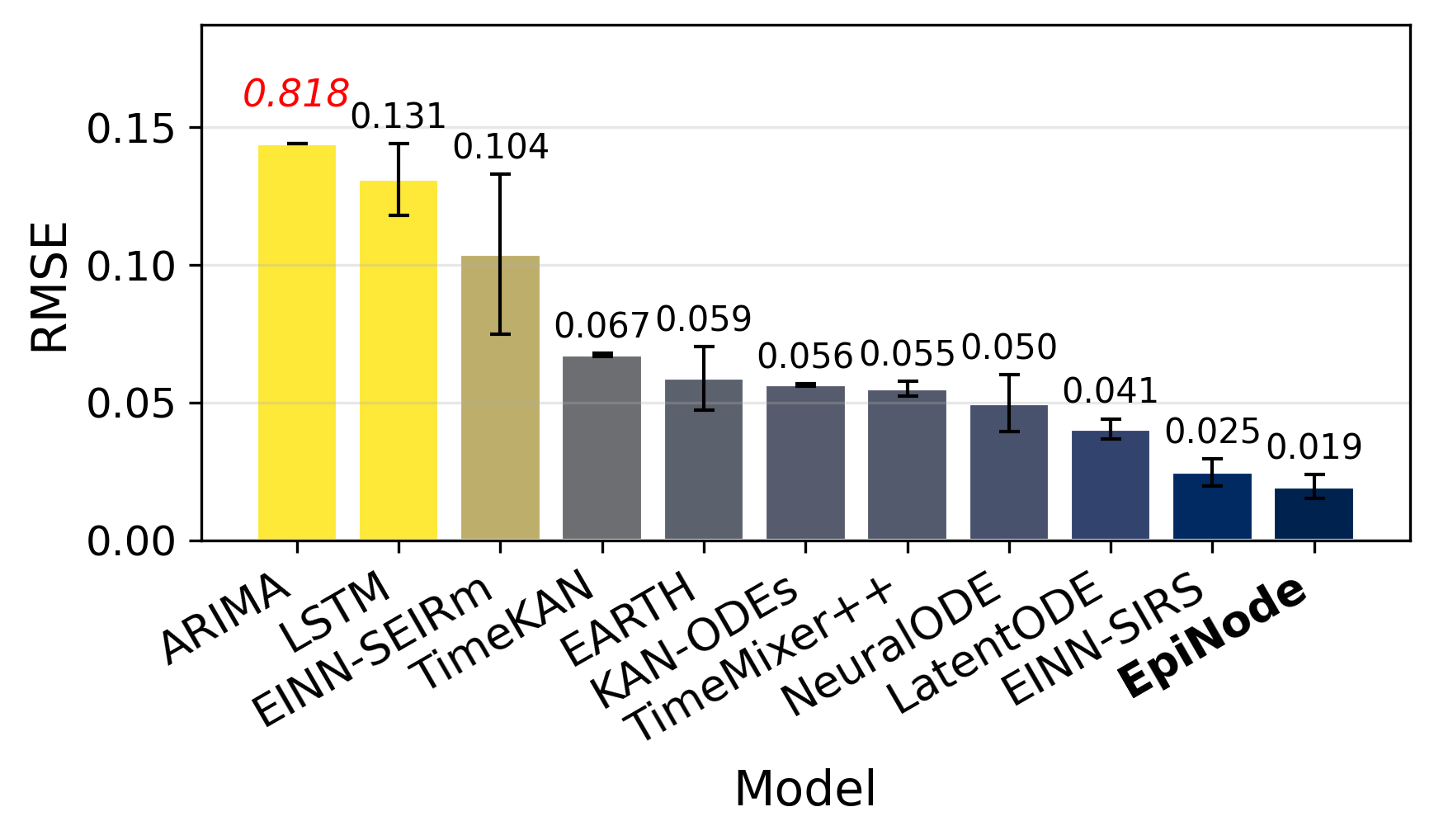}
    \caption{SIRS (Varying)}
    \label{fig:benchmark_overall_sirs_varying_rmse}
\end{subfigure}
\hfill
\begin{subfigure}[t]{0.32\linewidth}
    \centering
    \includegraphics[width=\linewidth]{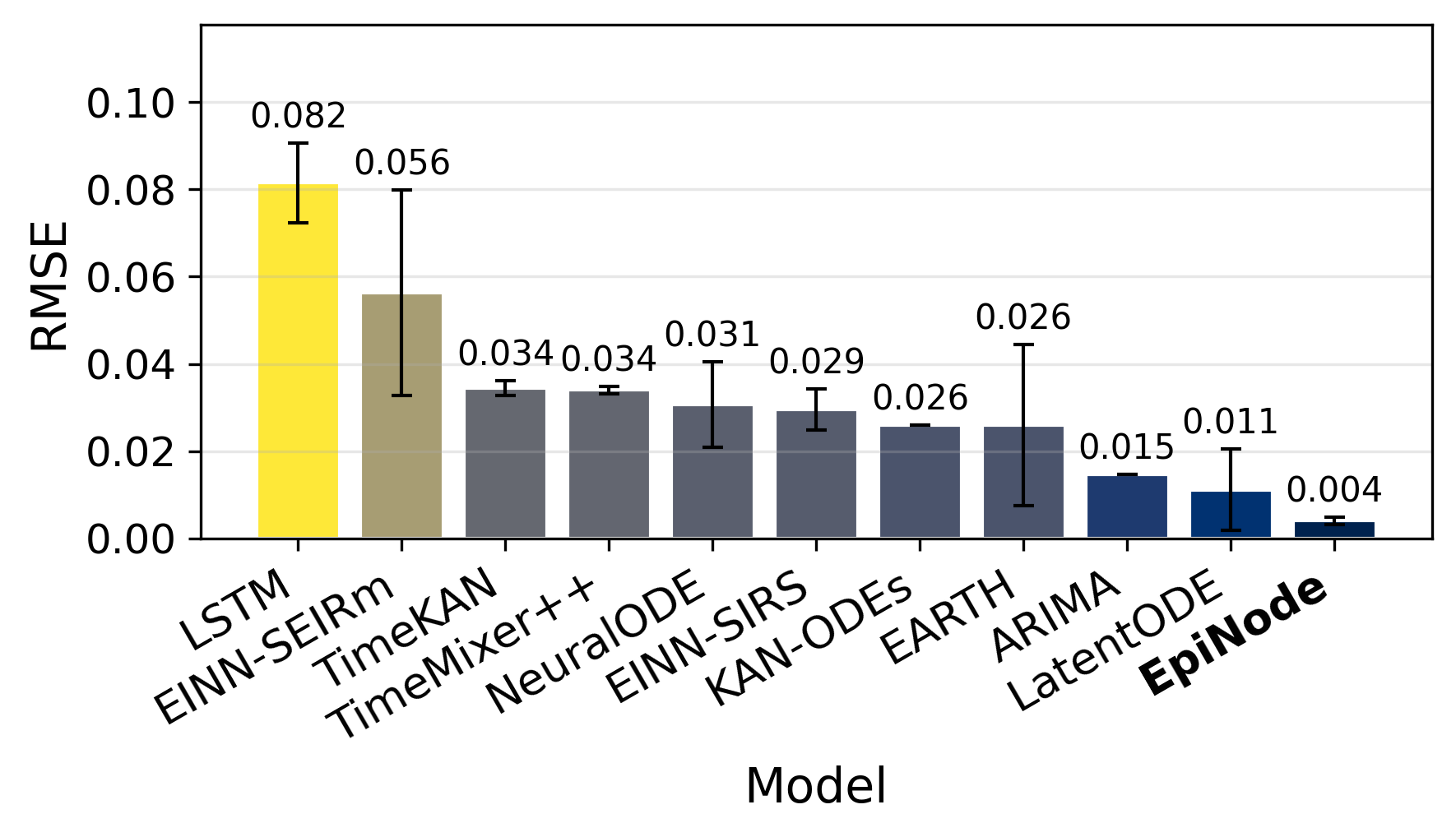}
    \caption{SEIRS}
    \label{fig:benchmark_overall_seirs}
\end{subfigure}
\hfill
\begin{subfigure}[t]{0.32\linewidth}
    \centering
    \includegraphics[width=\linewidth]{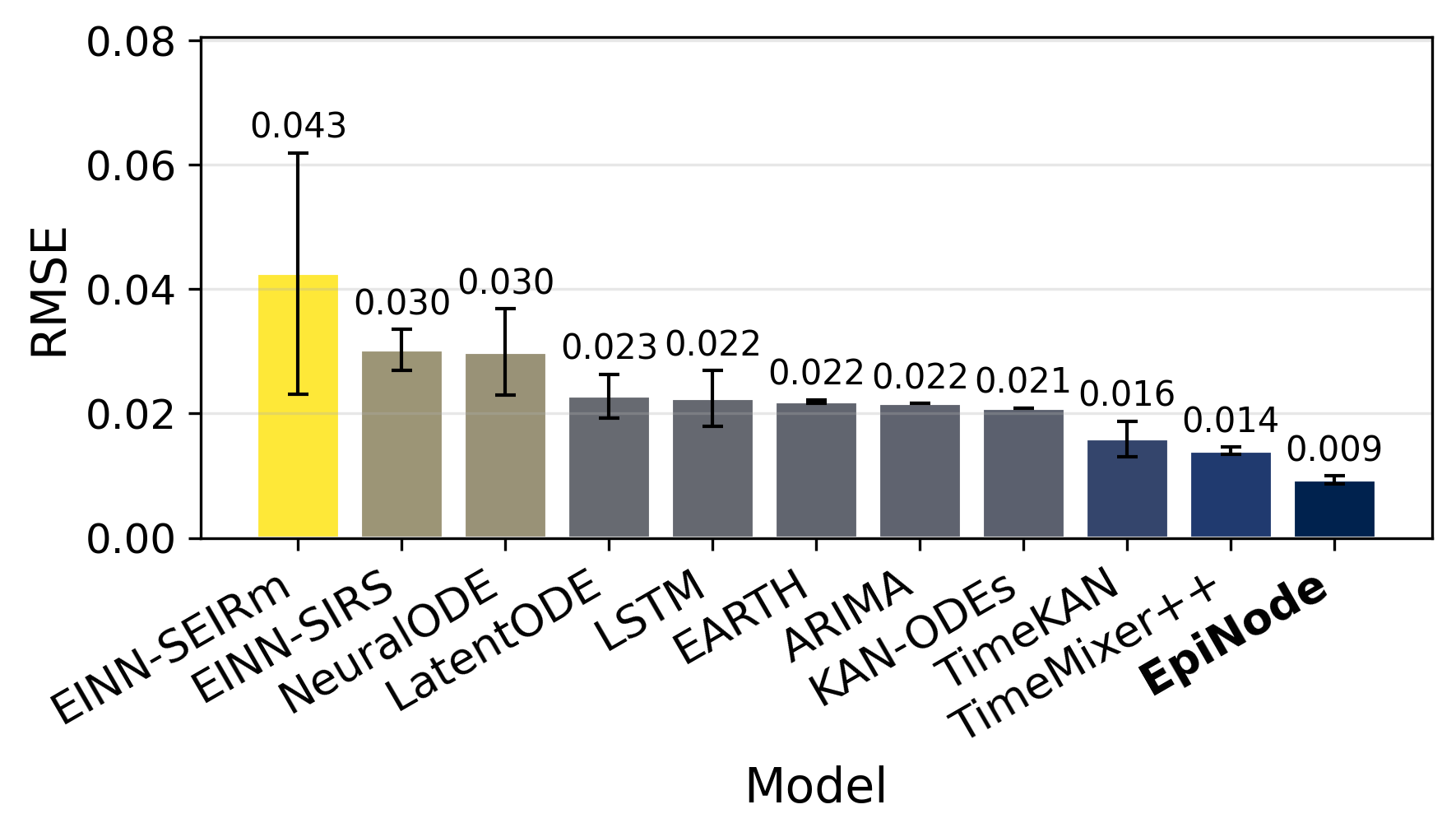}
    \caption{ILI}
    \label{fig:benchmark_overall_ili_rmse}
\end{subfigure}
\caption{Overall benchmark RMSE on (a) non-stationary, (b) physics-mismatched, and (c) real-world datasets.}
\vspace{-2mm}
\label{fig:benchmark_overall_rmse}
\end{figure*}

Figure~\ref{fig:benchmark_forecasts} compares forecast trajectories across synthetic and real datasets at representative train/forecast splits. Table~\ref{tab:forecast_accuracy_benchmark} and Figures~\ref{fig:benchmark_overall_rmse},~\ref{fig:benchmark_overall_rmse_synthetic} show the overall RMSE across both synthetic and real datasets report the corresponding RMSE means and standard deviations. \model{} attains the lowest mean RMSE on all five datasets.

On SIRS data with fixed parameters (Figure~\ref{fig:forecasts_sirs-fixed}), while most models achieve reasonable short-term accuracy, \model{} maintains stable long-horizon rollouts without drift in unobserved compartments (RMSE $0.0022$ $vs$ $0.0046$ for the next-best LatentODE). 
On SIRS data with time-varying parameters (Figure~\ref{fig:forecasts_sirs-varying}), which introduces periodic forcing, the gap between \model{} and baselines widens. 
TimeMixer++ ($0.0550$) and LSTM ($0.1310$) flatten to near zero and miss subsequent waves entirely. LatentODE captures the shape of the first post-split peak but underestimates its magnitude and fails to recover later dynamics. EINN-SIRS tracks the first peak well but drifts over longer horizons, and EARTH overshoots progressively across waves.

On physics-mismatched settings (SIR in Figure~\ref{fig:forecasts_sir} and SEIRS in Figure~\ref{fig:forecasts_seirs}), \model{} remains stable despite under- or over-parameterization of the fundamental SIRS model. On SIR, which lacks immunity waning, the model reaches RMSE $0.0223$ ($27\%$ below EINN-SIRS at $0.0305$); on SEIRS, which includes an exposed compartment, \model{} still achieves RMSE $0.0041$ ($63\%$ below LatentODE at $0.0111$), indicating that the latent Neural ODEs compensate for unmodeled compartments.

For the real-world dataset (ILI HHS 4), Figures~\ref{fig:forecasts_ili} and~\ref{fig:forecasts_ili_zoom} show that \model{} consistently outperforms strong univariate baselines and continuous-time models, achieving RMSE $0.0093$ compared to KAN-ODEs ($0.0209$), EARTH ($0.0219$), and LatentODE ($0.0228$). In particular, \model{} exhibits superior stability during post-split rollouts, avoiding the oscillatory or mean-reverting failures observed in purely data-driven models.

\subsubsection{Peak Errors}
\label{sec:peak}

\begin{figure}[htbp]
    \centering
    \includegraphics[width=.8\columnwidth]{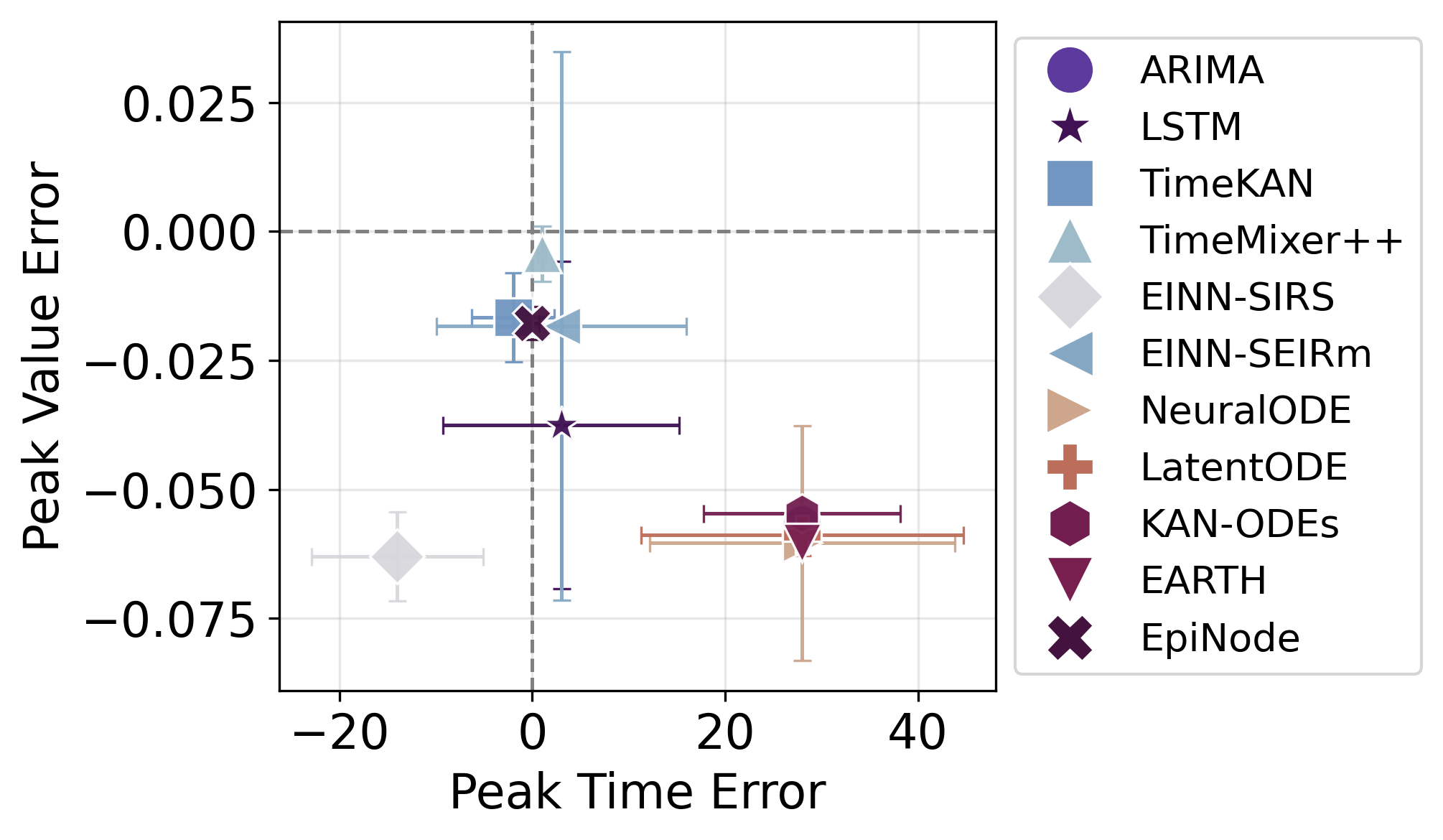}
    \caption{Peak error (magnitude \& timing) of ILI}
\vspace{-3mm}
    \label{fig:peak_hhs4}
\end{figure}

On ILI (Figure~\ref{fig:peak_hhs4}), \model{} achieves the best peak timing among all methods, with a bias of $0.0$ weeks ($\sigma = 0.7$) and a competitive peak-magnitude bias of $-0.018$ ($\sigma \approx 3\times 10^{-3}$). 
TimeMixer++ achieves the smallest peak magnitude error ($-0.004$) and low timing bias ($+1$ week), but its overall forecast RMSE remains higher than \model{}'s due to weaker tracking outside the peak region.
The remaining methods split into two regimes. The first is heavily biased predictors whose peak far from the true peak (ARIMA, NeuralODE, LatentODE, KAN-ODEs, and EARTH all at $+28$ weeks, EINN-SIRS at $-14$ weeks). The second is lower-bias but high-variance predictors (LSTM $+3 \pm 12$ weeks, EINN-SEIRm $+3 \pm 13$ weeks, TimeKAN $-2 \pm 4$ weeks). 

    \vspace{-1mm}

\subsection{Applications} \label{sec:application}

\subsubsection{Parameter inference}
\label{sec:parameter-inference}


\begin{figure}[h]
    \centering
    \begin{subfigure}[h]{\columnwidth}
        \centering
        \includegraphics[width=\linewidth]{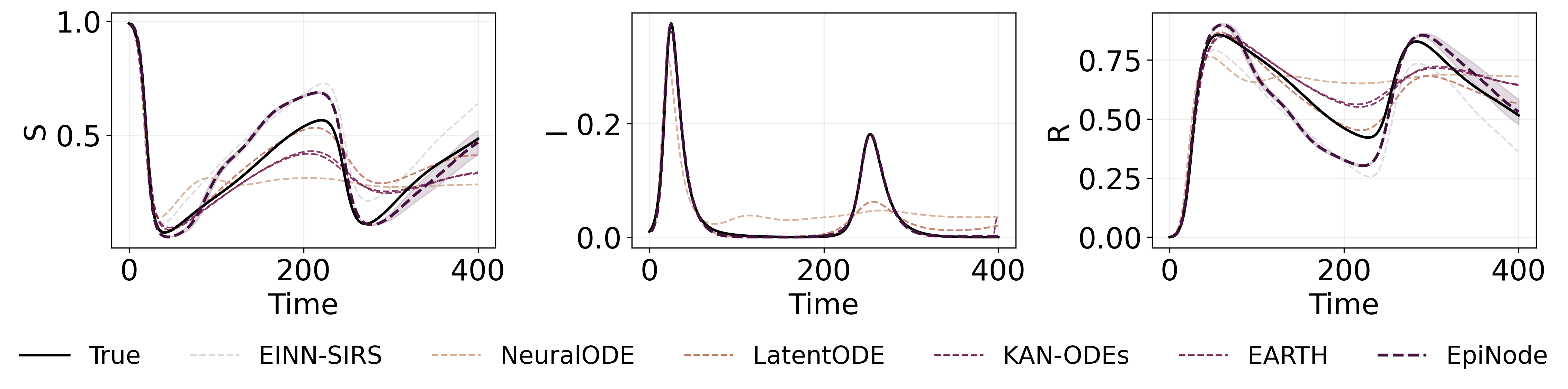}
        \caption{\small SIRS (Varying) compartments} \label{fig:compartments_sirs_varying}
    \end{subfigure} 
    
    
    \begin{subfigure}[h]{\columnwidth}
        \centering
        \includegraphics[width=\linewidth]{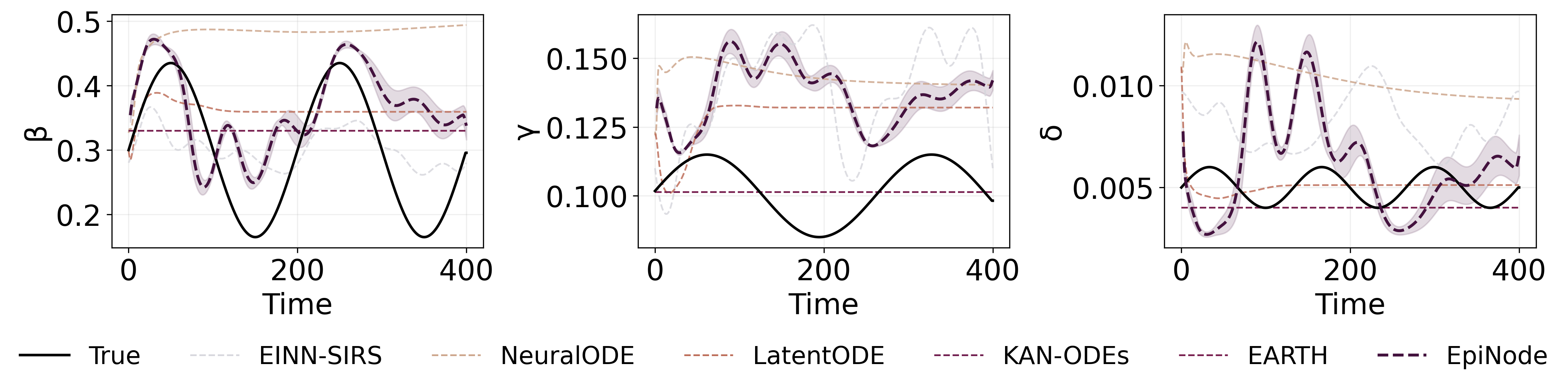}
        \caption{\small SIRS (Varying) parameters}\label{fig:parameters_sirs_varying}
    \end{subfigure}
    \caption{Unobserved compartments (a) and parameters (b) inferred from the full infection series.} 
\vspace{-2mm} 
\label{fig:compartment&parameter}
\end{figure}

\begin{table}[h]
\centering
\fontsize{7.5}{9}\selectfont
\caption{Sign agreement for inferred parameters (\%, mean $\pm$ std).}
\label{tab:sign_agreement}
\begin{tabular}{lccc}
\toprule
\textbf{Method} & \textbf{$\Delta\beta(t)$} & \textbf{$\Delta\gamma(t)$} & \textbf{$\Delta\delta(t)$} \\
\midrule
EINN-SIRS & 59.8 $\pm$ 2.9 & \textbf{52.0 $\pm$ 5.4} & \textbf{43.3 $\pm$ 7.9} \\
LatentODE & 32.8 $\pm$ 12.6 & 20.5 $\pm$ 12.1 & 16.8 $\pm$ 9.5 \\
NeuralODE & 18.1 $\pm$ 14.2 & 28.0 $\pm$ 24.4 & 17.6 $\pm$ 21.8 \\
\midrule
\textbf{EpiNode} & \textbf{66.7 $\pm$ 1.1} & \underline{50.6} $\pm$ 1.2 & \underline{41.3} $\pm$ 0.5 \\
\bottomrule
\end{tabular}

\vspace{1mm}
\footnotesize
\emph{Note: \footnotesize
EARTH and KAN-ODEs are omitted as they infer constant trajectories, yielding undefined values across parameters. \ \ \ \ \ \ \ \ \ \ \ } 
\end{table}
\vspace{-3mm}

Across synthetic data with ground-truth compartments and parameters (Figures~\ref{fig:compartment&parameter}, \ref{fig:compartments}, \ref{fig:parameters}), \model{} reconstructs the unobserved $S(t)$ and $R(t)$ as well as the time-varying $\beta(t)$, $\gamma(t)$, and $\delta(t)$ using only infection counts as input from the full time series. Most baselines either flatten to constants or drift away from the true dynamics, particularly for the parameter rates. 
For SIRS (Varying) as shown in Table~\ref{tab:sign_agreement}, only EpiNode and EINN-SIRS exceed the 50\% (random) success level for recovering changes in the transmission rate $\beta(t)$, with EpiNode achieving the highest agreement (66.7\%). This indicates that \model{} tracks the timing of changes in transmission rather than merely fitting the observed infection curve.
Across methods, inference of $\gamma(t)$ and $\delta(t)$ is generally less consistent than that of $\beta(t)$, as reflected by their lower sign agreement. Nevertheless, EpiNode ranks second for both parameters, achieving sign agreements of 50.6\% and 41.3\%, respectively.

Figure~\ref{fig:hhs_parameter} shows inferred parameters for real data, where ground-truth parameters are unavailable. For a representative region HHS 4, (Figure~\ref{fig:params_hhs4}), the estimated $\beta(t)$ exhibits clear seasonal oscillations aligned with major ILI waves, while $\gamma(t)$ remains stable and $\delta(t)$ varies smoothly at lower amplitude. Aggregated across all HHS regions (Figure~\ref{fig:params_all_hhs}), $\beta(t)$ shows consistent seasonal modulation with moderate regional heterogeneity, reflecting shared seasonal forcing and region-specific transmission intensity. The inferred parameters remain smooth, bounded, and temporally coherent across regions, supporting stable and interpretable parameter reconstruction from real-world surveillance data.



    \vspace{-1mm}
\section{Conclusion}

This work highlights the importance of integrating multi-scale structure, continuous-time latent dynamics, and mechanistic constraints for epidemic forecasting under partial observability. By explicitly decomposing the observed signal into trend, seasonal, and residual components \cite{timekan, zhang2022detrendattendrethinkingattention,He_2022}, EpiNode separates these effects into structured, low-dimensional control signals that guide the latent neural ODE.
Our experimental results demonstrate that \model{} produces interpretable time-varying parameter trajectories and generates forecasts consistent with known epidemic behavior. These parameter estimates provide insights beyond point forecasts, enabling retrospective analysis and hypothesis generation about the underlying drivers of epidemic dynamics. In addition, the mechanistic component can be replaced with alternative formulations, such as multi-strain models \cite{andreasen1997pathogen}. Such extensions could better capture pathogen interactions and complex epidemic dynamics.


Despite these advantages, EpiNode has limitations. 
The current formulation focuses on single-region, deterministic dynamics and does not explicitly model uncertainty, spatial coupling, or intervention effects.
Moreover, TSR decomposition is treated as a preprocessing step rather than a learned component, and performance may depend on the choice of decomposition method.
Addressing these limitations by incorporating probabilistic latent dynamics, spatial interactions, intervention effects, or learnable decomposition modules represents promising directions for future work.

\newpage
\section*{Acknowledgements}



We thank the ICML 2026 reviewers and the area chair for their thoughtful and constructive feedback, which materially improved the presentation of results.

This work is supported in part by US National Science Foundation grants 
CCF-1918770,  
IIS-2509636, 
IIS-2312794, 
and DBI-2412389. 
Any opinions, findings, and conclusions or recommendations expressed in this material are those of the authors and do not necessarily reflect the views of the sponsors.


\section*{Impact Statement}
This work develops \model{}, a hybrid neural--physical framework for epidemic dynamics forecasting under partial observability that integrates multi-scale signal decomposition, controlled neural ODEs, and mechanistic SIRS dynamics within a unified architecture. By jointly forecasting infection trajectories and inferring bounded, interpretable time-varying epidemiological parameters, \model{} bridges predictive performance and epidemiological insight, and generalizes across seasonal and non-seasonal diseases as well as single- and multi-wave regimes. Such accurate and interpretable forecasts have the potential to support public-health planning, early warning, and retrospective analysis by improving understanding of disease transmission and peak behavior.

At the same time, forecasts and inferred parameters derived from such models should be interpreted with care, as they depend on data quality, modeling assumptions, and incomplete observations. Misuse or overreliance on model outputs without appropriate domain expertise could lead to misguided decisions. We emphasize that \model{} is intended as a decision-support and analysis tool rather than a standalone policy-making system. We do not foresee any novel ethical concerns beyond those commonly associated with applying machine learning to public-health data, and we encourage responsible use in conjunction with epidemiological expertise and transparent communication of model limitations and uncertainty.


\section*{Software and Data}
We release the full implementation at \url{https://github.com/yiqisu/EpiNode.git}.





\newpage
\bibliography{node}
\bibliographystyle{icml2026}

\newpage
\appendix

\setcounter{table}{0}
\renewcommand{\thetable}{A\arabic{table}}

\setcounter{figure}{0}
\renewcommand{\thefigure}{A\arabic{figure}}

\onecolumn

\section{Failure Modes}
\label{sec:failure}
We illustrate the bidirectional autoencoder-based Neural ODE (AE-NODE) architecture described in Section~\ref{sec:failure-modes}.

\begin{figure}[htbp]
   \centering
   \includegraphics[width=.7\columnwidth]{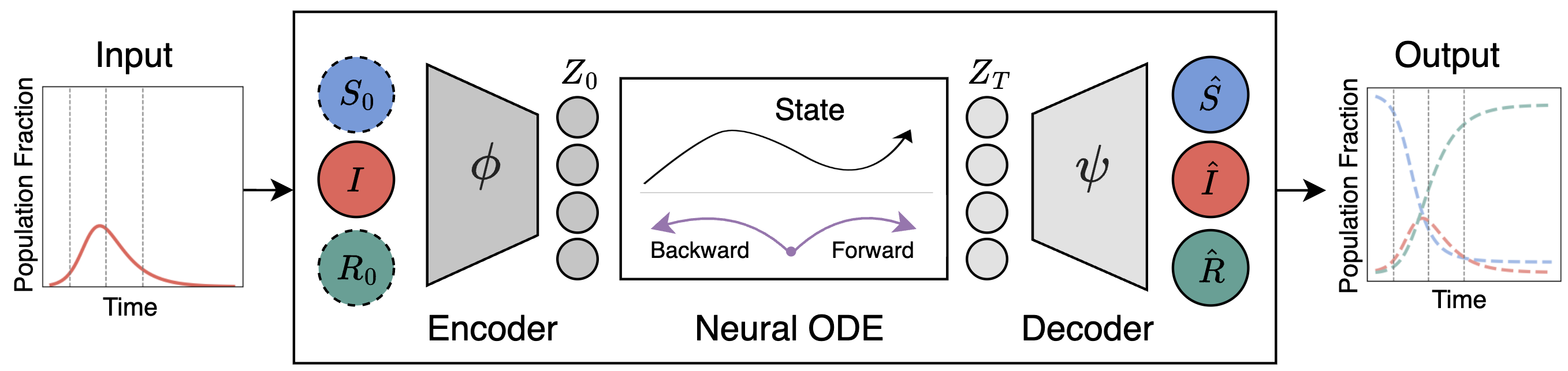}
   \caption{Overview of bidirectional AE-NODE framework.
   }
   \label{fig:ae-nde-bid}
\end{figure}

\section{Training and Inference Procedure}
\label{sec:algorithm}
Algorithm~\ref{alg:tsrode} summarizes the end-to-end training and inference procedure for \model. Given a partially observed infection time series, the algorithm first performs trend–season–residual decomposition to extract multi-scale control signals, which optionally undergo time-delay embedding. These controls drive a set of latent continuous-time Neural ODEs whose states are fused to infer time-varying epidemiological parameters. The inferred parameters are then used to advance the mechanistic SIRS model via numerical integration, producing epidemic state forecasts. Model parameters are optimized by minimizing the prediction error on infections over the training window, with gradients propagated through both the Neural ODE solvers and the mechanistic dynamics. At inference time, the learned model is rolled out beyond the observation window to generate long-horizon forecasts, peak estimates, and interpretable parameter trajectories.

\section{Signal Decomposition Methods}
\label{sec:decomposition-details}
We compare several signal decomposition techniques that differ in their assumptions about temporal structure, frequency separation, and modeling capacity. Each method decomposes the observed infection time series into components that are subsequently used to control latent continuous-time dynamics.

\paragraph{Moving average (MA).}
The moving average decomposition applies a sliding-window average to smooth short-term fluctuations and extract a low-frequency trend component. Residuals are obtained by subtracting the smoothed signal from the original series. MA provides a simple baseline that captures coarse trends but does not explicitly model seasonality or frequency structure, and is sensitive to window size selection.

\paragraph{Seasonal–trend decomposition using Loess (STL).}
STL decomposes a time series into additive trend, seasonal, and residual components using locally weighted regression (LOESS). It assumes a fixed seasonal period and smooth temporal evolution, making it effective for stationary or weakly non-stationary seasonal patterns. However, STL does not enforce explicit frequency separation and may struggle when seasonal dynamics vary over time or across epidemic waves.

\paragraph{Variational mode decomposition (VMD).}
VMD decomposes a signal into a predefined number of intrinsic mode functions, each constrained to be band-limited around a learned center frequency. The decomposition is obtained by solving a variational optimization problem in the frequency domain that jointly minimizes bandwidth and reconstruction error. VMD produces components with well-separated frequency content and is robust to noise, making it well suited for isolating multi-scale epidemic dynamics.

\paragraph{Wavelet.}
Wavelet-based decomposition represents the signal using a set of scaled and shifted wavelet basis functions, yielding a multi-resolution time–frequency representation. This approach captures both local temporal variations and global structure. While wavelets provide strong localization in time and frequency, the resulting components are not necessarily narrowband, and performance depends on the choice of wavelet family and decomposition depth.

\paragraph{SSA–VMD (hybrid decomposition).}
SSA–VMD combines Singular Spectrum Analysis (SSA) with VMD to leverage the strengths of both methods. SSA first separates the signal into dominant subspaces corresponding to trend and oscillatory 

\begin{center}
\begin{minipage}{\linewidth}
\hrule
\vspace{1mm}
\refstepcounter{algorithm}
\textbf{Algorithm~\thealgorithm} EpiNode training and forecasting
\label{alg:epinode}
\vspace{1mm}
\hrule
\vspace{1mm}
 \label{alg:tsrode}
\begin{algorithmic}[1]
\REQUIRE Observations $\{(t_i, I_i)\}_{i=0}^{T-1}$, split index $t_{\mathrm{split}}$, delay params $(m,\tau)$, trend-fit window $L_{\mathrm{fit}}$, step size schedule $\Delta t_i=t_{i+1}-t_i$; model components: VMD (\S\ref{sec:vmd}), channel-aware extrapolation rules \eqref{eq:extrap-trend}--\eqref{eq:extrap-resid}, collaborative ODE fields $f_{\theta_c}$ \eqref{eq:controlled-ode}, fusion MLP \eqref{eq:fuse}, parameter decoder $g_\phi$ \eqref{eq:param}, SIRS solver \eqref{eq:sirs}
\ENSURE Forecasted states $\{\hat{\mathbf{y}}(t_i)\}_{i=0}^{T_{\mathrm{eff}}-1}$, parameters $\{\beta(t_i),\gamma(t_i),\delta(t_i)\}_{i=0}^{T_{\mathrm{eff}}-1}$

\STATE \textbf{Causal TSR decomposition:} $\{u_k\}_{k=1}^{K} \leftarrow \mathrm{VMD}\bigl(\{I_i\}_{i=0}^{t_{\mathrm{split}}-1}\bigr)$
\STATE Sort modes by center frequency: $u_{\sigma(1)},\ldots,u_{\sigma(K)}$ with $\omega_{\sigma(1)}\le\cdots\le\omega_{\sigma(K)}$ \hfill\eqref{eq:tsr}
\STATE Assign $T_i\!\leftarrow\!\sum_{k=1}^{\lfloor K/3\rfloor}u_{\sigma(k)}(t_i)$,\; $S_i\!\leftarrow\!\sum_{k=\lfloor K/3\rfloor+1}^{\lfloor 2K/3\rfloor}u_{\sigma(k)}(t_i)$,\; $R_i\!\leftarrow\!$ remainder

\STATE \textbf{Channel-aware extrapolation.} Estimate $(a,b)\leftarrow \arg\min_{a,b}\sum_{j=t_{\mathrm{split}}-L_{\mathrm{fit}}}^{t_{\mathrm{split}}-1}(T_j - a\,t_j - b)^2$ and $p_S\leftarrow$ dominant period of $\{S_j\}_{j=0}^{t_{\mathrm{split}}-1}$ via periodogram (fallback $52$ weekly, $7$ daily). For each $i\in[t_{\mathrm{split}}, T)$:
\STATE \quad $T_i \leftarrow a\,t_i + b$ \hfill\eqref{eq:extrap-trend}
\STATE \quad $k_i \leftarrow \lceil (i - t_{\mathrm{split}} + 1)/p_S \rceil$;\, $S_i \leftarrow S_{i - p_S k_i}$ \hfill\eqref{eq:extrap-season}
\STATE \quad $R_i \leftarrow \bar{R}_{\mathrm{train}}$ \hfill\eqref{eq:extrap-resid}
\STATE Concatenate causal VMD output and extrapolated extensions to obtain $\{T_i,S_i,R_i\}_{i=0}^{T-1}$ on the full window.

\STATE \textbf{Time-delay embedding:} construct $\mathbf{u}_i^{(c)} = [c_i, c_{i-\tau}, \ldots, c_{i-(m-1)\tau}]^\top$ for $c\in\{T,S,R\}$ \hfill\eqref{eq:delay}
\STATE Set offset $\delta_0 = (m-1)\tau$, effective length $T_{\mathrm{eff}} = T - \delta_0$, effective split $t_{\mathrm{split}}^{\mathrm{eff}} = t_{\mathrm{split}} - \delta_0$

\STATE \textbf{Initialize:} $\hat{\mathbf{y}}(t_{\delta_0}) = [1-I_{\delta_0},\, I_{\delta_0},\, 0]^\top$ \hfill\eqref{eq:state},\eqref{eq:init}
\STATE $\mathbf{h}^{(T)}(t_{\delta_0})=\mathbf{0}$, $\mathbf{h}^{(S)}(t_{\delta_0})=\mathbf{0}$, $\mathbf{h}^{(R)}(t_{\delta_0})=\mathbf{0}$ \hfill\eqref{eq:latents}

\FOR{epoch $=1$ to $E$}
    \STATE Set loss $\mathcal{L}\leftarrow 0$
    \FOR{$i=0$ to $T_{\mathrm{eff}}-1$}
        \STATE \textbf{Latent ODE step (if $i>0$):}
        \IF{$i>0$}
            \STATE $\mathbf{h}^{(c)}(t_i)\leftarrow \mathrm{ODEINT}(f_{\theta_c}, \mathbf{h}^{(c)}(t_{i-1}), [t_{i-1},t_i];\, \mathbf{u}_{i-1}^{(c)})$ for $c\in\{T,S,R\}$ \hfill\eqref{eq:controlled-ode}
        \ENDIF

        \STATE \textbf{Fuse latents:} $\mathbf{h}(t_i)\leftarrow \mathrm{Fuse}([\mathbf{h}^{(T)}(t_i);\mathbf{h}^{(S)}(t_i);\mathbf{h}^{(R)}(t_i)])$ \hfill\eqref{eq:fuse}
        \STATE \textbf{Decode rates:} $(\tilde\beta,\tilde\gamma,\tilde\delta)\leftarrow g_\phi(\mathbf{h}(t_i))$; apply affine scaling \hfill\eqref{eq:param}

        \STATE \textbf{Physics rollout:} $\hat{\mathbf{y}}(t_{i+1}) \leftarrow \mathrm{RK4Step}(\hat{\mathbf{y}}(t_i), \beta(t_i),\gamma(t_i),\delta(t_i), \Delta t_i)$ \hfill\eqref{eq:sirs}
        \STATE Apply simplex stabilization / mass correction \hfill\eqref{eq:mass}

        \IF{$i < t_{\mathrm{split}}^{\mathrm{eff}}$}
            \STATE $\mathcal{L}\leftarrow \mathcal{L} + w_i(\hat{I}(t_i)-I_{i+\delta_0})^2$ \hfill\eqref{eq:loss}
        \ENDIF
    \ENDFOR
    \STATE $\mathcal{L}\leftarrow \mathcal{L}/t_{\mathrm{split}}^{\mathrm{eff}}$ \hfill\eqref{eq:loss}
    \STATE Update parameters of $\{f_{\theta_c}\}_{c\in\{T,S,R\}}$, $\mathrm{Fuse}$, $g_\phi$ by backpropagation through ODE solvers
\ENDFOR

\STATE \textbf{return} $\{\hat{\mathbf{y}}(t_i)\}_{i=0}^{T_{\mathrm{eff}}-1}$ and $\{\beta(t_i),\gamma(t_i),\delta(t_i)\}_{i=0}^{T_{\mathrm{eff}}-1}$
\end{algorithmic}
\vspace{1mm}
\hrule
\end{minipage}
\end{center}

modes using low-rank trajectory matrices. VMD is then applied to selected components to further refine frequency separation. This hybrid approach improves robustness in noisy settings but introduces additional complexity and hyperparameters.

\paragraph{Neural Koopman.}
Neural Koopman methods learn a latent linear dynamical system by lifting the observed time series into a higher-dimensional feature space using neural networks. Decomposition is achieved by projecting dynamics onto learned Koopman modes and eigenfunctions. This approach is fully data-driven and expressive, but does not explicitly enforce band-limited or frequency-localized components, which can lead to entangled modes under strong non-stationarity.

\section{Benchmarking Models}
\label{benchmark-details}
\begin{itemize}
    \item Autoregressive integrated moving-average (ARIMA) models serve as a classical statistical baseline for epidemic forecasting. ARIMA captures linear temporal dependencies through autoregressive and moving-average components and is commonly used for short-term epidemic prediction under near-stationary conditions.

    \item We include standard recurrent neural networks (RNN) and gated variants (LSTM/GRU) as nonlinear sequence modeling baselines. RNN-based models can capture complex temporal dependencies from data but operate as black-box predictors and do not enforce epidemiological constraints.
    
    \item TimeKAN \cite{timekan} and TimeMixer++ \cite{timemixer++} represent recent state-of-the-art univariate forecasters that leverage decomposition or multiscale mixing to capture long-range temporal patterns, providing strong data-driven baselines.
    
    \item To incorporate epidemiological structure, we include EINN \cite{einn}, a physics-informed neural model that embeds epidemic priors into neural forecasting without explicit continuous-time latent dynamics. As EINN employs SEIRm physics, we adapt the framework in two ways: 1) replace SEIRm with SIRS, 2) predict I rather than m. In both options, I is the only observed compartment.
    
    \item We further evaluate ODE-based continuous-time models, including Neural ODE \cite{neural-odes}, Latent ODE \cite{latent-odes}, and KAN-ODE \cite{kan-odes}, which model temporal evolution via learned differential equations but do not enforce epidemic physics by default. 
    
    \item Finally, we consider EARTH \cite{pmlr-v267-wan25b}, a graph-based neural ODE model that integrates epidemiological dynamics with spatial coupling, evaluated only in multi-region settings where adjacency information is available. Given our focus on single region forecasting, we treat the region as a graph with $N=1$ node and set adjacency $A = [1]$. 
\end{itemize}

Table~\ref{tab:model-comparison} summarizes the modeling capabilities of representative baselines and highlights the gaps that motivate our design.

\vspace{3mm}
\begin{table}[htbp]
\centering
\caption{
Comparison of modeling capabilities across baselines. }
\label{tab:model-comparison}
\small
\resizebox{\textwidth}{!}{%
\begin{tabular}{lccccccccc}
\toprule
\textbf{Model} 
& \textbf{Univariate} 
& \textbf{Cont.-Time} 
& \textbf{Physics} 
& \textbf{TV Params} 
& \textbf{Interpretable} 
& \textbf{Non-Seasonality} 
& \textbf{Seasonality} 
& \textbf{Stable Long-Horizon} \\
\midrule
ARIMA & \cmark  & \xmark & \xmark & \xmark & \xmark & \xmark & \xmark & \xmark \\
RNN (LSTM) & \cmark  & \xmark & \xmark & \xmark & \xmark & \xmark & \xmark & \xmark \\
TimeKAN & \cmark & \xmark & \xmark & \xmark & \xmark & \xmark & \cmark & \xmark \\
TimeMixer++ & \cmark  & \xmark & \xmark & \xmark & \xmark & \xmark & \cmark & \xmark \\
\midrule
EINN & \cmark  & \xmark & \cmark & \xmark & \cmark & \xmark & \xmark & \xmark \\
\midrule
Neural ODE & \cmark  & \cmark & \xmark & \xmark & \xmark & \xmark & \xmark & \xmark \\
Latent ODE & \cmark  & \cmark & \xmark & \xmark & \xmark & \xmark & \xmark & \xmark \\
KAN-ODE & \cmark  & \cmark & \xmark & \xmark & \xmark & \xmark & \xmark & \xmark \\
EARTH & $\dagger$ & \cmark  & \cmark & \cmark & \cmark & \cmark & \cmark & \cmark \\
\midrule
\textbf{\model{} (ours)} 
& \cmark & \cmark & \cmark & \cmark & \cmark & \cmark & \cmark & \cmark \\
\bottomrule
\end{tabular}
}
\vspace{2pt}
{\footnotesize\raggedright\textit{
``Univariate'' denotes native support for single-series forecasting. ``TV Params'' denotes inference of time-varying transmission parameters. ``Stable Long-Horizon'' indicates empirically demonstrated long-horizon robustness. $\dagger$ EARTH is designed for multi-region forecasting; in the univariate setting, it reduces to a one-node graph without spatial interactions.}\par}
\end{table}

\section{Dataset Details}
To comprehensively evaluate epidemic forecasting performance under partial observability, time-varying dynamics, and physics mismatch, we benchmark our approach on a diverse suite of synthetic and real-world datasets, spanning multiple disease regimes, observation settings, and experimental conditions.

\subsection{Synthetic Datasets}
\paragraph{SIRS with time-fixed and time-varying parameters. }
We generate synthetic epidemics using the SIRS model under multiple parameter regimes. In the fixed setting, transmission, recovery, and immunity-loss rates remain constant over time, serving as a baseline for identifiability under stationarity.
In the periodic setting, parameters vary smoothly and periodically to emulate seasonal forcing, capturing recurring epidemic patterns.

\paragraph{Different epidemic physics. }
To assess robustness to physics mismatch, we additionally simulate data from alternative compartmental models. The SIR setting removes immunity waning, testing the model’s ability to adapt when the assumed SIRS structure is overparameterized. 
The SEIRS setting introduces an exposed compartment, increasing latent-state complexity and evaluating performance when the true dynamics deviate from the assumed model class.

\subsection{Real-World Datasets}

Influenza-like illness (ILI) surveillance data (\href{https://gis.cdc.gov/grasp/fluview/fluportaldashboard.html}{https://gis.cdc.gov/grasp/fluview/fluportaldashboard.html}) provides a canonical example of strongly seasonal epidemic dynamics.
We consider two settings: single-wave segments, obtained by isolating individual seasonal outbreaks, and multi-wave sequences spanning multiple years. This allows evaluation of both early outbreak forecasting and long-term seasonal recurrence. We evaluate across multiple U.S. Department of Health and Human Services (HHS) regions (HHS~1--10, covering the entire continental United States) and across multiple time periods, capturing heterogeneity in epidemic progression and reporting practices.

\section{Choice of Train--Test Splits}
\label{sec:splits-detail}

\model{} follows a principled, regime-aware evaluation design tailored to epidemic dynamics, with split placement determined by the underlying wave structure of each dataset.

\paragraph{Early-stage splits for single-peak dynamics.}
For single-wave epidemics, the realistic forecasting task is early-stage prediction. We use small training fractions (e.g., $0.1$ for SIR) 
so that the model is required to extrapolate the epidemic trajectory \emph{through and beyond} the peak rather than merely interpolating between observed peaks.

\paragraph{Cycle-aware splits for multi-peak dynamics.}
For multi-wave epidemics the goal is to forecast future waves from historical cycles. We place $t_{\mathrm{split}}$ after observing one or more prior cycles, typically near the start of a new wave to be forecasted (e.g., $0.6$ for SIRS (Varying), $0.7$ for ILI), so the model must generalize from prior cycles to subsequent ones.



\section{Implementation via Bounded Parameterization.}
Rather than fixing parameters or imposing strong smoothness priors, we implement these ranges through a bounded neural parameterization, where raw network outputs are mapped via a sigmoid function and affine scaling into the prescribed intervals. This approach follows prior work on constrained neural modeling of dynamical systems and epidemic processes \cite{raissi2019physics,neural-odes}. By encoding epidemiological knowledge as soft constraints, the model remains expressive while producing parameter trajectories that are interpretable, numerically stable, and consistent with known disease characteristics.

\section{Hyperparameter selection.}
The number of VMD modes $K$ and bandwidth penalty $\alpha$ are selected per dataset via grid search. 
The time-delay embedding parameters $(\tau, m)$ are similarly tuned per dataset. 

\section{Supplementary Results}
\label{sec:app_results}

\subsection{Performance Evaluation}
\label{sec:evaluation_appendix}

\subsubsection{Forecasting Accuracy}
\label{sec:app_forecast_accuracy}

\begin{figure*}[htbp]
\centering
\makebox[\textwidth][c]{%
\begin{subfigure}[t]{0.42\textwidth}
    \centering
    \includegraphics[width=\linewidth]{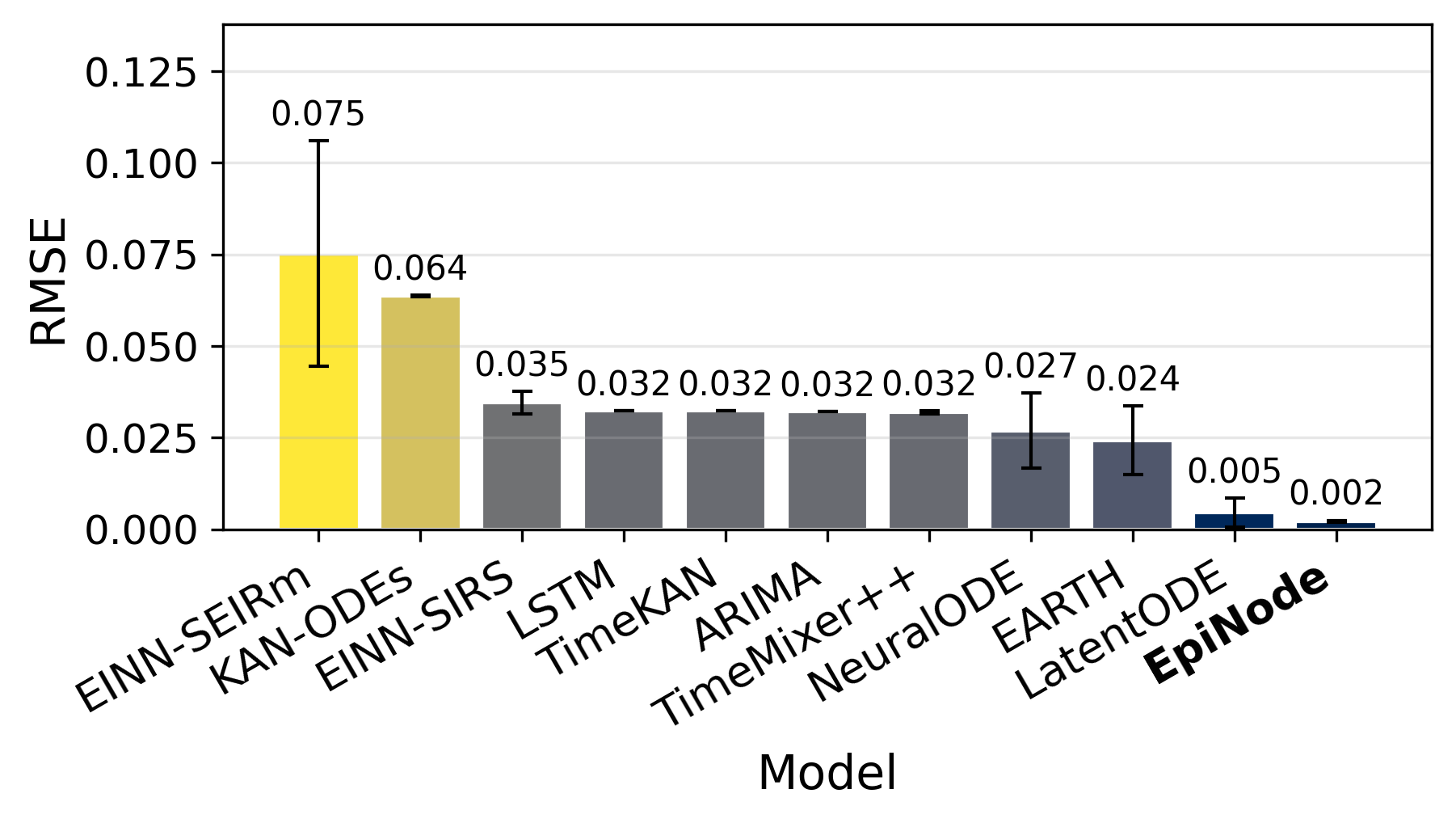}
    \caption{SIRS (Fixed)}
    \label{fig:benchmark_overall_sirs_fixed}
\end{subfigure}
\hspace{0.04\textwidth}
\begin{subfigure}[t]{0.42\textwidth}
    \centering
    \includegraphics[width=\linewidth]{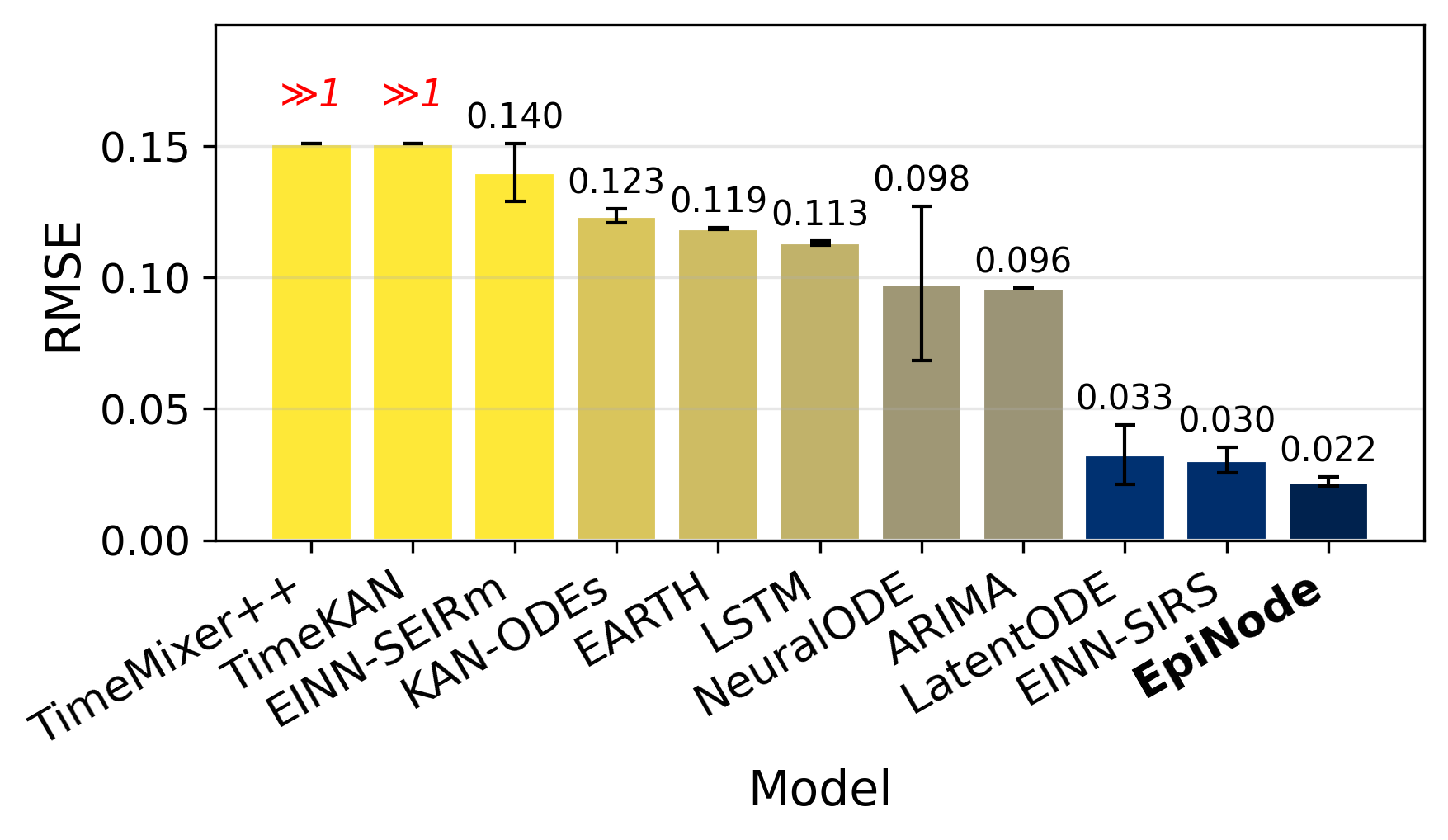}
    \caption{SIR}
    \label{fig:benchmark_overall_sir}
\end{subfigure}%
}

\caption{Overall benchmark RMSE on stationary synthetic datasets.}
\label{fig:benchmark_overall_rmse_synthetic}
\end{figure*}

Figure~\ref{fig:benchmark_overall_rmse_synthetic} complements the benchmark comparison (Figure~\ref{fig:benchmark_overall_rmse}) with overall RMSE on two time-fixed synthetic datasets (SIRS, SIR). \model{} achieves the lowest mean RMSE on all three datasets. 
The win on SIRS (Fixed) shows that the model is accurate when the assumed physics matches the data, and the wins on SIR (Fixed) and SEIRS (Fixed) show that the latent control signals absorb the residual structure introduced by the mismatch, so the advantage extends from non-stationary settings highlighted in the main paper to stationary regimes with both matched and mismatched physics. Univariate Transformer-style forecasters (TimeKAN, TimeMixer++) are off-scale on SIR (Fixed).

\subsubsection{Peak Errors}
\label{sec:peak_results}

\begin{figure}[htbp]
    \centering
    \includegraphics[width=.9\columnwidth]{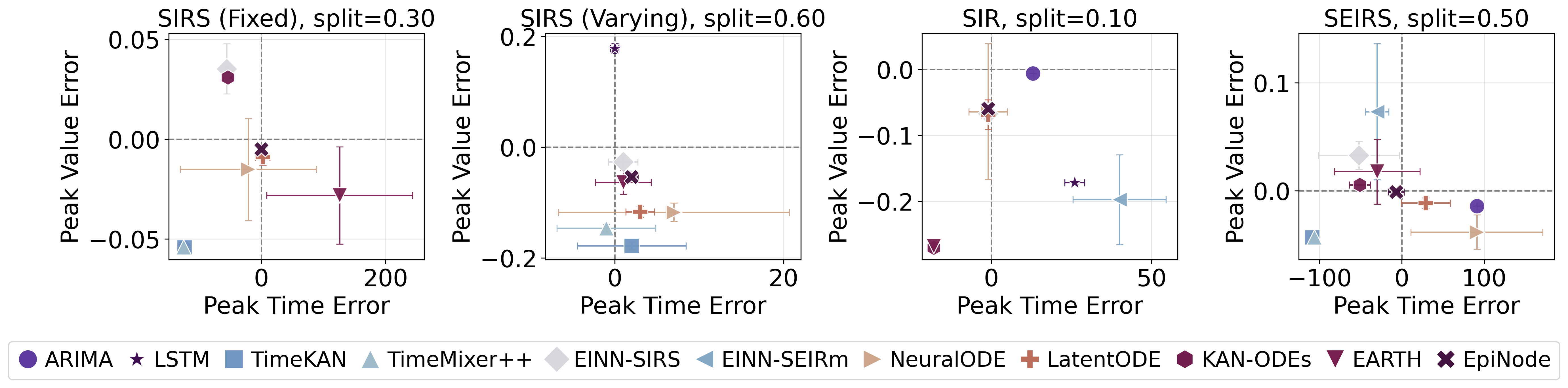}
    \caption{Peak error (magnitude \& timing) across synthetic datasets} 
    \label{fig:peak}
\end{figure}

 Figure~\ref{fig:peak} and Tables~\ref{tab:peak_timing},~\ref{tab:peak_magnitude} report peak time and peak value errors on the synthetic datasets. 
\model{} achieves the smallest mean peak-timing error on four of the five datasets and the smallest or second-smallest mean peak-magnitude error on all five.
 In contrast, several baselines attain competitive pointwise RMSE but fail to accurately capture peak behavior. Errors in the inferred growth dynamics accumulate throughout the forecast horizon and are magnified near turning points, leading to systematically early peak predictions when growth is overestimated and delayed peaks when transmission dynamics are overly damped.

\renewcommand{\arraystretch}{1}
\begin{table*}[t]
\centering
\small
\caption{Peak time error across synthetic and real datasets (RMSE, mean $\pm$ std).}
\label{tab:peak_timing}
\begin{tabular}{lccccc}
\toprule
\textbf{Method} & \textbf{SIRS (Fixed)} & \textbf{SIRS (Varying)} & \textbf{SIR (Fixed)} & \textbf{SEIRS (Fixed)} & \textbf{ILI} \\
\midrule
ARIMA
& $-$125.0 & 147.0 & 13.0 & 91.0 & 28.0 \\

LSTM
& $-$125.0 (0.0) & \textbf{0.0} (0.5) & 26.0 (3.1) & $-$33.0 (4.8) & 3.0 (12.3) \\



EINN-SIRS
& $-$56.0 (3.2) & \underline{1.0} (1.7) & $-$1.0 (0.8) & $-$52.0 (49.3) & $-$14.0 (8.9) \\

EINN-SEIRm
& 131.0 (63.1) & 118.0 (43.6) & 40.0 (14.5) & $-$30.0 (14.1) & \underline{3.0} (12.9) \\

NeuralODE
& $-$21.0 (109.4) & 7.0 (13.7) & $-$1.0 (6.0) & 91.0 (80.0) & 28.0 (15.8) \\

LatentODE
& \underline{2.0} (11.7) & 3.0 (1.7) & \underline{$-$1.0} (1.0) & \underline{29.0} (29.7) & 28.0 (16.7) \\

KAN-ODEs
& $-$54.0 (6.6) & $-$13.0 (0.0) & $-$18.0 (0.0) & $-$51.0 (12.8) & 28.0 (10.2) \\

EARTH
& 126.0 (117.4) & 1.0 (3.3) & $-$18.0 (0.0) & $-$30.0 (52.0) & 28.0 (0.0) \\

\midrule
\textbf{EpiNode (Ours)}
& \textbf{0.0} (1.6)
& 2.0 (0.7)
& \textbf{$-$1.0} (0.0)
& \textbf{$-$7.0} (9.5)
& \textbf{0.0} (0.7) \\
\bottomrule
\end{tabular}

\vspace{2pt}
{\footnotesize\raggedright\textit{\ \ \ \ \ \ \ \ \ \ \ \ \ \ \ \ \ \ \ \ \ \ \ \ \ Note: ARIMA is deterministic and therefore no standard deviation is reported. }\par}
\end{table*}

\renewcommand{\arraystretch}{1}
\begin{table*}[t]
\centering
\small
\caption{Peak value error across synthetic and real datasets (RMSE, mean $\pm$ std).}
\label{tab:peak_magnitude}
\begin{tabular}{lccccc}
\toprule
\textbf{Method} & \textbf{SIRS (Fixed)} & \textbf{SIRS (Varying)} & \textbf{SIR (Fixed)} & \textbf{SEIRS (Fixed)} & \textbf{ILI} \\
\midrule

ARIMA
& -0.0544 & 1.1983 & \textbf{$-$0.0063} & -0.0139 & -0.0567 \\

LSTM
& $-$0.0550 (0.0001) & 0.1779 (0.0086) & $-$0.1720 (0.0019) & 0.1475 (0.0014) & $-$0.0375 (0.0317) \\



EINN-SIRS
& 0.0352 (0.0125) & \textbf{$-$0.0267} (0.0113) & $-$0.0674 (0.0104) & 0.0328 (0.0129) & $-$0.0630 (0.0087) \\

EINN-SEIRm
& 0.1436 (0.0780) & 0.0722 (0.0888) & $-$0.1978 (0.0682) & 0.0734 (0.0629) & \underline{$-$0.0184} (0.0531) \\

NeuralODE
& $-$0.0151 (0.0256) & $-$0.1173 (0.0165) & $-$0.0640 (0.1029) & $-$0.0381 (0.0160) & $-$0.0604 (0.0228) \\

LatentODE
& \underline{$-$0.0090} (0.0041) & $-$0.1164 (0.0100) & $-$0.0687 (0.0224) & \underline{$-$0.0112} (0.0049) & $-$0.0588 (0.0041) \\

KAN-ODEs
& 0.0309 (0.0003) & $-$0.1130 (0.0069) & $-$0.2707 (0.0023) & 0.0055 (0.0003) & $-$0.0547 (0.0001) \\

EARTH
& $-$0.0282 (0.0243) & $-$0.0633 (0.0214) & $-$0.2679 (0.0002) & 0.0179 (0.0300) & $-$0.0603 (0.0007) \\

\midrule
\textbf{EpiNode (Ours)}
& \textbf{$-$0.0050} (0.0003)
& \underline{$-$0.0535} (0.0084)
& \underline{$-$0.0594} (0.0026)
& \textbf{$-$0.0010} (0.0016)
& \textbf{$-$0.0178} (0.0032) \\
\bottomrule
\end{tabular}

\vspace{2pt}
{\footnotesize\raggedright\textit{\ \ \ \ \ \ \ \ \ Note: ARIMA is deterministic and therefore no standard deviation is reported. }\par}
\end{table*}

\subsection{Applications}
\label{sec:application_appendix}

\subsubsection{Parameter Inference}
\label{sec:inference_appendix}
Across synthetic data with ground truth, \model{} recovers the unobserved $S(t), R(t)$ trajectories (Figure~\ref{fig:compartments}) and the time-varying $\beta(t), \gamma(t), \delta(t)$ within their prescribed bounds (Figure~\ref{fig:parameters}). However, baselines drift, flatten, or produce no parameter outputs. 
On real ILI data across all ten CDC HHS regions (Figure~\ref{fig:hhs_parameter}), the inferred $\beta(t)$ is smooth, bounded, and shows the expected winter-peaking modulation, with consistent regional patterns and no per-region tuning.

\begin{figure}[h]
    \centering
    \includegraphics[width=.9\linewidth]{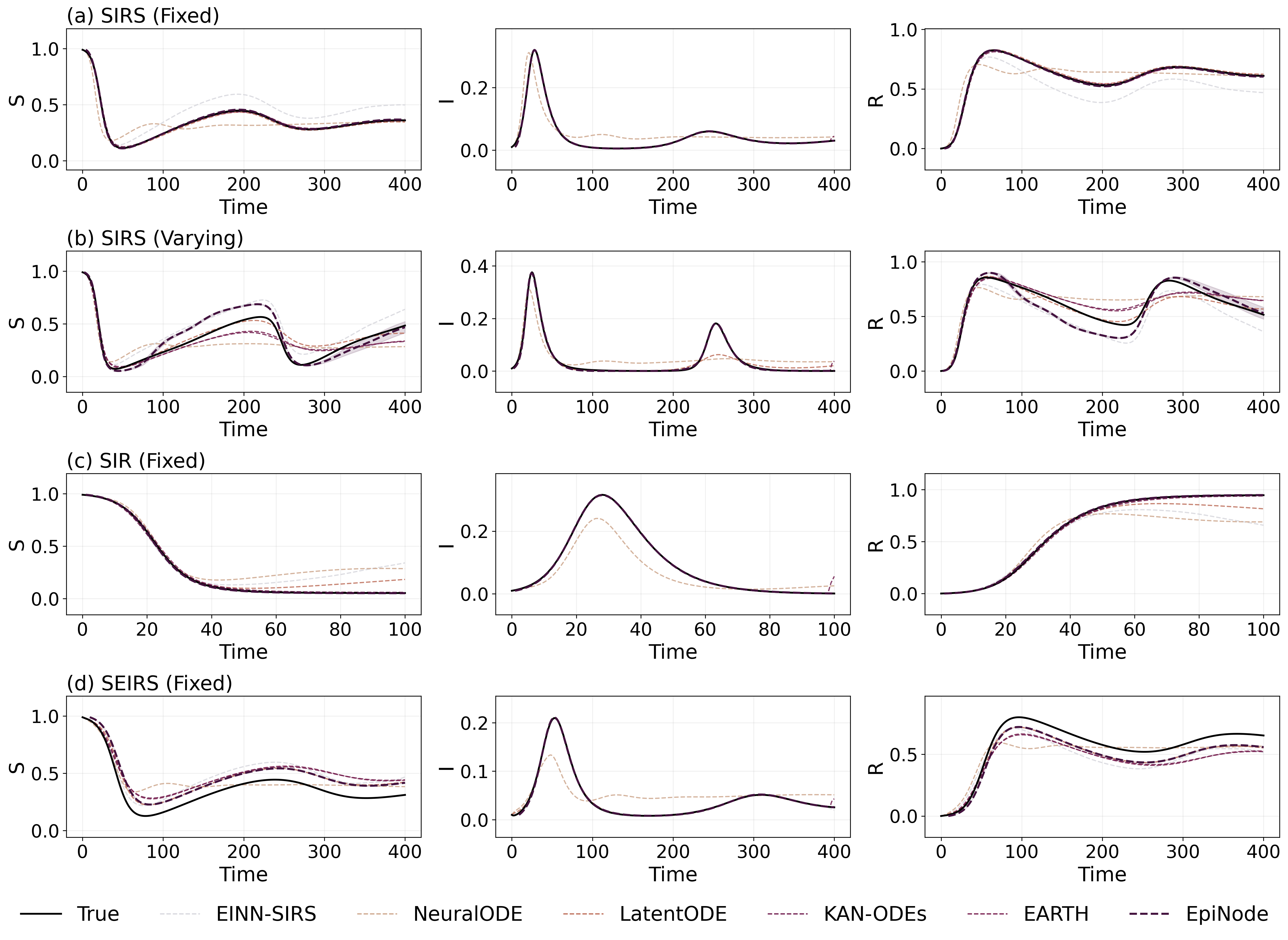}
    \caption{\small True and predicted compartments across synthetic datasets inferred from the full infection series.}
    \label{fig:compartments}
\end{figure}

\begin{figure}[h]
    \centering
    \includegraphics[width=.9\linewidth]{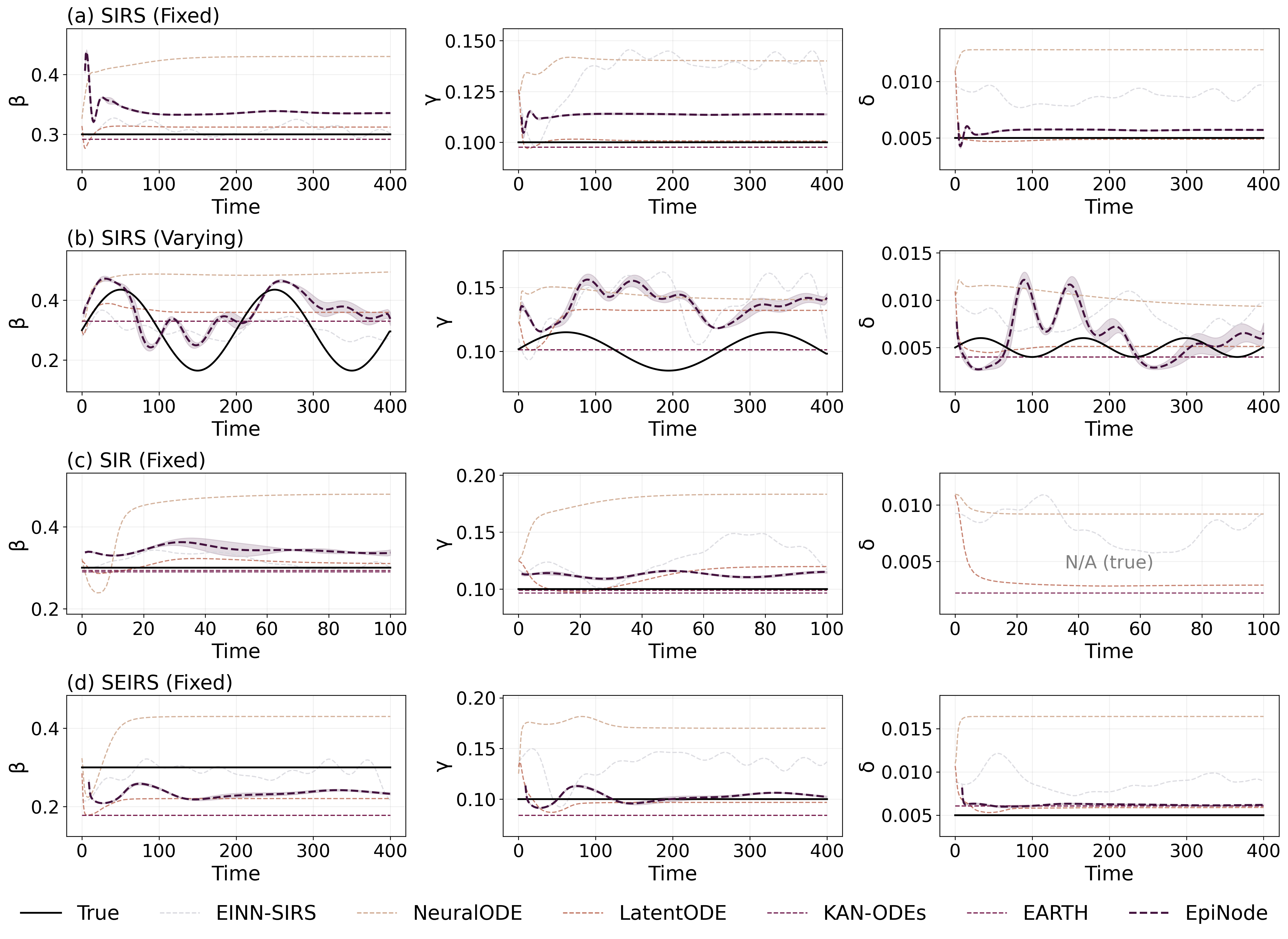}
    \caption{\small True and predicted parameters across synthetic datasets inferred from the full infection series.}
    \label{fig:parameters}
\end{figure}

\begin{figure*}[h]
    \centering
    \begin{subfigure}[h]{0.217\textwidth}
        \centering
        \includegraphics[width=\linewidth]{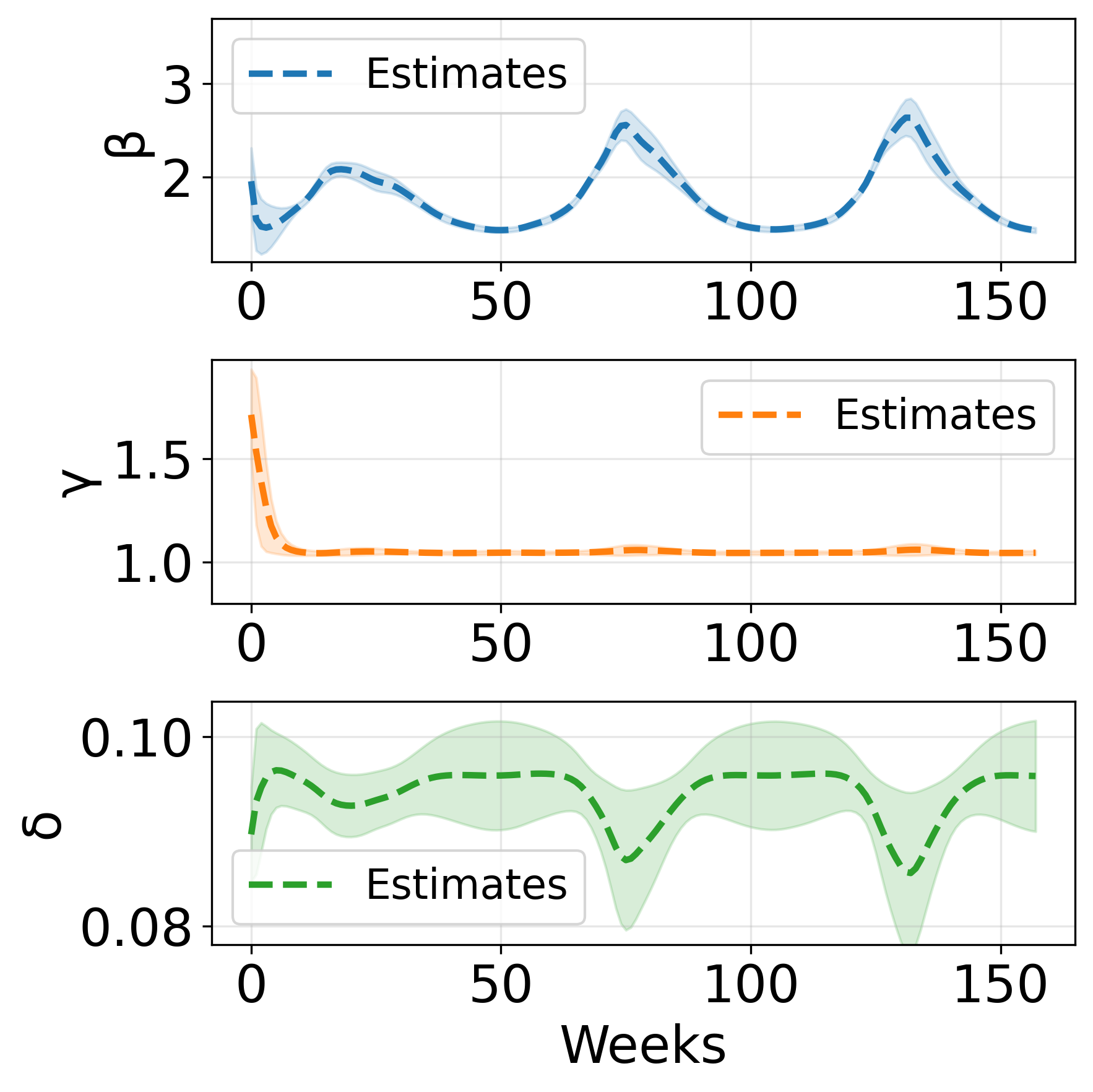}
        \caption{\small ILI (HHS 4)}
        \label{fig:params_hhs4}
    \end{subfigure}\hfill
    \begin{subfigure}[h]{0.77\textwidth}
        \centering
        \includegraphics[width=\linewidth]{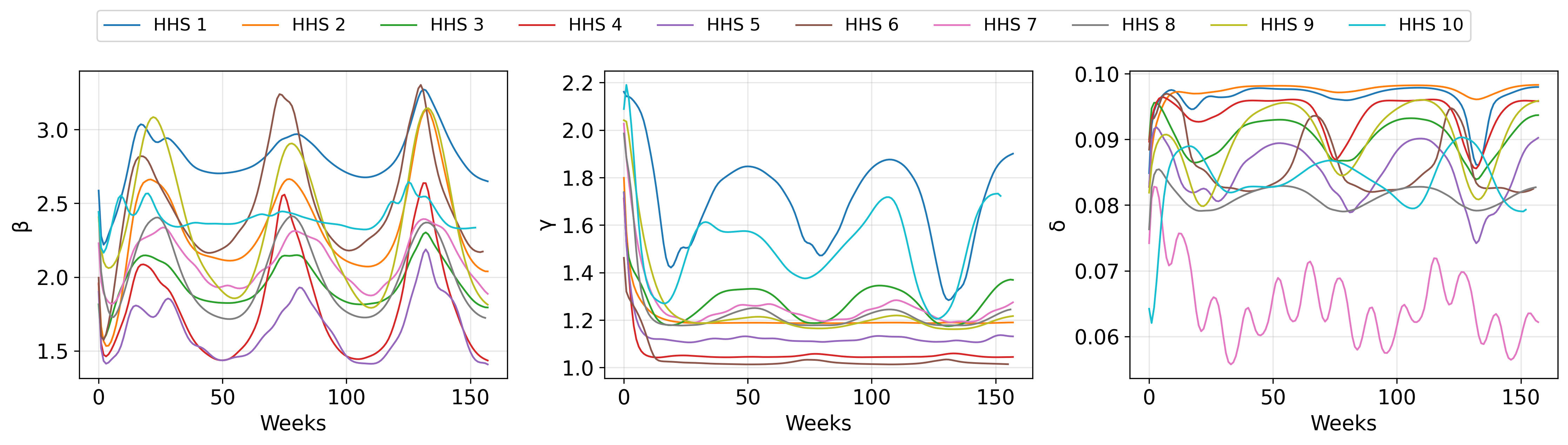}
        \caption{\small ILI (All 10 HHS regions)}
        \label{fig:params_all_hhs}
    \end{subfigure}

    \caption{
    Predicted time-varying parameters inferred from the full infection series.
    (a) Estimated parameter rates for a representative HHS region (HHS~4).
    (b) Estimated parameter rates across all ten HHS regions.
    }
    \label{fig:hhs_parameter}
\end{figure*}

\subsubsection{Regional Dynamics}
\label{sec:hhs}


\begin{figure*}[h]
    \centering
    \begin{subfigure}[t]{0.198\textwidth}
        \centering
        \includegraphics[width=\linewidth]{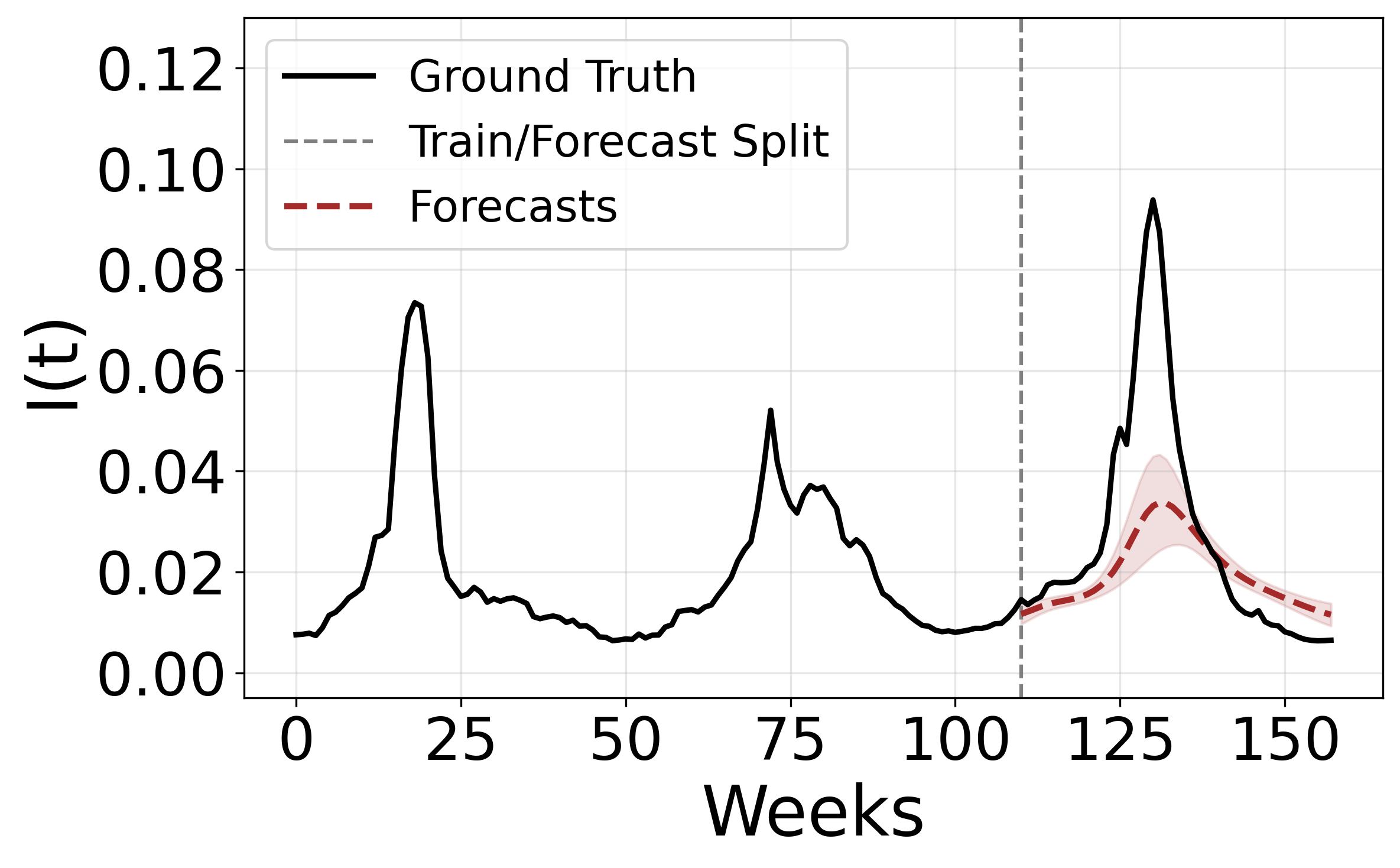}
        \caption{\small HHS 1}
    \end{subfigure}\hfill
    \begin{subfigure}[t]{0.198\textwidth}
        \centering
        \includegraphics[width=\linewidth]{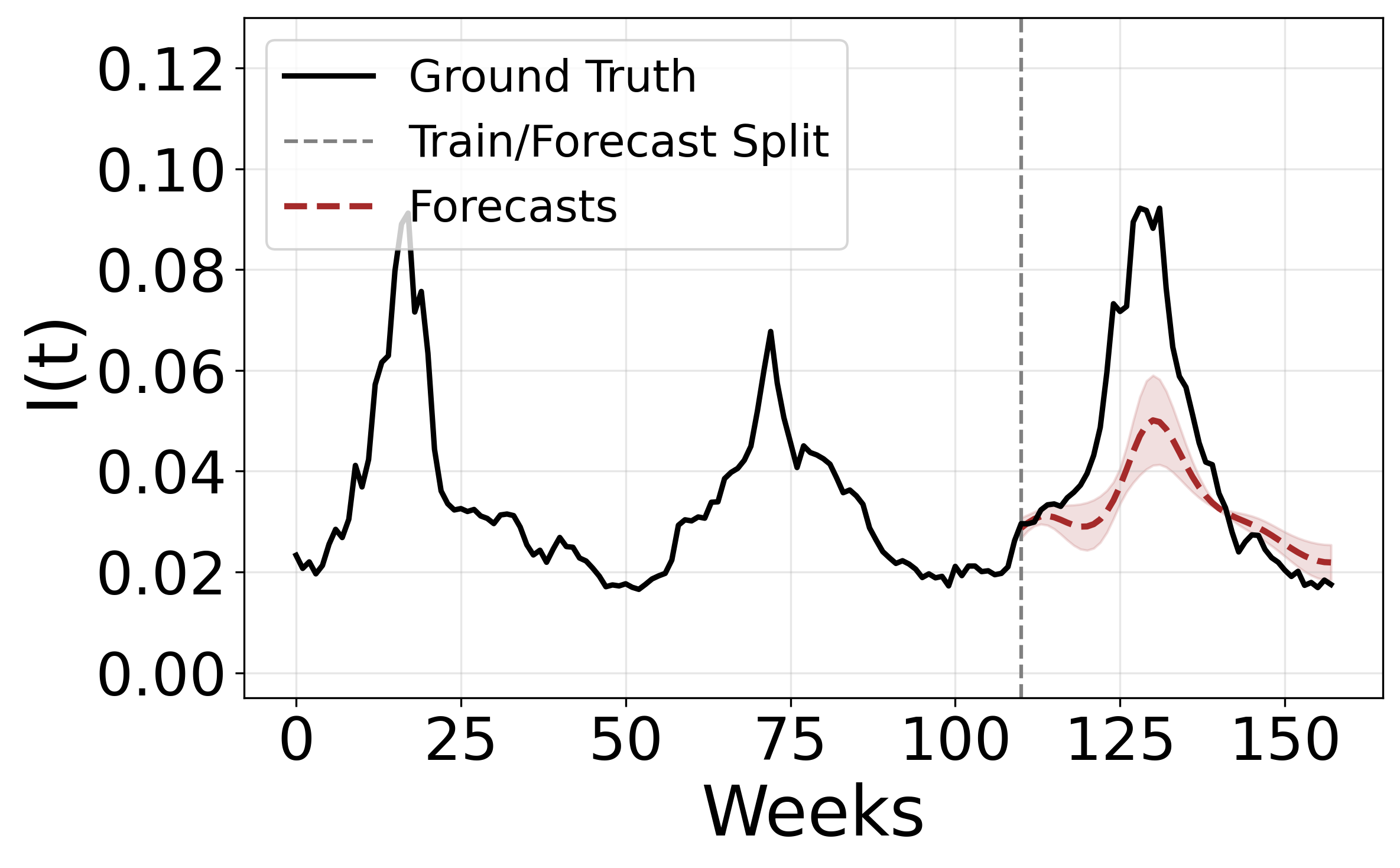}
        \caption{\small HHS 2}
    \end{subfigure}\hfill
    \begin{subfigure}[t]{0.198\textwidth}
        \centering
        \includegraphics[width=\linewidth]{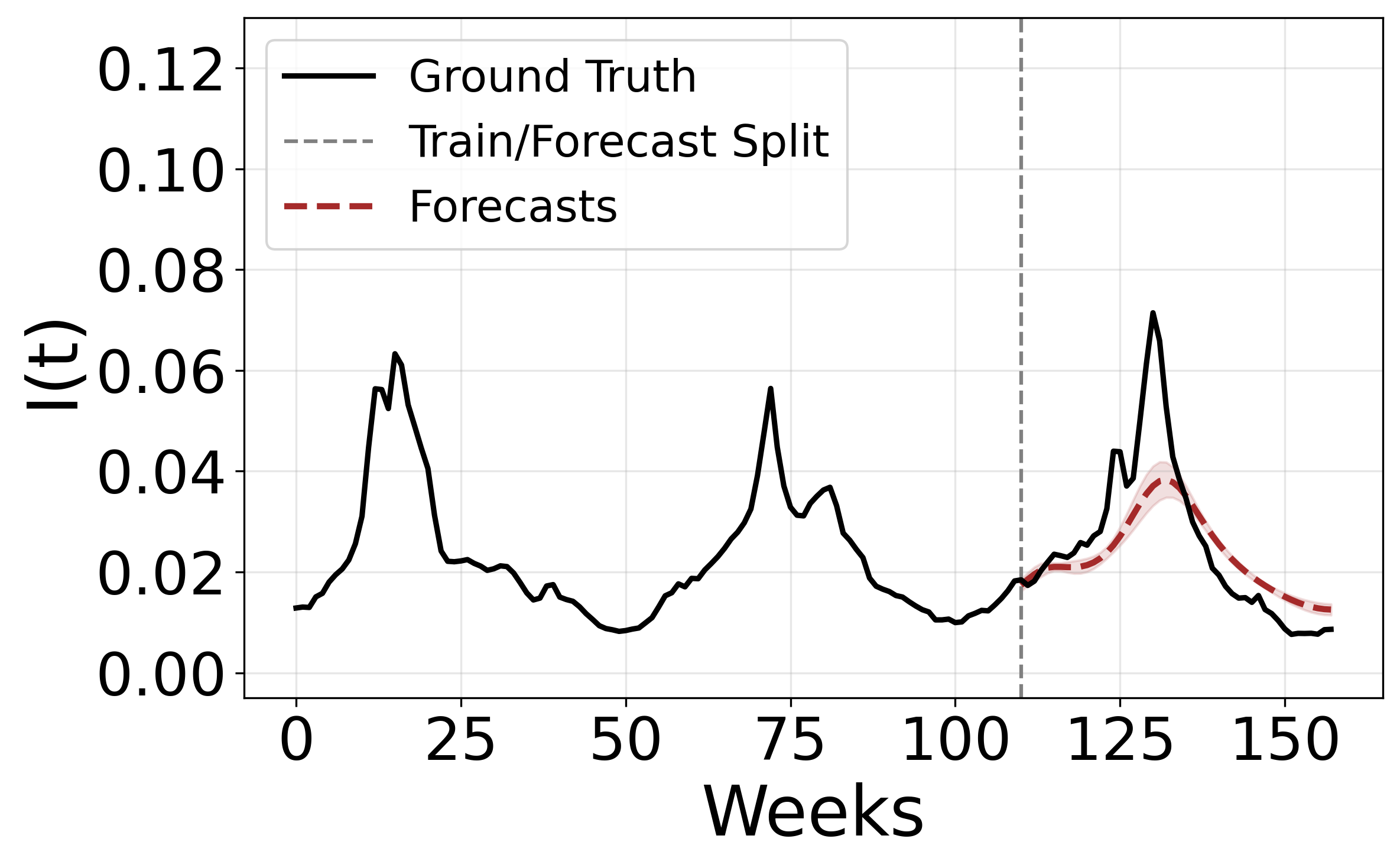}
        \caption{\small HHS 3}
    \end{subfigure}\hfill
    \begin{subfigure}[t]{0.198\textwidth}
        \centering
        \includegraphics[width=\linewidth]{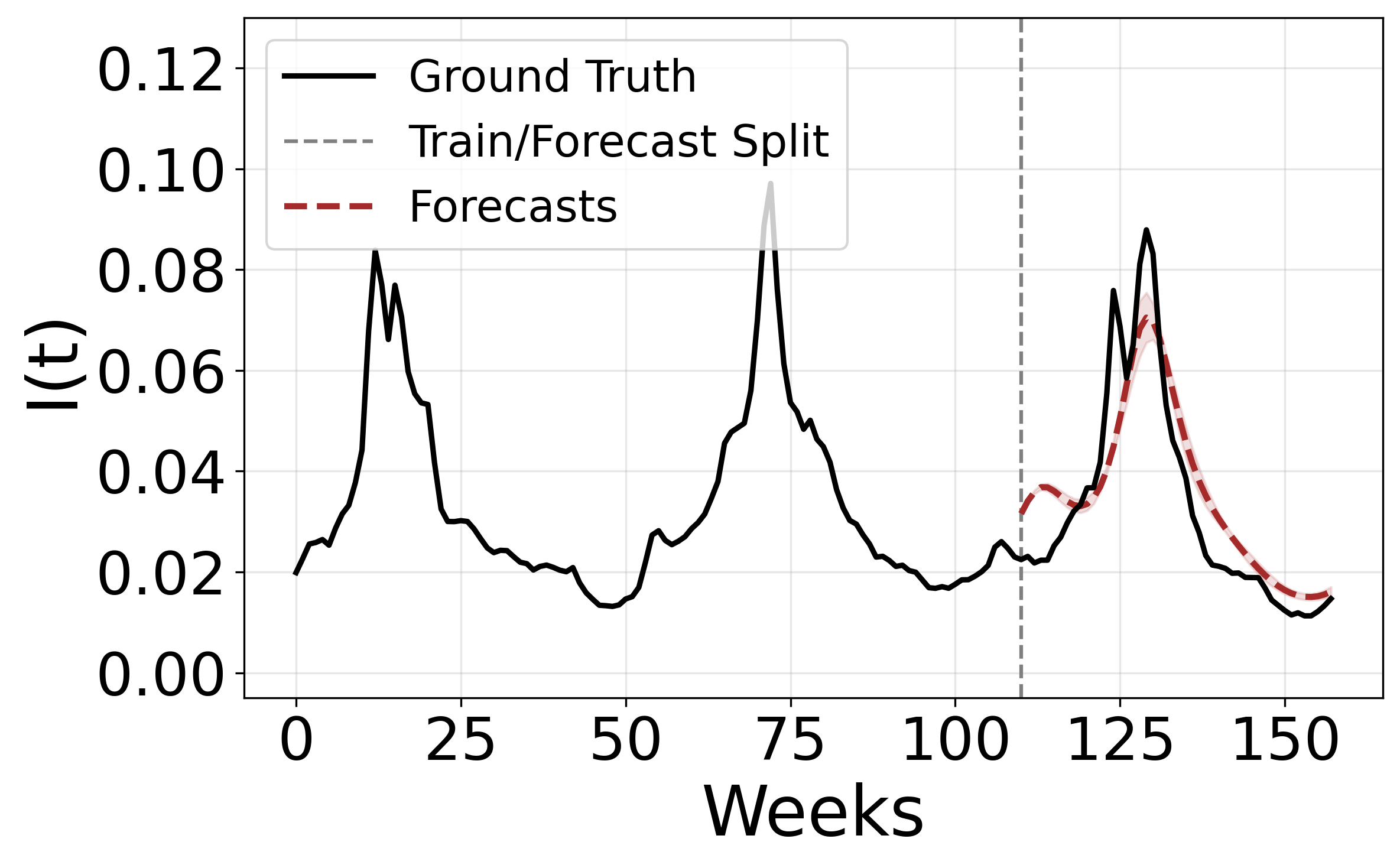}
        \caption{\small HHS 4}
    \end{subfigure}\hfill
    \begin{subfigure}[t]{0.198\textwidth}
        \centering
        \includegraphics[width=\linewidth]{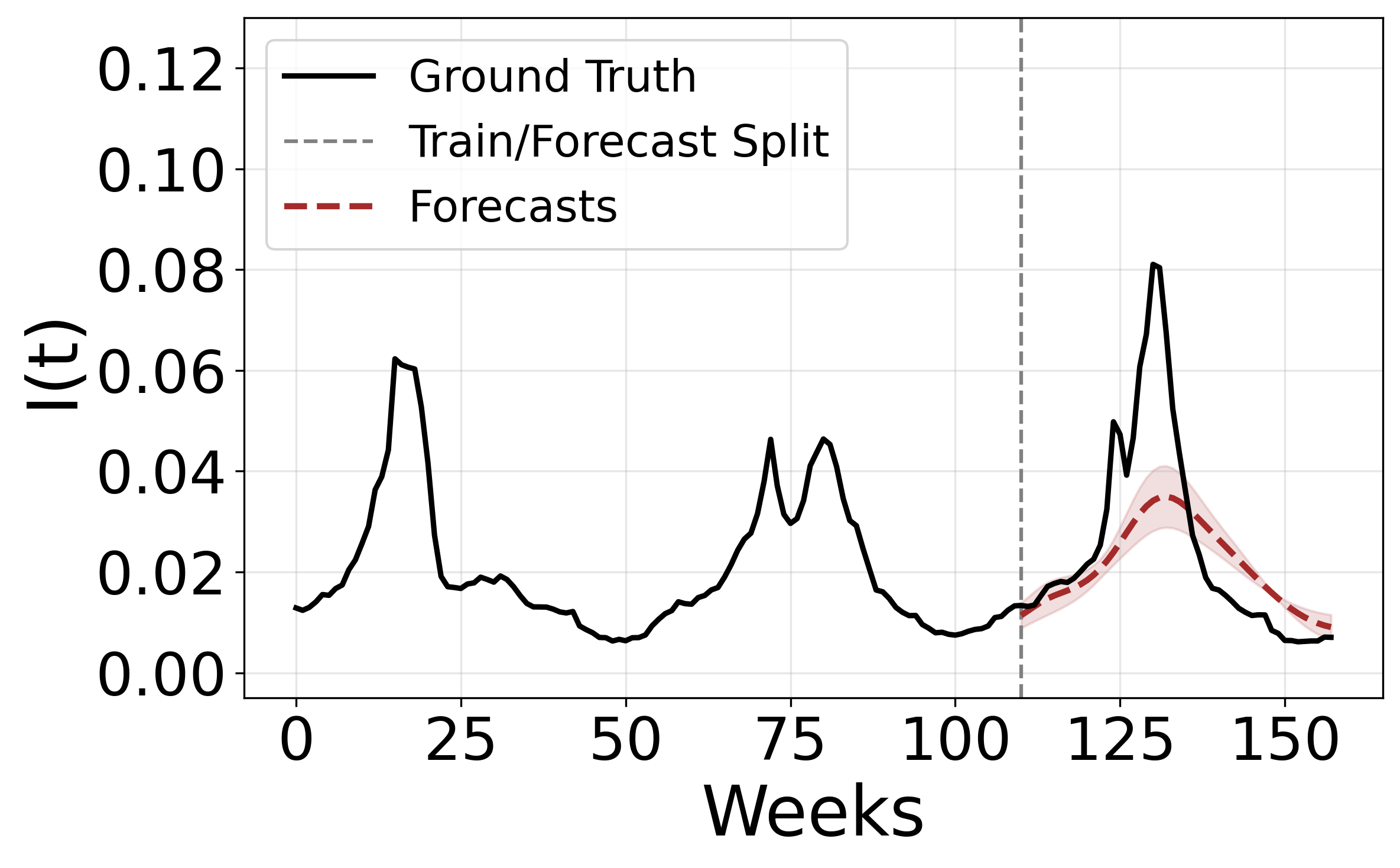}
        \caption{\small HHS 5}
    \end{subfigure}

    \vspace{2mm}

    \begin{subfigure}[t]{0.198\textwidth}
        \centering
        \includegraphics[width=\linewidth]{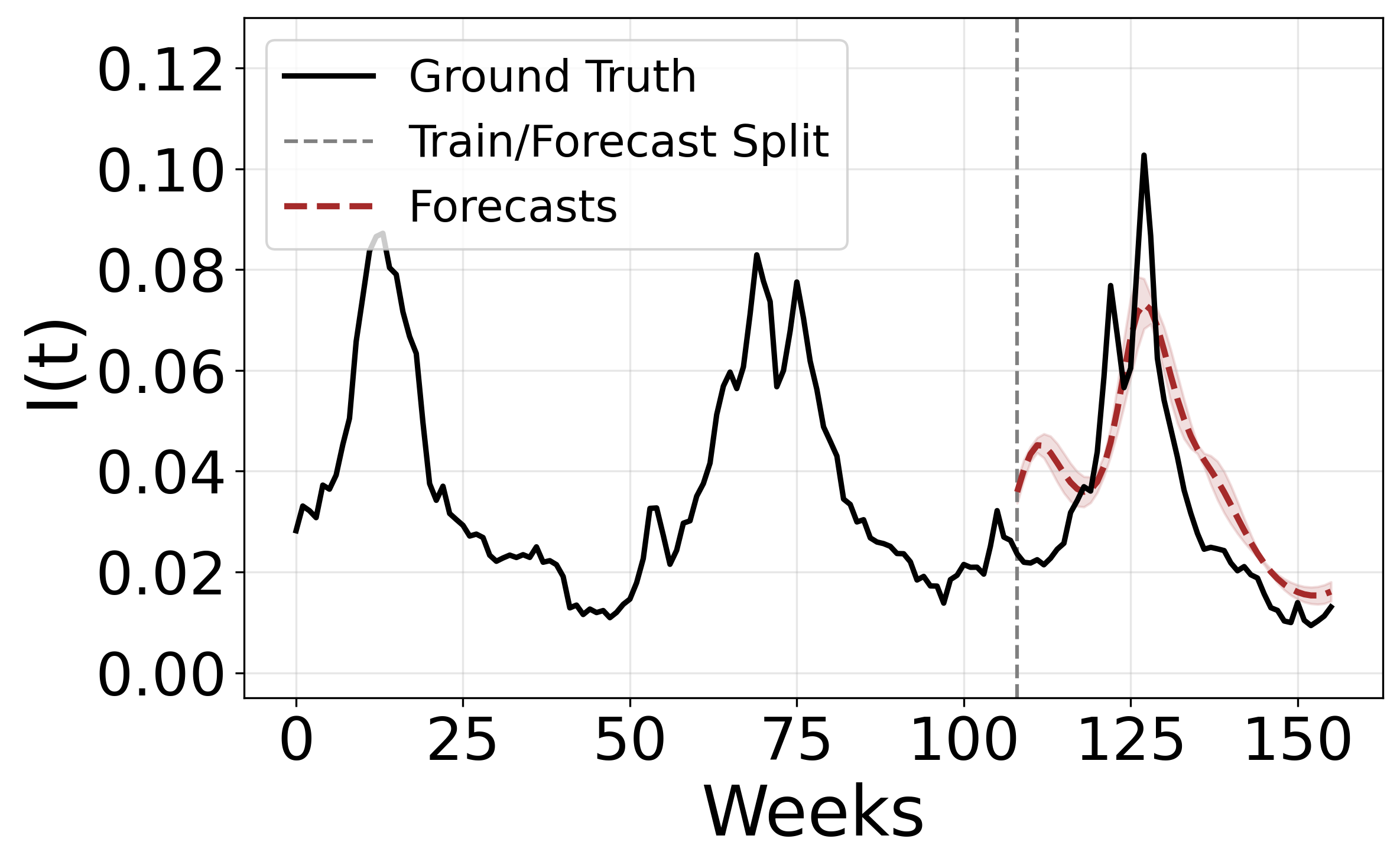}
        \caption{\small HHS 6}
    \end{subfigure}\hfill
    \begin{subfigure}[t]{0.198\textwidth}
        \centering
        \includegraphics[width=\linewidth]{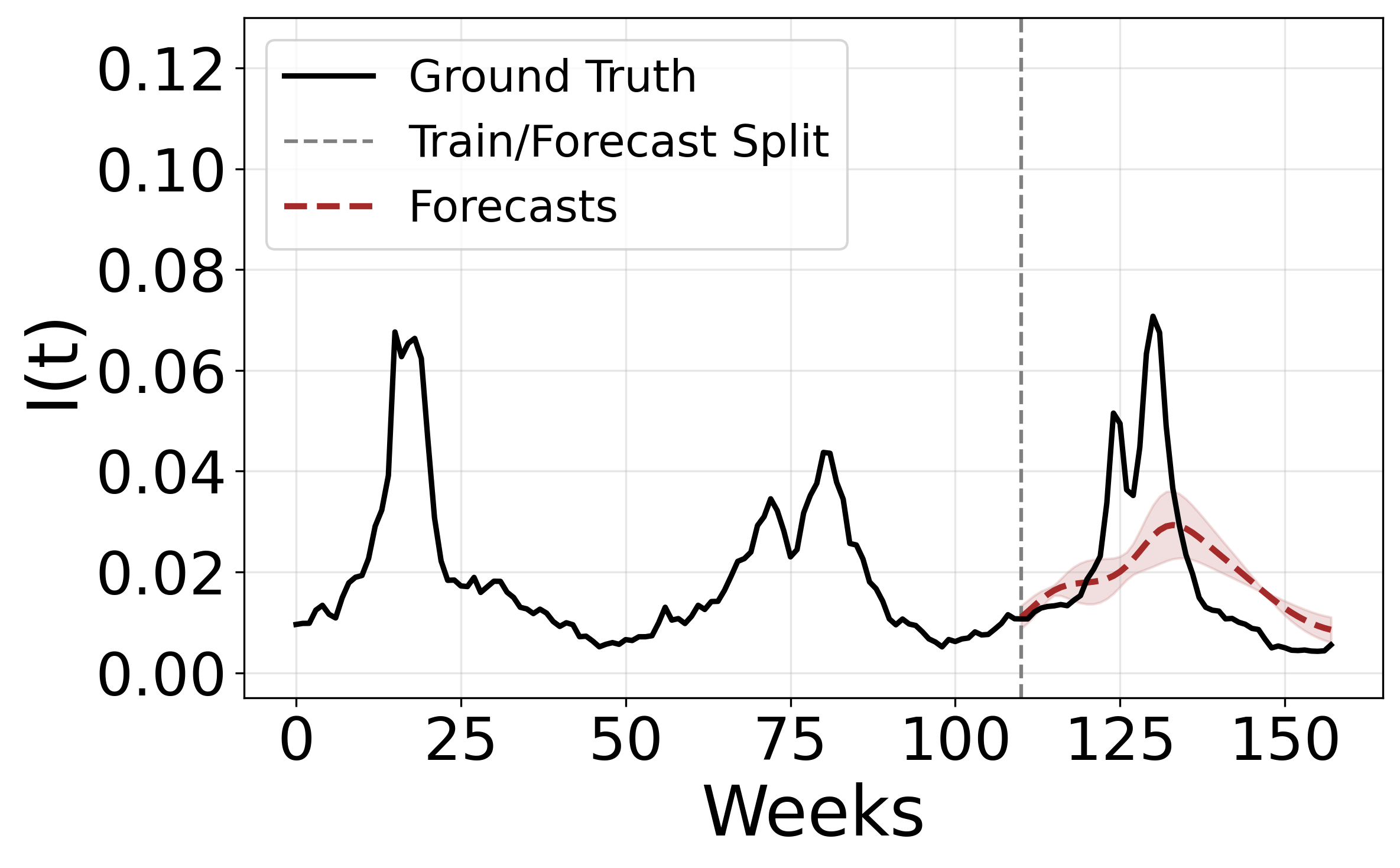}
        \caption{\small HHS 7}
    \end{subfigure}\hfill
    \begin{subfigure}[t]{0.198\textwidth}
        \centering
        \includegraphics[width=\linewidth]{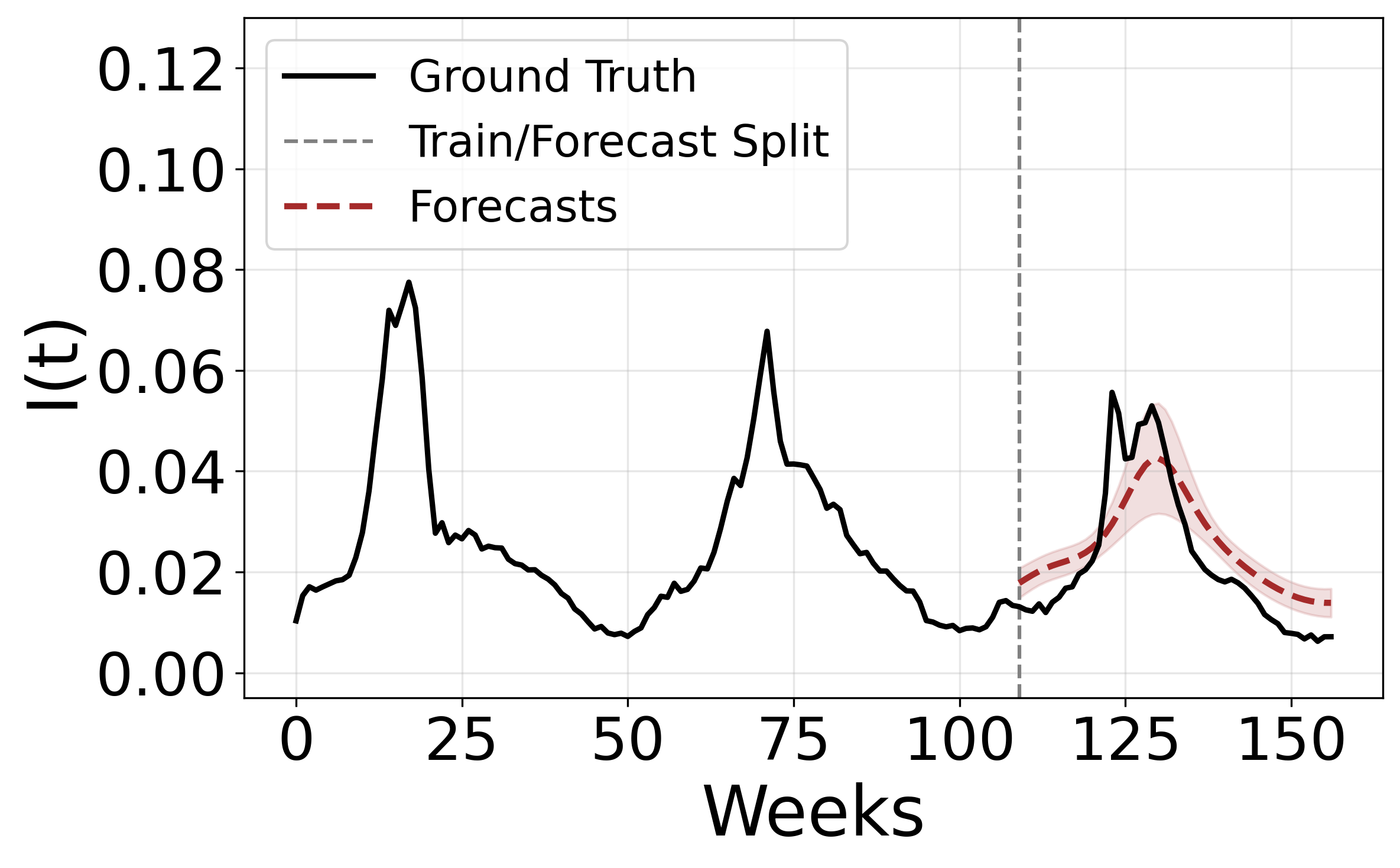}
        \caption{\small HHS 8}
    \end{subfigure}\hfill
    \begin{subfigure}[t]{0.198\textwidth}
        \centering
        \includegraphics[width=\linewidth]{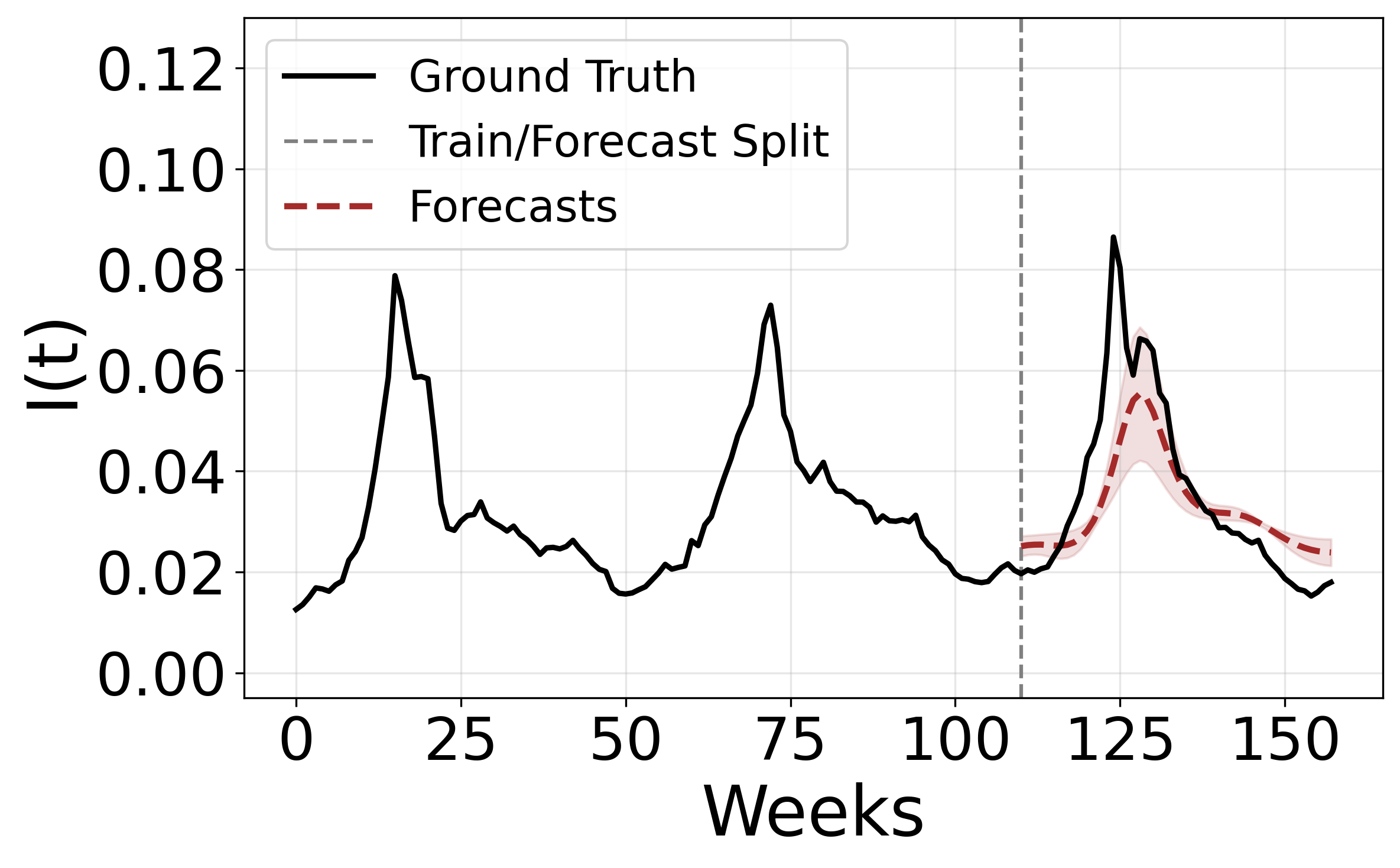}
        \caption{\small HHS 9}
    \end{subfigure}\hfill
    \begin{subfigure}[t]{0.198\textwidth}
        \centering
        \includegraphics[width=\linewidth]{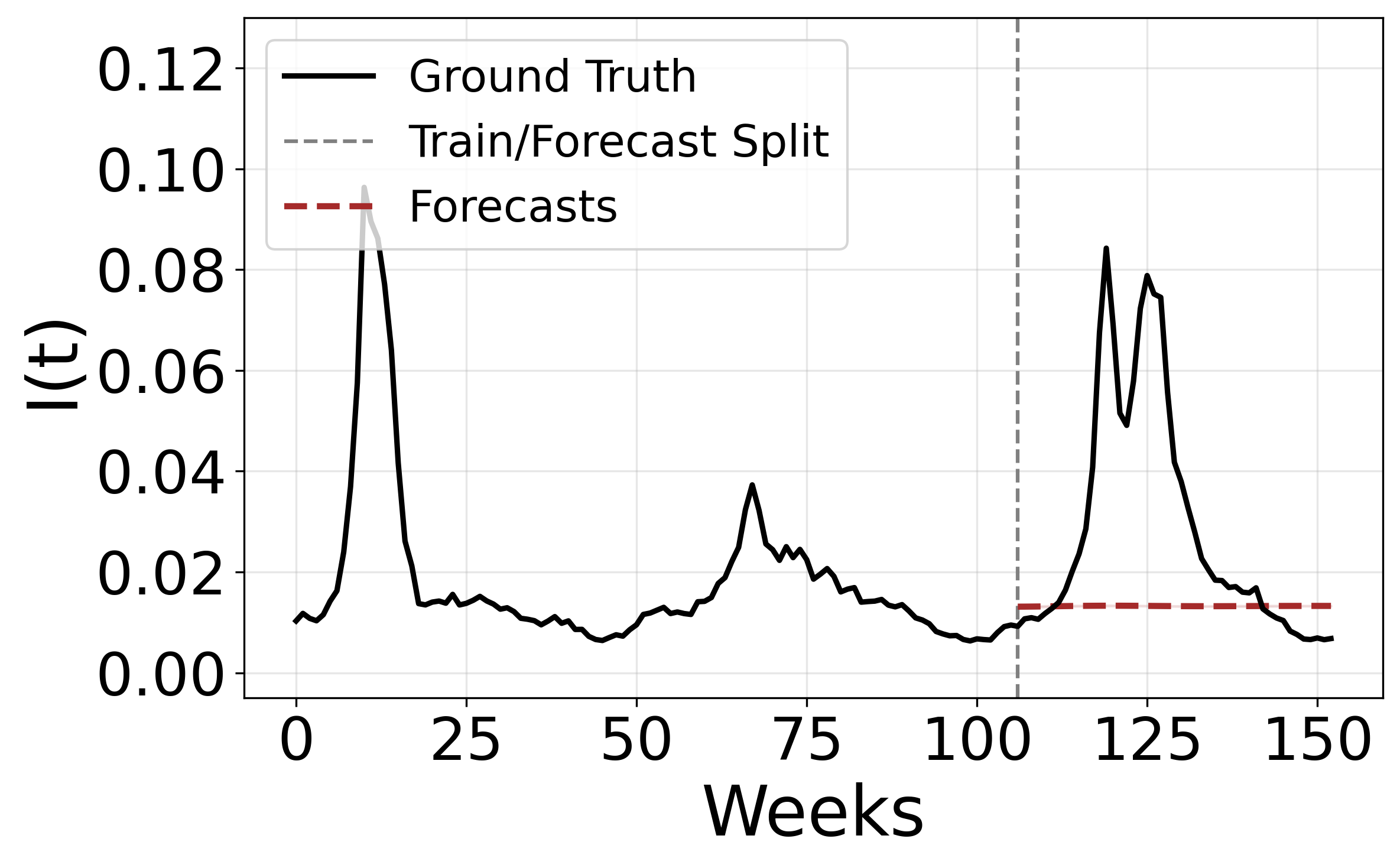}
        \caption{\small HHS 10}
    \end{subfigure}

    \caption{Structured regional dynamics enable accurate forecasting across all HHS regions at split $0.7$.}
    \label{fig:hhs_all}
\end{figure*}

Figure~\ref{fig:hhs_all} shows the forecasts across all ten HHS regions.
In HHS 4 and HHS 6, the proposed model accurately captures both the timing and magnitude of the post-split infection peak, because these regions have regular seasonal structure and smoothly varying trends that the TSR decomposition cleanly separates. 
In contrast, forecasts in HHS 7 and HHS 10 are less accurate, particularly around the post-split surge. These regions exhibit more irregular dynamics and abrupt changes in peak amplitude that are weakly represented in the training window, producing larger residual components and a pronounced distribution shift at forecast time. The inferred transmission dynamics therefore underestimate the rapid increase and underpredict the epidemic peak. These findings show that forecast accuracy depends strongly on the structural regularity of regional epidemic dynamics.

\subsection{Ablation Studies}\label{sec:ablation}
We conduct extensive ablations to isolate the contribution of key architectural components.

\begin{figure}[h]
    \centering
    \begin{subfigure}[t]{0.7\textwidth}
        \centering
        \includegraphics[width=\linewidth]{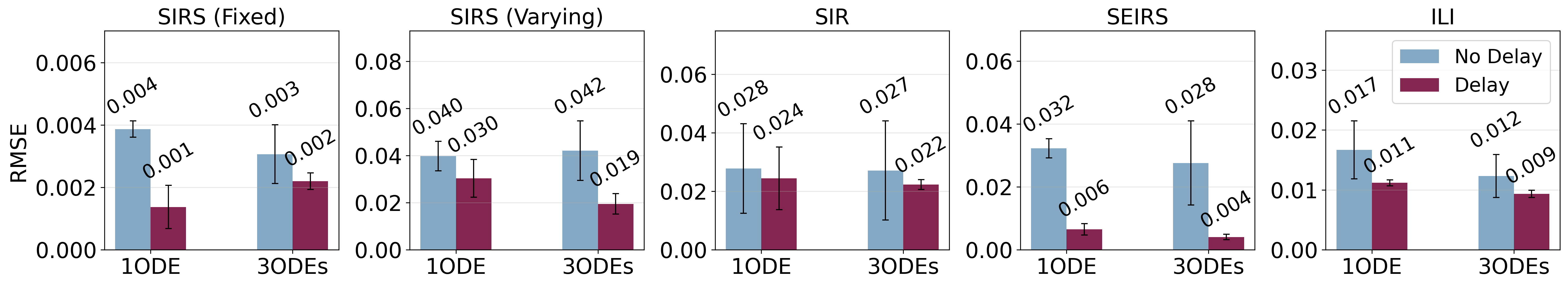}
        \caption{Architecture \& time-delay ablation}
        \label{fig:ablation_arch_delay}
    \end{subfigure}\hfill
    \begin{subfigure}[t]{0.28\textwidth}
        \centering
        \includegraphics[width=\linewidth, height=2.1cm]{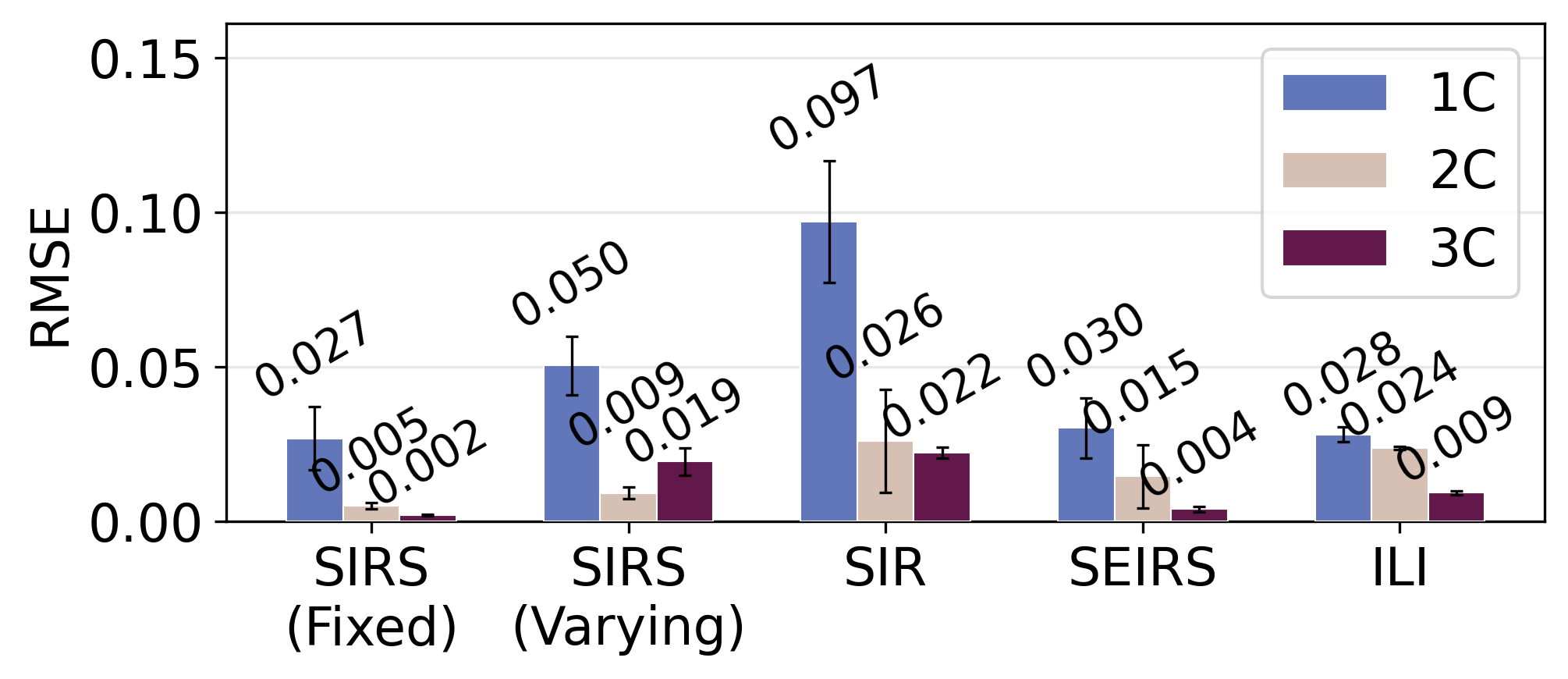}
        \caption{Number of decomposed components}
        \label{fig:ablation_comp}
    \end{subfigure}\hfill
    \begin{subfigure}[t]{0.99\textwidth}
        \centering
        \includegraphics[width=\linewidth]{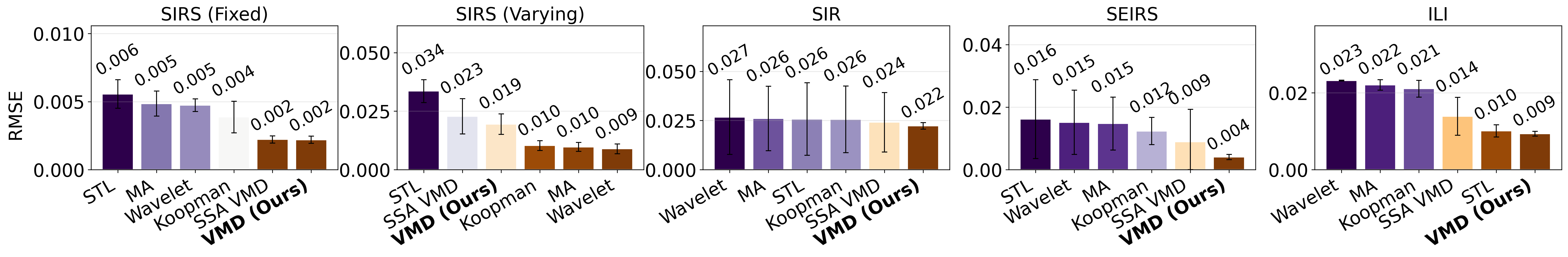}
        \caption{Decomposition methods}
        \label{fig:ablation_decomp}
    \end{subfigure}
    \caption{Ablation results across datasets. (a) Comparison of 1ODE vs. 3ODEs with and without time-delay embedding. (b) Comparison of 1-component $vs$ 2-components $vs$ 3-component decomposition. (c) Comparison of signal decomposition methods.}
\label{fig:ablation_arch_delay_decomp}
\end{figure}

\subsubsection{Single Latent ODE variant}
\label{sec:ablation_arc}

Figure~\ref{fig:ablation_arch_delay} compares architectures with a single latent ODE (1ODE) versus three collaborative latent ODEs (3ODEs) corresponding to trend, seasonal, and residual components. Under identical training conditions, 3ODEs consistently achieves lower error than 1ODE variants, particularly when time-delay embedding is enabled.  This indicates that disentangling multi-scale dynamics into separate latent flows improves identifiability and long-horizon stability.

\subsubsection{Number of Decomposed Components}
\label{sec:ablation_noComp}
Figure~\ref{fig:ablation_comp} shows the comparison among 1 component (C) $vs$ 2C $vs$ 3C. On four of five datasets, RMSE decreases with one to two to three components, suggesting that three components provide the most effective signals to separate slow structural evolution from seasonal forcing. For SIRS (Varying), where two components achieve the lowest RMSE and adding a third slightly increases error, likely because the periodic forcing in this dataset is already well captured by a single seasonal mode.

\subsubsection{Time-delay embedding}
\label{sec:ablation_delay_results}

Incorporating time-delay embedding further improves accuracy by providing temporal context to each component (Figure~\ref{fig:ablation_arch_delay}). An exception occurs in the SIRS setting with time-fixed parameters, where the dynamics are noise-free and stationary; in this case, time-delay embedding offers little benefit, as the system evolution is determined by the current state.

\subsubsection{TSR decomposition method}
\label{sec:ablation_decomp_meth}
Figure~\ref{fig:ablation_decomp} shows that VMD-based decomposition ranks best on four of five datasets, likely due to its explicit frequency localization and robustness to noise. The exception is SIRS (Varying), where VMD places in the middle and Wavelet decomposition performs best.

\end{document}